\documentclass{article}

\usepackage[utf8]{inputenc}
\usepackage{authblk}
\usepackage{setspace}
\usepackage[margin=1.25in]{geometry}
\usepackage{float}
\usepackage{graphicx}
\graphicspath{{./figures/}}
\usepackage{subcaption}

\usepackage{amsmath}
\usepackage{amssymb}
\usepackage{amsfonts}

\usepackage{algorithmic}

\usepackage{textcomp}
\usepackage{xcolor}
\usepackage{tikz}
\usetikzlibrary{positioning,arrows.meta,fit,backgrounds,calc}

\usepackage{lineno}
\usepackage{hyperref}

\usepackage[
    backend=bibtex,
    style=nejm,
    citestyle=numeric-comp,
    sorting=none
]{biblatex}

\title{Vision-Language Models for Egocentric Video:
From Hand-Object Interaction to Embodied AI}

\author[1]{Mohammad Zamani}
\author[1,*]{Fatemeh Ziaeetabar}

\affil[1]{Department of Computer Science,
School of Mathematics, Statistics and Computer Science,
College of Science, University of Tehran,
Tehran, Iran}

\affil[*]{Address correspondence to:
\href{mailto:fziaeetabar@ut.ac.ir}{fziaeetabar@ut.ac.ir}}

\date{}

\begin{document}

\maketitle


\begin{abstract}
Egocentric video captures activities from the wearer's perspective, providing
an action-centered view of human attention, object manipulation, and interaction
with the surrounding environment. This perspective is increasingly important
for assistive technologies, wearable and immersive computing, and embodied
intelligent systems, yet it introduces distinctive challenges, including
continuous camera motion, occlusion, small active objects, viewpoint-dependent
appearance, and long-range temporal dependencies. Vision--language models
(VLMs) offer promising capabilities for addressing these challenges by
connecting visual observations with semantic knowledge and natural-language
supervision. This survey presents a structured and critical review of VLMs for
egocentric video understanding, tracing the progression from conventional
recognition architectures to multimodal foundation models. We organize the
literature around core tasks, datasets, and developments in hand--object
interaction, temporal reasoning, frame and clip selection, multimodal
representation learning, prompting, model adaptation, and semantic alignment.
Particular attention is devoted to graph-based reasoning as a structured means
of modeling spatial, temporal, and semantic relations among hands, objects,
actions, and scene context. We further examine applications ranging from action
recognition and anticipation to video question answering, assistive systems,
and human-to-robot skill transfer. Our synthesis shows that current models
remain more reliable at identifying visible objects than interpreting actions,
evolving interactions, and user intent, especially in long activities. We
identify temporally grounded and relationally explicit reasoning, efficient
long-video processing, cross-domain generalization, multimodal integration,
privacy, and trustworthy evaluation as key research priorities toward robust
and deployable embodied intelligence.

\end{abstract}


\section{INTRODUCTION}

\subsection{Motivation}
Egocentric video, captured from a first-person viewpoint using wearable
cameras, records scenes the way a human experiences them and reveals the
wearer's attention, behavior, and goals, making it a powerful substrate for
analyzing human behavior that third-person imagery cannot
match~\cite{li2025challenges,rodin2021predicting}. Because the data originate
from wearable devices, first-person understanding underpins augmented and
virtual reality, human--computer interaction, robotics, and assistive
technology, where accurately reading the wearer's current and future actions
enables real-time guidance---from natural AR interaction to warning a user
before they touch a dangerous object~\cite{li2025challenges,rodin2021predicting}.

Perhaps most significantly, the first-person perspective directly exposes
how people interact with nearby objects using their hands, positioning
egocentric video as a natural bridge between human demonstration and
embodied artificial intelligence \cite{plizzari2024outlook}. Recent work
treats egocentric human video as a scalable source of embodiment data: for
example, EgoMimic collects egocentric human videos paired with 3D hand
tracking and co-trains a single policy on human and robot data, achieving
strong gains on long-horizon manipulation tasks and generalizing to entirely
new scenes \cite{kareer2024egomimic}. This trajectory, from understanding
hand--object interaction toward learning embodied behavior, motivates the
present survey, which examines how vision-language models can serve as the
unifying interface that connects egocentric perception with language-based
reasoning and action. As wearable sensing hardware improves and large
multimodal models mature, first-person AI is increasingly seen as a key
bridge between perception and action \cite{li2025challenges}.

Beyond consolidating this literature, the present survey adopts a perspective
that distinguishes it from existing egocentric overviews: we treat explicit
relational structure---scene graphs, hand--object interaction graphs, and
graph-guided reasoning---as a first-class organizing principle rather than a
peripheral technique. A first-person action is defined less by appearance than
by a relation unfolding in time---which hand engages which object, and
how---and we argue throughout that graph-based, object-centric reasoning is a principled response to the well-documented failure of dense
vision-language models to capture that structure. In this sense the survey is
not merely a survey of egocentric vision-language models but a \emph{graph-aware}
survey of them, in which hand--object and scene-graph structure serves as the
connective tissue linking low-level interaction perception to language-grounded,
embodied reasoning.

\subsection{Why Vision-Language Models for Egocentric Video?}
Vision-language models learn a shared image--text representation from
large-scale paired data---CLIP, for example, replaces fixed-label supervision
with natural-language supervision to enable strong zero-shot
transfer~\cite{radford2021clip}. Building on this paradigm, EgoVLP pioneered
egocentric video-language pretraining with the EgoClip set and the
egocentric-aware EgoNCE objective~\cite{lin2022egovlp}, and EgoVLPv2 pushed
cross-modal fusion into the backbone~\cite{pramanick2023egovlpv2}. Beyond retrieval
and recognition, such models now power first-person assistants like
Vinci~\cite{huang2024vinci} and are being pushed toward explicit reasoning via
reinforcement learning for temporal grounding~\cite{xu2026temporal}.
Vision-language models thus offer a unified interface connecting egocentric
visual streams with language-based queries, recognition, and
reasoning~\cite{lin2022egovlp,radford2021clip}.

\subsection{Challenges in Egocentric Video Understanding}
Despite this progress, egocentric video understanding faces distinctive
obstacles. Data scale is a fundamental bottleneck: the largest egocentric
corpus, Ego4D, offers only $3{,}670$ hours against the billions of
image--text pairs available exocentrically, and downstream benchmarks remain
limited in scale and quality~\cite{li2025challenges}. The footage itself is
uncurated, long-form, and dominated by continuous camera motion that
complicates stable learning~\cite{grauman2022ego4d,plizzari2024outlook,li2025challenges}.
Finally, benchmarks skewed toward kitchen hand--object interaction constrain how
broadly the knowledge in vision-language models can be
evaluated~\cite{dong2025truly}. Section~\ref{sec:challenges} analyzes these
perception challenges in depth.

\subsection{Contributions of This Survey}
A distinguishing feature of this survey, relative to prior egocentric overviews,
is its \emph{graph-aware} lens: throughout, we foreground explicit relational
structure as the bridge between first-person perception and language-grounded
reasoning. The main contributions are summarized as follows:
\begin{itemize}
    \item We provide a comprehensive survey of vision-language models for
    egocentric video understanding, structured along a clear trajectory
    that progresses from hand--object interaction to embodied artificial
    intelligence.

    \item We establish the foundations of the field by formalizing
    egocentric video, its core tasks (action recognition, action
    anticipation, captioning, retrieval, and hand--object interaction
    understanding), and the perception challenges unique to the
    first-person setting, including ego-motion, occlusion, fine-grained
    manipulation, and long temporal dependencies.

    \item We review and comparatively analyze the major egocentric video
    datasets and benchmarks, and we discuss their requirements, biases, and
    limitations.

    \item We trace the evolution of video understanding models from
    classical convolutional and two-stream architectures, through
    transformer-based models, to modern vision-language models, and we
    examine why standard vision-language models underperform on egocentric
    video together with the egocentric-specific architectures designed to
    address this gap.

    \item We offer a focused treatment of hand--object interaction
    understanding, covering hand detection and tracking, object-centric and
    affordance-based interaction, and fine-grained interaction stages, as a
    central component of egocentric understanding.

    \item We survey spatiotemporal reasoning as well as graph-based and
    object-centric reasoning for egocentric video, including scene graphs,
    hand--object interaction graphs, graph-guided frame sampling, and
    prompting and semantic-alignment techniques for vision-language models.

    \item We discuss applications toward embodied intelligence, spanning
    assistive systems, human activity understanding, AR/VR and wearable AI,
    and human--robot collaboration, and we identify open challenges and
    future research directions to guide subsequent work.
\end{itemize}

\subsection{Comparison with Existing Surveys}
\label{sec:survey-comparison}
Several recent, well-cited surveys already map parts of this landscape, but each
covers a different slice of it, and none adopts the graph-aware lens that
organizes our review. The closest first-person surveys are complementary in
scope rather than competing: Rodin \textit{et al.}~\cite{rodin2021predicting}
focus on the anticipation/prediction family, Bandini and
Zariffa~\cite{bandini2023analysis} on hand analysis, Plizzari
\textit{et al.}~\cite{plizzari2024outlook} on a forward-looking ``outlook'' of
applications, and Li \textit{et al.}~\cite{li2025challenges} on a broad
subject/object/environment taxonomy of egocentric vision. Orthogonal to these,
Zhang \textit{et al.}~\cite{zhang2024vlmsurvey} survey vision-language models for
general (non-egocentric) vision, and Li \textit{et al.}~\cite{li2024sggsurvey}
survey scene-graph generation without an egocentric or embodied focus.
Table~\ref{tab:survey-comparison} contrasts their coverage against ours along the
dimensions this survey spans. The pattern it exposes is the gap we fill: the one
survey that centers relational/graph structure is not egocentric, the egocentric
surveys do not treat that structure as an organizing principle, and none unifies
egocentric vision-language modeling, hand--object interaction, graph-based
reasoning, and embodied AI within a single trajectory. Ours is, to our
knowledge, the first to do so---and the only entry checked on both the
\emph{egocentric} and \emph{graph/relational} columns together.

\begin{table*}[t]
\centering
\caption{Coverage of representative recent surveys versus this one
(Section~\ref{sec:survey-comparison}). \checkmark~= covered in depth,
$\sim$~= partial/incidental, $\times$~= not covered. ``Graph/rel.'' denotes
scene-graph, hand--object-interaction-graph, or relational structure used as an
organizing lens---the axis on which this survey is distinctive.}
\label{tab:survey-comparison}
\footnotesize
\renewcommand{\arraystretch}{1.25}
\setlength{\tabcolsep}{4pt}
\resizebox{\linewidth}{!}{%
\begin{tabular}{p{3.2cm}p{1.5cm}cccccccc}
\hline
\textbf{Survey} & \textbf{Venue/Yr} & \textbf{Ego} & \textbf{Data/tasks} & \textbf{VLMs} & \textbf{HOI} & \textbf{Spat.--temp.} & \textbf{Graph/rel.} & \textbf{Prompt/align.} & \textbf{Embodied} \\
\hline
Rodin \textit{et al.}~\cite{rodin2021predicting} & CVIU '21 & \checkmark & $\sim$ & $\times$ & $\sim$ & $\sim$ & $\times$ & $\times$ & $\sim$ \\
Bandini \& Zariffa~\cite{bandini2023analysis} & TPAMI '23 & \checkmark & $\sim$ & $\times$ & \checkmark & $\times$ & $\times$ & $\times$ & $\sim$ \\
Plizzari \textit{et al.}~\cite{plizzari2024outlook} & IJCV '24 & \checkmark & \checkmark & $\sim$ & $\sim$ & $\sim$ & $\times$ & $\times$ & \checkmark \\
Li \textit{et al.}~\cite{li2025challenges} & MIR '26 & \checkmark & \checkmark & $\sim$ & \checkmark & $\sim$ & $\times$ & $\times$ & $\sim$ \\
Zhang \textit{et al.}~\cite{zhang2024vlmsurvey} & TPAMI '24 & $\times$ & $\sim$ & \checkmark & $\times$ & $\times$ & $\times$ & \checkmark & $\times$ \\
Li \textit{et al.}~\cite{li2024sggsurvey} & Neurocomp. '24 & $\times$ & $\sim$ & $\sim$ & $\sim$ & \checkmark & \checkmark & $\times$ & $\times$ \\
\textbf{This survey} & --- & \checkmark & \checkmark & \checkmark & \checkmark & \checkmark & \checkmark & \checkmark & \checkmark \\
\hline
\end{tabular}
}
\end{table*}

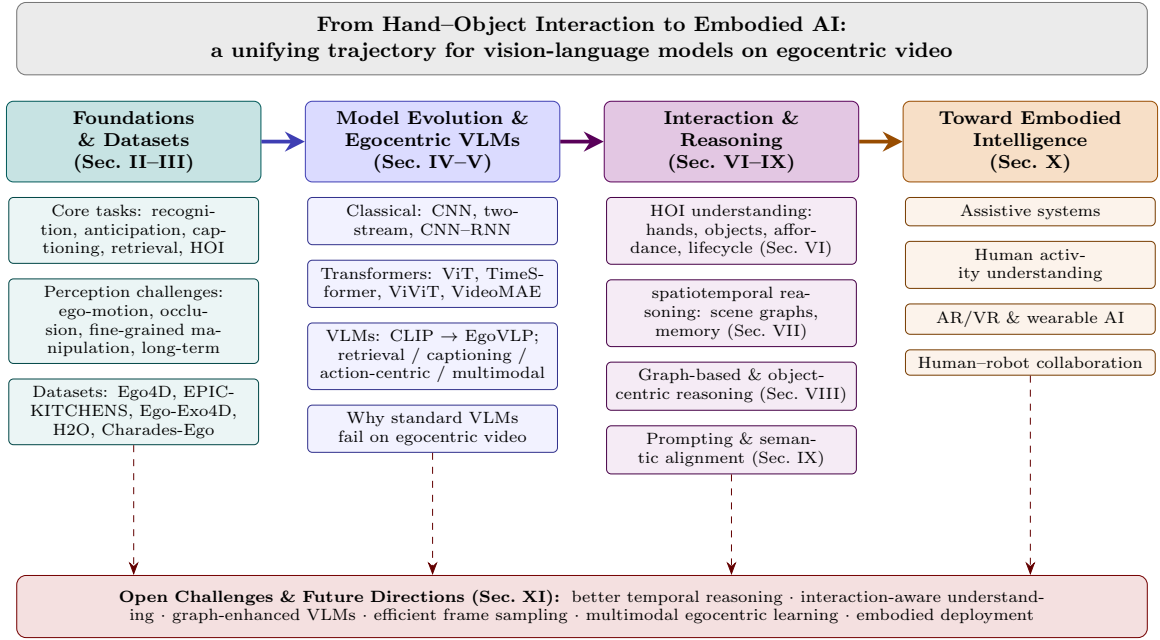
\begin{figure*}[t]
\centering
\colorlet{c1}{teal}
\colorlet{c2}{blue!65}
\colorlet{c3}{violet}
\colorlet{c4}{orange!85!black}
\colorlet{c5}{red!65!black}
\resizebox{\textwidth}{!}{%
\begin{tikzpicture}[
  font=\footnotesize,
  >={Stealth[length=3mm]},
  stage/.style={rounded corners=3pt, draw=#1!70!black, fill=#1!22,
    text width=3.55cm, align=center, inner sep=4pt, minimum height=0.95cm,
    font=\footnotesize\bfseries},
  item/.style={rounded corners=2pt, draw=#1!55!black, fill=#1!8,
    text width=3.55cm, align=center, inner sep=3pt, font=\scriptsize},
  thesis/.style={rounded corners=4pt, draw=black!55, fill=black!8,
    text width=16.6cm, align=center, inner sep=6pt, font=\small\bfseries},
  chal/.style={rounded corners=4pt, draw=c5!70!black, fill=c5!12,
    text width=16.6cm, align=center, inner sep=6pt, font=\scriptsize},
  flow/.style={-{Stealth[length=3mm,width=3mm]}, line width=1.6pt, #1!70!black},
]

\node[thesis] (thesis) at (6.75,0)
  {From Hand--Object Interaction to Embodied AI:\\
   a unifying trajectory for vision-language models on egocentric video};

\node[stage=c1] (h1) at (0,-1.55)    {Foundations \& Datasets\\(Sec.~II--III)};
\node[stage=c2] (h2) at (4.5,-1.55)  {Model Evolution \&\\Egocentric VLMs\\(Sec.~IV--V)};
\node[stage=c3] (h3) at (9.0,-1.55)  {Interaction \&\\Reasoning\\(Sec.~VI--IX)};
\node[stage=c4] (h4) at (13.5,-1.55) {Toward Embodied\\Intelligence\\(Sec.~X)};

\node[item=c1, below=2.2mm of h1] (a1)
  {Core tasks: recognition, anticipation, captioning, retrieval, HOI};
\node[item=c1, below=2.2mm of a1] (a2)
  {Perception challenges: ego-motion, occlusion, fine-grained manipulation, long-term};
\node[item=c1, below=2.2mm of a2] (a3)
  {Datasets: Ego4D, EPIC-KITCHENS, Ego-Exo4D, H2O, Charades-Ego};

\node[item=c2, below=2.2mm of h2] (b1)
  {Classical: CNN, two-stream, CNN--RNN};
\node[item=c2, below=2.2mm of b1] (b2)
  {Transformers: ViT, TimeSformer, ViViT, VideoMAE};
\node[item=c2, below=2.2mm of b2] (b3)
  {VLMs: CLIP $\to$ EgoVLP; retrieval / captioning / action-centric / multimodal};
\node[item=c2, below=2.2mm of b3] (b4)
  {Why standard VLMs fail on egocentric video};

\node[item=c3, below=2.2mm of h3] (e1)
  {HOI understanding: hands, objects, affordance, lifecycle (Sec.~VI)};
\node[item=c3, below=2.2mm of e1] (e2)
  {spatiotemporal reasoning: scene graphs, memory (Sec.~VII)};
\node[item=c3, below=2.2mm of e2] (e3)
  {Graph-based \& object-centric reasoning (Sec.~VIII)};
\node[item=c3, below=2.2mm of e3] (e4)
  {Prompting \& semantic alignment (Sec.~IX)};

\node[item=c4, below=2.2mm of h4] (d1)
  {Assistive systems};
\node[item=c4, below=2.2mm of d1] (d2)
  {Human activity understanding};
\node[item=c4, below=2.2mm of d2] (d3)
  {AR/VR \& wearable AI};
\node[item=c4, below=2.2mm of d3] (d4)
  {Human--robot collaboration};

\draw[flow=c2] (h1) -- (h2);
\draw[flow=c3] (h2) -- (h3);
\draw[flow=c4] (h3) -- (h4);

\node[chal] (chal) at (6.75,-8.55)
  {\textbf{Open Challenges \& Future Directions (Sec.~XI):}\;
   better temporal reasoning $\cdot$ interaction-aware understanding $\cdot$
   graph-enhanced VLMs $\cdot$ efficient frame sampling $\cdot$
   multimodal egocentric learning $\cdot$ embodied deployment};

\draw[dashed, c5!60!black, -{Stealth[length=2mm]}] (a3.south) -- (a3.south |- chal.north);
\draw[dashed, c5!60!black, -{Stealth[length=2mm]}] (b4.south) -- (b4.south |- chal.north);
\draw[dashed, c5!60!black, -{Stealth[length=2mm]}] (e4.south) -- (e4.south |- chal.north);
\draw[dashed, c5!60!black, -{Stealth[length=2mm]}] (d4.south) -- (d4.south |- chal.north);

\end{tikzpicture}%
}
\caption{Roadmap of this survey. The paper is organized as a single trajectory
from hand--object interaction to embodied AI. Foundations and datasets
(Sec.~\ref{sec:foundations}--\ref{sec:datasets}) supply the substrate; the model
lineage advances from classical and transformer architectures to egocentric
vision-language models (Sec.~\ref{sec:evolution}--\ref{sec:vlm}); interaction and
reasoning form the conceptual core, spanning hand--object interaction,
spatiotemporal reasoning, graph-based reasoning, and prompting and semantic
alignment (Sec.~\ref{sec:hoi-understanding}--\ref{sec:prompting-alignment}); and
these converge on applications toward embodied intelligence
(Sec.~\ref{sec:applications}). Open challenges (Sec.~\ref{sec:open-challenges})
cut across every stage, shown by the dashed feedback links.}
\label{fig:roadmap}
\end{figure*}

\subsection{Survey Methodology and Paper Selection}
\label{sec:methodology}
To keep this survey both comprehensive and representative of the current state
of the art, we followed a structured search and selection procedure for
identifying the literature reviewed in this paper. We searched the major computer-vision, machine-learning, and robotics venues---CVPR, ICCV, ECCV, NeurIPS, ICLR, ICML,
AAAI, ACM MM, ICRA, and CoRL---together with the leading journals in the field
(IEEE TPAMI, IJCV, and CVIU), and supplemented these with keyword searches over
Google Scholar, IEEE Xplore, the ACM Digital Library, DBLP, and arXiv. The
queries combined egocentric-vision terms (``egocentric,'' ``first-person,''
``hand--object interaction,'' ``Ego4D,'' ``EPIC-KITCHENS'') with
modeling terms (``vision-language model,'' ``video transformer,''
``scene graph,'' ``action anticipation,'' ``embodied learning'').

From the resulting pool we applied three inclusion criteria, in decreasing order
of priority. \emph{Relevance:} a work had to bear directly on the survey's
trajectory from hand--object interaction to embodied intelligence, or to supply a
foundational method (e.g., CLIP, ViT) on which that trajectory builds.
\emph{Venue and impact:} we preferentially selected papers published at the
top-tier venues listed above or already well cited relative to their publication
date, using citation count as a signal of community uptake rather than a hard
threshold. \emph{Recency:} we emphasized work from the last five years
(2020--2026), which captures the shift to vision-language and embodied models,
while retaining the seminal earlier papers that define the field's foundations.
Because egocentric vision is a fast-moving area in which many recent and relevant contributions first appear as preprints, we also included selected arXiv manuscripts that had not yet completed peer review. Such works are explicitly identified as preprints in the reference list and are interpreted cautiously throughout the survey. The literature search covered publications available through the end of July
2026. Works that were redundant with a stronger included reference, or that fell outside the first-person scope, were excluded.

\subsection{Organization of the Paper}
The remainder of this paper is organized as follows, and is depicted as a visual
roadmap in Fig.~\ref{fig:roadmap}.
Section~\ref{sec:foundations} presents the foundations of egocentric video
understanding, including its definition, core tasks, and characteristic
perception challenges. Section~\ref{sec:datasets} reviews major egocentric
video datasets, providing a comparative analysis and discussing their
biases and limitations. Section~\ref{sec:evolution} traces the evolution of
video understanding models from classical and transformer-based
architectures to vision-language models.
Section~\ref{sec:vlm} examines vision-language models for egocentric video
understanding, analyzing why standard models fail and surveying egocentric
architectures. Section~\ref{sec:hoi-understanding} focuses on hand--object interaction
understanding, while Section~\ref{sec:spatiotemporal} covers
spatiotemporal reasoning and Section~\ref{sec:graph} addresses
graph-based and object-centric reasoning.
Section~\ref{sec:prompting-alignment} discusses prompting and semantic alignment in
vision-language models. Section~\ref{sec:applications} presents
applications toward embodied intelligence.
Section~\ref{sec:open-challenges} outlines open challenges and future
directions, and Section~\ref{sec:conclusion} concludes the paper.

\section{Foundations of Egocentric Video Understanding}\label{sec:foundations}

\subsection{What is Egocentric Video?}
Egocentric vision, also called first-person vision, captures visual and
multimodal data through cameras or sensors worn on the human body, offering
a perspective that simulates human visual experience
\cite{li2025challenges}. The paradigm was articulated early by Kanade and
Hebert, who argued that sensing the environment and the subject's
activities from a wearable viewpoint is advantageous precisely because it
records the scene from the subject's own vantage point, in contrast to the
traditional third-person, or surveillance, perspective
\cite{kanade2012firstperson}. Because the camera is worn, its movements are driven by the wearer's intentions and attention, and the hands and
manipulated objects often occupy the central field of view, although they are frequently affected by self-occlusion and object occlusion \cite{rodin2021predicting,bandini2023analysis}.

This setting differs markedly from exocentric vision, in which the camera
is stable and disjoint from the user. The two viewpoints are complementary:
the egocentric perspective captures close-by hand--object interactions and
the wearer's attention, whereas the exocentric perspective captures full
body pose and surrounding scene context \cite{grauman2024egoexo4d}. The cost
of the first-person viewpoint is that the camera moves with the body,
causing fast motion and sudden illumination changes that can degrade video
quality \cite{bandini2023analysis}. Egocentric recordings are also typically
long, unscripted, captured during daily activities, and accompanied by
multimodal signals such as audio, eye gaze, and inertial data
\cite{li2025challenges,grauman2022ego4d}.

The modality is recorded with wearable devices mounted on the head or chest,
including action cameras such as GoPro and, more recently, purpose-built
smart glasses such as Project Aria \cite{bandini2023analysis,engel2023projectaria}.
Its growth has been driven by large-scale datasets: Ego4D offers 3{,}670
hours of daily-life video from 931 wearers across 74 locations
\cite{grauman2022ego4d}, the EPIC-KITCHENS series provides densely
annotated unscripted kitchen activity \cite{damen2022rescaling}, and
Ego-Exo4D adds simultaneously captured first- and third-person video of
skilled activities \cite{grauman2024egoexo4d}. Together, these resources
establish first-person perception as a distinct research frontier
\cite{grauman2022ego4d}.

\subsection{Core Tasks in Egocentric Video Understanding}
\label{sec:core-tasks}

Egocentric video understanding has crystallized around five canonical tasks, each inherited from third-person video analysis but reshaped by the first-person viewpoint---severe camera motion, hand occlusions, off-center objects, and the tight coupling of perception to the wearer's intent~\cite{kanade2012firstperson,plizzari2024outlook}. Table~\ref{tab:core-tasks} defines each task alongside its defining first-person difficulty. Nearly all of them are anchored by two benchmarks: EPIC-KITCHENS-100, with 89{,}977 fine-grained (verb, noun) action segments over 100 hours in 45 kitchens, and Ego4D, with 3{,}670 hours of daily-life video across 74 locations~\cite{damen2022rescaling,grauman2022ego4d}.

\begin{table}[t]
\centering
\caption{The five canonical egocentric video-understanding tasks
(Section~\ref{sec:core-tasks}): what each is and its defining first-person
difficulty.}
\label{tab:core-tasks}
\footnotesize
\renewcommand{\arraystretch}{1.3}
\begin{tabular}{p{1.8cm}p{2.7cm}p{2.5cm}}
\hline
\textbf{Task} & \textbf{What it is} & \textbf{Key difficulty} \\
\hline
Action recognition~\cite{damen2022rescaling} & Classify a trimmed clip as a (verb, noun) action & Long-tailed compositions; blur; small active objects \\
Action anticipation~\cite{furnari2020rulstm,girdhar2021avt} & Predict future action(s) before they occur & Inherent uncertainty; many plausible futures \\
Video captioning~\cite{krishna2017densecap,yang2023vid2seq,zhao2023lavila} & Generate language narrations of events & Grounding fine HOI; long-video coherence \\
Video retrieval~\cite{wray2021semantic,bain2021frozen} & Cross-modal text$\leftrightarrow$video matching & Many-to-many relevance; long-horizon queries \\
HOI understanding~\cite{shan2020understanding,cheng2023hands23} & Detect hands/objects, contact, and state change & Interaction (verb) vs.\ mere object recognition \\
\hline
\end{tabular}
\end{table}

Methodologically these tasks track the broader shift from convolutional and recurrent models to video transformers and, most recently, egocentric video-language pretraining such as EgoVLP~\cite{lin2022egovlp}, developed in Section~\ref{sec:evolution}. Hand--object interaction understanding is the most distinctively egocentric of the five, since the wearer's hands dominate the frame~\cite{bandini2023analysis}, and we treat it in depth in Section~\ref{sec:hoi-understanding}. A single theme unites all five: current models name objects well but falter on the interaction and long-horizon structure that define first-person activity---EgoVLMs collapse under single verb substitutions~\cite{dong2025truly}, and billion-parameter models still trail humans badly on long-form question answering~\cite{mangalam2023egoschema}. That gap motivates the interaction- and graph-centric treatment of the rest of this survey.

\subsection{Challenges of Egocentric Perception}
\label{sec:challenges}

Egocentric video compounds the difficulties of generic video with challenges intrinsic to the first-person viewpoint---an unconstrained head-mounted camera framing a narrow, shifting slice of the world dominated by close-range interaction~\cite{plizzari2024outlook,li2025challenges,kanade2012firstperson}. Four stand out.

\subsubsection{Ego-motion}
\label{sec:egomotion}
The camera is rigidly coupled to the wearer's head, so head rotations and gaze shifts induce large, unpredictable motion and blur with no analogue in the stabilized footage on which standard backbones are trained, degrading clip features and breaking trackers~\cite{plizzari2024outlook,li2025challenges}. Mitigations either suppress it---hand-focused stabilization yields a $+17\%$ (and $+22\%$ combined) Top-1 gain on Ego-Exo4D keystep recognition~\cite{chavis2025dexfocus}---or exploit it as signal, reading head trajectory from IMU and visual--inertial SLAM as a cue to attention and 6-DoF pose~\cite{tan2023egodistill,engel2023projectaria}.

\subsubsection{Occlusion and Partial Visibility}
\label{sec:occlusion}
Acting hands and held objects persistently occlude the most informative region, while the narrow field of view truncates objects and lets them leave and re-enter view as attention shifts~\cite{darkhalil2022visor,tang2023egotracks}. This breaks trackers tuned on framed footage---EgoTracks scores far below conventional benchmarks because objects vanish and must be re-detected over minutes---and motivates occlusion-robust, amodal, high-capacity models such as HaMeR~\cite{tang2023egotracks,pavlakos2024hamer}.

\subsubsection{Fine-grained Manipulation}
\label{sec:finegrained}
Daily activity is dominated by subtle near-field manipulations---``take'' vs.\ ``put,'' or one small object among many---reflected in EPIC-KITCHENS-100's compositional space of 97 verbs and 300 nouns~\cite{damen2022rescaling}. Because the discriminative evidence lies in brief contact moments, global-context models confuse fine classes and even large EgoVLMs are flipped by single verb or noun swaps~\cite{dong2025truly}; progress leans on contact-rich datasets such as HOI4D, Assembly101, and ARCTIC~\cite{liu2022hoi4d,sener2022assembly101,fan2023arctic}.

\subsubsection{Long Temporal Dependencies}
\label{sec:longterm}
Egocentric recordings run for minutes to hours, with steps that derive meaning only from the whole activity---far beyond the few-second clips standard backbones ingest~\cite{grauman2022ego4d,plizzari2024outlook}. In the original EgoSchema evaluation, the tested billion-parameter models scored below $33\%$, whereas humans reached approximately $76\%$~\cite{mangalam2023egoschema}. This gap has motivated memory-augmented transformers such as MeMViT and MC-ViT, although many systems still rely on appearance shortcuts rather than genuine temporal reasoning~\cite{xu2026temporal,wu2022memvit,balazevic2024mcvit}.

\section{Egocentric Video Datasets}\label{sec:datasets}
The progress of vision-language models for first-person understanding is
inseparable from the datasets that train and evaluate them. Egocentric
corpora differ markedly from the third-person collections that have driven
mainstream video understanding, both in the visual phenomena they capture and in the annotation effort they demand \cite{plizzari2024outlook}. This section first articulates what egocentric understanding requires of a dataset (Section~\ref{subsec:requirements}), then surveys five widely used datasets that span the design space (Section~\ref{subsec:major}), compares them along dimensions relevant to vision-language modeling (Section~\ref{subsec:comparison}), and finally examines the biases and limitations that constrain how far models trained on them can generalize (Section~\ref{subsec:biases}).
 
\subsection{Dataset Requirements for Egocentric Understanding}
\label{subsec:requirements}
First-person, or first-person-vision (FPV), data imposes requirements that
are largely absent from third-person video. Because the camera is worn, it
records exactly what is in front of the user, its motion is driven by the
wearer's own activity and attention, and the hands together with the objects they manipulate tend to occupy the center of the field of
view \cite{bandini2023analysis}. This intimate coupling between sensor and
behavior is precisely what makes the first-person perspective valuable: it
captures the environment and the subject's activities from a vantage point
that is more informative about intent and interaction than an external,
surveillance-style view \cite{kanade2012firstperson}.
 
Several properties follow from this viewpoint and define what an egocentric dataset must capture. First, the imagery is dominated by fast, unconstrained head and camera motion, severe occlusion, and background clutter, none of which are typical of curated third-person footage \cite{kwon2021h2o}. Second, because action from a first-person viewpoint is largely a matter of understanding how the hands engage objects, hand--object interaction is the central signal rather than a peripheral one \cite{kwon2021h2o, bandini2023analysis}. Third, everyday behavior unfolds as long, unscripted activity, so datasets must preserve long-term temporal structure rather than isolated, trimmed clips \cite{damen2022rescaling}. Fourth, the first-person
setting naturally affords rich multimodality---inertial measurement units
(IMU), eye gaze, audio, stereo, and three-dimensional scene
geometry---signals that are difficult to obtain from third-person video and that have been shown to be important for tasks such as anticipating future activity \cite{rodin2021predicting}. Finally, always-on wearable capture raises privacy and consent concerns that responsible datasets must
address \cite{plizzari2024outlook}. Recent surveys organize the resulting
research landscape around the understanding of the subject, the objects, the environment, and their interactions \cite{li2025challenges}, and they
emphasize that the field is still far from the always-on, personalized,
life-enhancing wearable systems that motivate it \cite{plizzari2024outlook}.
 
\subsection{Major Egocentric Datasets}
\label{subsec:major}
We review five datasets that together cover the principal axes of egocentric data design: from constrained three-dimensional hand-object capture to massive in-the-wild collection, and from ego-only recording to simultaneous first- and third-person views.
 
\subsubsection{EPIC-KITCHENS}
EPIC-KITCHENS-100~\cite{damen2022rescaling} is the largest unscripted dataset of daily kitchen activity: 100 hours from 37 participants across 45 kitchens, densely narrated into 89{,}977 (verb, noun) action segments (97 verbs, 300 nouns; 4{,}025 action classes). Its suite of recognition, anticipation, retrieval, and domain-adaptation challenges made it the standard egocentric testbed, and the VISOR extension~\cite{darkhalil2022visor} adds 272K pixel-level hand/object masks with hand--object relations.

\subsubsection{Ego4D}
Ego4D~\cite{grauman2022ego4d} is the field's scale anchor: 3{,}670 hours of
daily-life video from 931 wearers across 74 locations in nine countries.
Portions of the collection are accompanied by audio, eye gaze, stereo,
three-dimensional meshes, and/or synchronized recordings from multiple
egocentric cameras. The dataset defines five benchmark suites (Episodic Memory, Hands and Objects, Audio-Visual Diarization, Social, Forecasting). Its dense free-form narrations made it the de facto backbone for egocentric video-language pretraining; the derived EgoSchema benchmark~\cite{mangalam2023egoschema} exposes the
long-form reasoning gap: in its original evaluation, the tested large models scored below $33\%$, whereas humans reached approximately $76\%$.

\subsubsection{H2O}
H2O~\cite{kwon2021h2o} instead targets fine-grained 3D understanding of two hands manipulating objects: 571{,}645 synchronized multi-view RGB-D frames (4 subjects, 36 actions, 8 objects) with dense 3D poses of both hands, 6DoF object poses, meshes, and point clouds. It is a key resource for 3D two-hand--object grounding, despite its small subject count.

\subsubsection{Charades-Ego}
Charades-Ego~\cite{sigurdsson2018charadesego} is defined by paired first- and third-person capture of the same scripted activities: 68{,}536 instances over 68.8 hours, 112 actors, and the 157 Charades activity classes. The paired recordings make it an early testbed for the ego-exo correspondence problem.

\subsubsection{Ego-Exo4D}
Ego-Exo4D~\cite{grauman2024egoexo4d} couples first- and third-person capture at scale for skilled activity: 1{,}286 hours from 740 participants across 13 cities, in 5{,}035 takes spanning physical (sports, dance, music) and procedural (cooking, repair) domains. Each take pairs Project Aria glasses (RGB, gaze, IMU, audio) with four to five calibrated static cameras, and it adds expert-commentary language and benchmarks for keystep recognition, proficiency estimation, ego-exo correspondence, and 3D body/hand pose.

\subsection{Comparative Dataset Analysis}
\label{subsec:comparison}
The five datasets occupy complementary positions in the design space, as
summarized in Table~\ref{tab:dataset-comparison}. They differ by roughly an
order of magnitude or more in scale, from the 100 hours of EPIC-KITCHENS-100
to the 3{,}670 hours of Ego4D, and along a spectrum of naturalism
that runs from the constrained laboratory capture of H2O, through the scripted
activities of Charades-Ego, to the fully unscripted recording of EPIC-KITCHENS,
Ego4D, and Ego-Exo4D. They also differ in viewpoint: EPIC-KITCHENS and Ego4D
are ego-only, H2O augments the egocentric view with static cameras, and
Charades-Ego and Ego-Exo4D provide paired and simultaneous first- and
third-person views, respectively. Annotation richness varies in kind as well
as degree: H2O offers dense three-dimensional hand and object pose, Ego-Exo4D
adds three-dimensional body and hand pose together with point clouds, whereas
EPIC-KITCHENS and Charades-Ego remain largely two-dimensional.
 
These differences map directly onto the downstream tasks each dataset can
support for vision-language models. Language-grounded pretraining and
video-text retrieval are best served by EPIC-KITCHENS-100, Ego4D, and
Ego-Exo4D, whose free-form narrations provide aligned textual supervision;
long-form video question answering is probed by EgoSchema; three-dimensional
hand-object grounding relies on H2O and related datasets; and ego-exo transfer
and skill or proficiency estimation are uniquely supported by Ego-Exo4D and
Charades-Ego. This complementarity has motivated work on hand-object
interaction datasets that trade scale for annotation density, such as
Assembly101 \cite{sener2022assembly101}, ARCTIC \cite{fan2023arctic},
HOI4D \cite{liu2022hoi4d}, and HOT3D \cite{banerjee2024hot3d}, which provide
multi-view and three-dimensional supervision in more controlled settings.
 
\begin{table*}[t]
\centering
\caption{Comparison of major egocentric video datasets along dimensions
relevant to vision-language modeling. Ego4D scale refers to the full video collection, while annotations and auxiliary modalities cover subsets; H2O is reported in frames. Figures are version-dependent for Ego4D and Ego-Exo4D (see Section~\ref{subsec:biases}).}
\label{tab:dataset-comparison}
\small
\resizebox{\linewidth}{!}{%
\begin{tabular}{lccccc}
\hline
\textbf{Dimension} & \textbf{EPIC-KITCHENS-100} & \textbf{Ego4D} & \textbf{H2O} & \textbf{Charades-Ego} & \textbf{Ego-Exo4D} \\
\hline
Scale & 100\,h & 3{,}670\,h & 571{,}645 frames & 68.8\,h & 1{,}286\,h \\
Participants & 37 & 931 & 4 & 112 & 740 \\
Geographic spread & 4 cities & 9 countries & 3 envs.\ (lab) & worldwide homes & 13 cities \\
Naturalism & unscripted & unscripted & lab & scripted & unscripted \\
Viewpoint & ego & ego & ego + static & paired ego-exo & simultaneous ego-exo \\
Domain & kitchen & daily life & tabletop HOI & household & skilled activity \\
3D annotation & 2D (masks) & partial & rich 3D & 2D & rich 3D \\
Language supervision & strong & strong & weak & moderate & strong \\
\hline
\end{tabular}
}
\end{table*}
 
\subsection{Dataset Biases and Limitations}
\label{subsec:biases}
Despite their scale, egocentric datasets carry biases that constrain the
generalization of models trained on them. The most pervasive is geographic and
domain narrowness. EPIC-KITCHENS is confined to kitchen activity recorded in
only four Western cities, and even Ego4D, the most geographically diverse,
acknowledges that its 74 locations fall well short of global coverage, that
camera wearers are concentrated in urban and college-town areas, and that
pandemic-era collection skewed its footage toward stay-at-home
scenarios \cite{grauman2022ego4d}. These limitations are a concrete instance of
the dataset-bias phenomenon long recognized in computer vision, whereby
collections that set out to represent the visual world instead become closed
worlds that generalize poorly to one another \cite{torralba2011unbiased}.
 
A second source of bias lies in the annotations themselves. Free-form
narrations are authored by participants or crowd workers and are
overwhelmingly English-centric, so the very language signal used to align
video and text inherits lexical noise and culturally specific word choices;
Ego4D explicitly notes that its narrations, produced by annotators at two
sites, are biased toward local word usage \cite{grauman2022ego4d}. Demographic
skew compounds these effects, as does the small number of subjects in
densely annotated three-dimensional datasets such as H2O, which limits the
diversity of hand morphologies and manipulation styles represented. Finally,
always-on first-person capture raises privacy and consent concerns; datasets
mitigate these through informed consent and de-identification, or by recording
in controlled environments \cite{plizzari2024outlook}.
 
For vision-language models, the practical consequence is that systems
pretrained on Western, kitchen-heavy, English-narrated egocentric data risk
degraded performance when transferred to other cultures, languages, viewpoints,
and activity domains. Mitigations include explicit cross-domain and
domain-adaptation evaluation, balanced multi-domain pretraining mixtures, and
transparent reporting of narration provenance. Standardized capture platforms
such as Project Aria \cite{engel2023projectaria} help unify modalities across
newer datasets, but by concentrating data collection on common hardware they
may also introduce shared, sensor-induced biases that future work will need to
characterize.

\section{Evolution of Video Understanding Models}
\label{sec:evolution}

\begin{figure*}[t]
\centering
\colorlet{e1}{teal}
\colorlet{e2}{blue!65}
\colorlet{e3}{violet}
\colorlet{e4}{red!65!black}
\resizebox{\textwidth}{!}{%
\begin{tikzpicture}[
  font=\footnotesize,
  >={Stealth[length=3mm]},
  era/.style={rounded corners=3pt, draw=#1!70!black, fill=#1!22,
    text width=5.0cm, align=center, inner sep=4pt, minimum height=1.0cm,
    font=\footnotesize\bfseries},
  fam/.style={rounded corners=2pt, draw=#1!55!black, fill=#1!8,
    text width=5.0cm, align=center, inner sep=3pt, font=\scriptsize},
  gap/.style={rounded corners=2pt, draw=e4!65!black, fill=e4!10,
    text width=5.0cm, align=center, inner sep=3pt, font=\scriptsize\itshape},
  axis/.style={rounded corners=4pt, draw=black!55, fill=black!8,
    text width=17.0cm, align=center, inner sep=5pt, font=\small\bfseries},
  base/.style={rounded corners=4pt, draw=e1!70!black, fill=e1!12,
    text width=17.0cm, align=center, inner sep=5pt, font=\scriptsize},
  flow/.style={-{Stealth[length=3mm,width=3mm]}, line width=1.6pt, #1!70!black},
  step/.style={-{Stealth[length=2mm]}, draw=#1!60!black, line width=0.8pt},
]

\node[axis] (axis) at (6.0,0.35)
  {{Evolution of representation and supervision: closed-set visual models
 $\longrightarrow$ open-vocabulary vision-language models};};

\node[era=e1] (h1) at (0,-1.35)
  {\textbf{I. Classical} (2014--2019)\\Hand-engineered temporal modeling};
\node[era=e2] (h2) at (6.0,-1.35)
  {\textbf{II. Transformers} (2020--2023)\\Learned space-time attention};
\node[era=e3] (h3) at (12.0,-1.35)
  {\textbf{III. Vision-Language} (2021--)\\Language-supervised representation};

\node[fam=e1, below=2.4mm of h1] (a1)
  {\textbf{CNN / 3D-CNN}\\Sports-1M $\to$ C3D $\to$ P3D, R(2+1)D $\to$ I3D;
   SlowFast, X3D, TSM\\\emph{spatiotemporal convolutions; inflate 2D priors}};
\node[fam=e1, below=2.4mm of a1] (a2)
  {\textbf{CNN--RNN}\\LRCN, Ng \textit{et al.}, ConvLSTM, VideoLSTM; LSTA
   (first-person)\\\emph{variable-length recurrent aggregation}};
\node[fam=e1, below=2.4mm of a2] (a3)
  {\textbf{Two-stream}\\Two-Stream $\to$ TSN, TRN, ActionVLAD $\to$
   EPIC-Fusion (+audio)\\\emph{parallel appearance + motion pathways}};
\node[gap, below=2.4mm of a3] (g1)
  {Gap: fixed short clips miss long horizons; optical flow costly and brittle
   under ego-motion; closed label space};

\node[fam=e2, below=2.4mm of h2] (b1)
  {\textbf{ViT} (+ DeiT, Swin)\\patch tokenization; no inductive bias
   \emph{but data-hungry}};
\node[fam=e2, below=2.4mm of b1] (b2)
  {\textbf{TimeSformer} $\;\parallel\;$ \textbf{ViViT}\\divided / factorized
   space-time attention\\\emph{convergent solution to quadratic cost}};
\node[fam=e2, below=2.4mm of b2] (b3)
  {\textbf{VideoMAE} (ST-MAE, V2)\\tube masking at 90--95\%\\\emph{makes video
   transformers data-efficient}};
\node[fam=e2, below=2.4mm of b3] (b4)
  {\textbf{Breadth:} MViT/MViTv2, Video Swin, Motionformer, MaskFeat,
   UniFormer, MTV, MeMViT \emph{(memory, minutes-scale)}};
\node[gap, below=2.4mm of b4] (g2)
  {Gap: representations remain visual-only and lack an explicit natural-language interface};

\node[fam=e3, below=2.4mm of h3] (c1)
  {\textbf{CLIP}\\image--text contrastive, 400M pairs;
   76.2\% zero-shot ImageNet};
\node[fam=e3, below=2.4mm of c1] (c2)
  {\textbf{VideoCLIP}\\video--text contrastive: overlapping positives +
   retrieval-mined hard negatives};
\node[fam=e3, below=2.4mm of c2] (c3)
  {\textbf{Frozen-in-Time}\\joint image + video encoder, \emph{ViT\,/\,%
   TimeSformer-style};\\\emph{direct ancestor of EgoVLP}};
\node[fam=e3, below=2.4mm of c3] (c4)
  {\textbf{EgoVLP} (EgoVLPv2, LaViLa)\\EgoClip ($\sim$3.8M) + EgoNCE +
   EgoMCQ\\\emph{first-person transfer across five tasks}};
\node[gap, below=2.4mm of c4] (g3)
  {Open vocabulary achieved --- but verb/interaction and long-horizon
   understanding remain open (Sec.~\ref{sec:vlm-fail})};

\draw[flow=e2] (h1) -- (h2);
\draw[flow=e3] (h2) -- (h3);

\foreach \s/\t in {a1/a2, a2/a3} \draw[step=e1] (\s.south) -- (\t.north);
\foreach \s/\t in {b1/b2, b2/b3, b3/b4} \draw[step=e2] (\s.south) -- (\t.north);
\foreach \s/\t in {c1/c2, c2/c3, c3/c4} \draw[step=e3] (\s.south) -- (\t.north);

\draw[dashed, e2!60!black, -{Stealth[length=2mm]}] (a1.east) -- (b1.west);
\draw[dashed, e3!60!black, -{Stealth[length=2mm]}] (b4.east) -- (c4.west);

\node[base] (base) at (6.0,-9.55)
  {\textbf{Constant egocentric substrate:} EPIC-KITCHENS-100 and Ego4D supply
   the backbones, baselines, and benchmarks at every stage --- VideoMAE anchors
   most Ego4D challenge entries, and EgoVLP's EgoClip is curated from Ego4D};
\foreach \n in {g1, g2, g3}
  \draw[dashed, e1!60!black, -{Stealth[length=2mm]}]
    (\n.south) -- (\n.south |- base.north);

\end{tikzpicture}%
}
\caption{Evolution of video understanding models
(Section~\ref{sec:evolution}). Three eras, read left to right, each inheriting representational machinery from the preceding era while
changing its temporal modeling and supervision strategy.
\emph{Classical} models hand-engineer temporal structure through
spatiotemporal convolutions, recurrent aggregation, and parallel
appearance--motion streams (Section~\ref{sec:classical}).
\emph{Transformer} models replace those inductive biases with factorized
space-time attention, made trainable at video scale by masked pretraining
(Section~\ref{sec:transformer}). \emph{Vision-language} models replace the
fixed label space itself with natural-language supervision, a lineage running
CLIP $\to$ VideoCLIP $\to$ Frozen-in-Time $\to$ EgoVLP. Red boxes mark the gap each era left open and the
next set out to close; the horizontal dashed links mark what each era
inherited rather than replaced---image pretraining and convolutional backbones
carried into the transformer era, and transformer space-time encoders carried
into egocentric video-language pretraining.}
\label{fig:evolution}
\end{figure*}
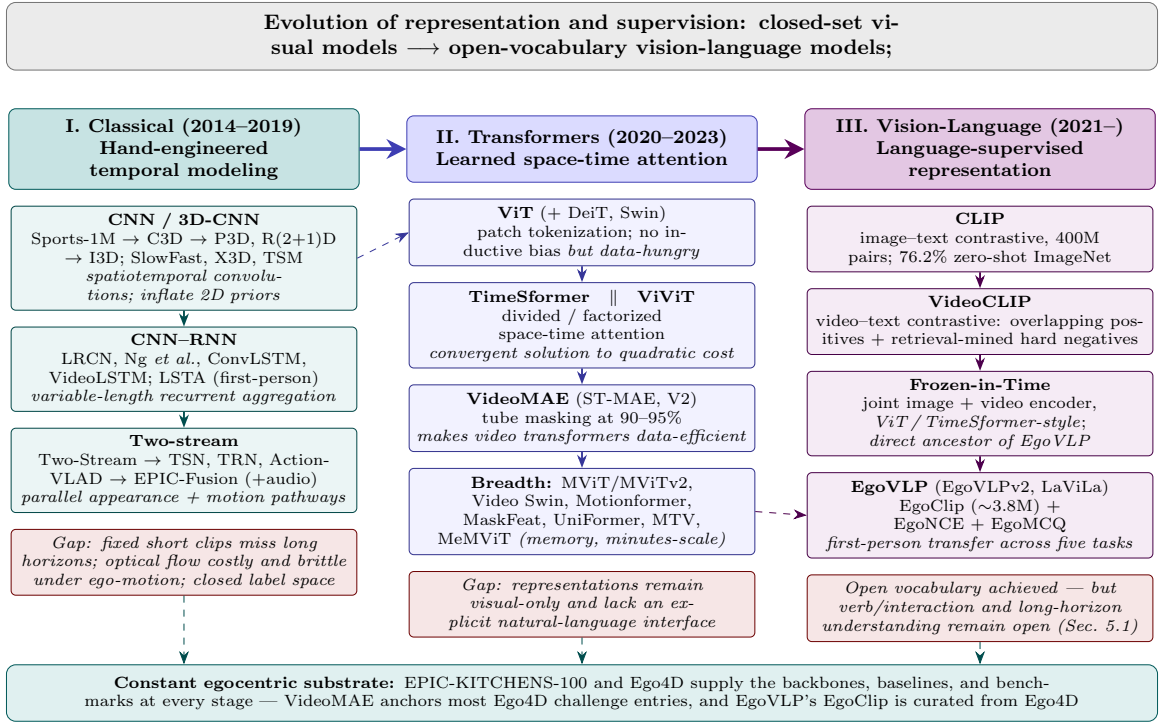

Modern egocentric video understanding inherits its core machinery from a
decade of progress on third-person action recognition. Before
transformer-based vision--language models reshaped the field, three families
of deep architectures defined the state of the art: pure convolutional
networks operating directly on stacks of frames, hybrid CNN--RNN pipelines
that delegated temporal modeling to recurrent units, and two-stream networks
that processed appearance and motion as parallel pathways. We review these
classical models because the egocentric literature, from EPIC-KITCHENS
\cite{damen2022rescaling} to Ego4D \cite{grauman2022ego4d}, still relies on
them as backbones, baselines, and conceptual scaffolding for the
hand--object reasoning that motivates this survey. Fig.~\ref{fig:evolution}
summarizes the trajectory this section traces, from hand-engineered temporal
modeling to language-supervised, open-vocabulary representation.

\subsection{Classical Video Understanding}
\label{sec:classical}

Three pre-transformer families defined video action recognition. Each solved
one bottleneck and left another open; Table~\ref{tab:classical} summarizes the
progression.

\subsubsection{CNN-Based Models}
\label{sec:cnn}
Convolutional models learn motion directly from pixels. After Karpathy
\textit{et al.}~\cite{karpathy2014largescale} showed that naive frame fusion
barely beats single-frame baselines, C3D~\cite{tran2015c3d} demonstrated that
homogeneous $3{\times}3{\times}3$ convolutions yield compact, transferable
spatiotemporal descriptors; P3D~\cite{qiu2017p3d} and
R(2+1)D~\cite{tran2018r2plus1d} then factorized these costly 3D kernels into
cheaper spatial and temporal parts, and I3D~\cite{carreira2017i3d} inflated 2D
ImageNet filters into 3D to inherit mature image pretraining. Later variants chased efficiency and range---the
dual-rate pathways of SlowFast~\cite{feichtenhofer2019slowfast}, the scaled
tiny backbone of X3D~\cite{feichtenhofer2020x3d}, the zero-cost channel shift
of TSM~\cite{lin2019tsm}, and non-local blocks for long-range
context~\cite{wang2018nonlocal}. \emph{Left unsolved:} fixed-length clips still
cannot span the minutes-long, goal-directed structure of egocentric activity.

\subsubsection{CNN--RNN Architectures}
\label{sec:cnnrnn}
A second family kept a per-frame CNN and delegated time to a recurrent network,
modeling variable-length structure that fixed clips could not. LRCN~\cite{donahue2015lrcn}
and the pooling/LSTM study of Ng~\textit{et al.}~\cite{ng2015beyond} aggregated
per-frame features over longer spans; ConvLSTM~\cite{shi2015convlstm} and
VideoLSTM~\cite{li2018videolstm} restored spatial locality inside the recurrent
cell; and encoder--decoder LSTMs previewed self-supervised
pretraining~\cite{srivastava2015unsupervised}. This family fit first-person
video naturally, where hands and gaze must be tracked over time: dedicated
first-person cue encoders~\cite{singh2016fpv}, hand- and object-supervised twin
streams~\cite{ma2016deeperfpv}, and the attention unit
LSTA~\cite{sudhakaran2019lsta} reached the state of the art on EGTEA and
EPIC-KITCHENS~\cite{damen2022rescaling}. \emph{Left unsolved:} recurrence
captured order but not the appearance--motion split that separates visually
similar first-person actions.

\subsubsection{Two-Stream Networks}
\label{sec:twostream}
The third family processed appearance and motion in parallel: an RGB stream and
a stacked-optical-flow stream fused late~\cite{simonyan2014twostream}, later
refined by better fusion placement and residual
cross-connections~\cite{feichtenhofer2016fusion,feichtenhofer2016stresnet},
sparse segment sampling in TSN~\cite{wang2016tsn}, multi-scale temporal
relations in TRN~\cite{zhou2018trn}, and learnable aggregation in
ActionVLAD~\cite{girdhar2017actionvlad}. It transferred well to first-person
video, where rapid hand and head motion is informative, culminating in
EPIC-Fusion~\cite{kazakos2019epicfusion}, which added audio to set the state of
the art on EPIC-KITCHENS. \emph{Left unsolved:} precomputed optical flow is
expensive and brittle under the severe ego-motion of wearable capture.

\begin{table*}[t]
\centering
\caption{The three classical (pre-transformer) video-understanding families:
what each generation contributed and the gap it left for egocentric
vision-language understanding (Section~\ref{sec:classical}).}
\label{tab:classical}
\footnotesize
\renewcommand{\arraystretch}{1.3}
\resizebox{\linewidth}{!}{%
\begin{tabular}{p{1.7cm}p{3.0cm}p{3.3cm}p{3.2cm}p{3.7cm}}
\hline
\textbf{Generation} & \textbf{Representative models} & \textbf{Key innovation} & \textbf{Problem solved} & \textbf{Left unsolved for egocentric} \\
\hline
CNN / 3D-CNN & C3D~\cite{tran2015c3d}, I3D~\cite{carreira2017i3d}, SlowFast~\cite{feichtenhofer2019slowfast}, TSM~\cite{lin2019tsm} & Spatiotemporal convolutions; inflate 2D pretraining into 3D & Learns motion from pixels with mature image priors & Fixed short clips miss long-horizon, goal-directed activity \\
CNN--RNN & LRCN~\cite{donahue2015lrcn}, ConvLSTM~\cite{shi2015convlstm}, LSTA~\cite{sudhakaran2019lsta} & Per-frame CNN + recurrent temporal aggregation & Variable-length modeling; tracks hands/gaze over time & Weak appearance--motion separation; closed vocabulary \\
Two-stream & Two-Stream~\cite{simonyan2014twostream}, TSN~\cite{wang2016tsn}, EPIC-Fusion~\cite{kazakos2019epicfusion} & Parallel appearance + motion (and audio) pathways & Explicit motion and multimodal cues for fine actions & Optical flow costly/brittle under ego-motion; no language grounding \\
\hline
\end{tabular}
}
\end{table*}

Taken together, these families established the representational
vocabulary---spatiotemporal convolutions, recurrent temporal aggregation, and
parallel appearance--motion streams---inherited by the transformer-based and
vision--language models examined next. Their common limitation, a fixed label
space with no language interface, is exactly what the vision-language models of
Section~\ref{sec:vlm} set out to remove.

\subsection{Transformer-Based Video Understanding}
\label{sec:transformer}

Attention-based architectures reshaped video understanding in a few years. The lineage is direct: the Vision Transformer (ViT) establishes patch tokenization, TimeSformer and ViViT carry self-attention to the temporal axis through \emph{factorized} schemes, and VideoMAE makes these data-hungry video transformers trainable via self-supervised masking. We take these four as the pillars and connect each to the egocentric setting.

\subsubsection{Vision Transformer (ViT)}
\label{sec:vit}
ViT~\cite{dosovitskiy2021vit} imports the self-attention primitive of Vaswani~\textit{et al.}~\cite{vaswani2017attention} into vision as the first pure-transformer image classifier: an image becomes a sequence of linearly embedded patches processed by a standard Transformer encoder. Its thesis---that an inductive-bias-free design matches CNNs given enough data---also names its weakness for egocentric corpora, which are far smaller than web scale. Two extensions relax that data demand: DeiT~\cite{touvron2021deit} trains on ImageNet-1k alone via heavy augmentation and a distillation token, while Swin~\cite{liu2021swin} adds hierarchical, shifted-window local attention with linear complexity, seeding the later Video Swin.

\subsubsection{TimeSformer}
\label{sec:timesformer}
TimeSformer~\cite{bertasius2021timesformer} applies attention directly to frame patches and compares five space-time schemes, finding that \emph{divided space-time attention}---temporal then spatial within each block---gives the best accuracy/efficiency trade-off by decoupling the quadratic cost of joint attention. The distinction is egocentric-relevant: space-only attention suffices on appearance-dominated Kinetics but the temporal component is essential on motion-centric data such as Something-Something and EPIC-KITCHENS, exactly the hand--object regime.

\subsubsection{VideoMAE}
\label{sec:videomae}
Masked modeling attacks the data appetite directly. Following image MAE~\cite{he2022mae}, VideoMAE~\cite{tong2022videomae} masks video with a \emph{tube} strategy at an extreme $90$--$95\%$ ratio---enabled by temporal redundancy---and trains strongly even on a few thousand videos without extra data, which is why its backbones dominate Ego4D and EPIC-KITCHENS leaderboards. The concurrent ST-MAE~\cite{feichtenhofer2022stmae} uses \emph{random} spacetime masking (the key contrast), and VideoMAE~V2~\cite{wang2023videomaev2} scales the paradigm to a billion-parameter video foundation model via dual masking.

\subsubsection{ViViT}
\label{sec:vivit}
ViViT~\cite{arnab2021vivit} tokenizes a clip (optionally via tubelet embedding) and proposes four factorizations that trade cost for accuracy; its factorized self-attention variant coincides with TimeSformer's divided attention, showing the two core video transformers converged on the same principle, while its factorized encoder is often the best compute trade-off. It, too, reports EPIC-KITCHENS results, tying the design to the egocentric theme.

\subsubsection{Architectural Breadth and Egocentric Applications}
\label{sec:transformer_breadth}
Beyond the four pillars, a cluster of variants advanced complementary axes: the multiscale MViT/\allowbreak MViTv2 family~\cite{fan2021mvit,li2022mvitv2}, Video Swin's 3D shifted windows~\cite{liu2022videoswin}, Motionformer's trajectory attention along motion paths~\cite{patrick2021motionformer}, feature-prediction pretraining in MaskFeat~\cite{wei2022maskfeat}, the convolution--attention hybrid UniFormer~\cite{li2023uniformer}, and efficiency-oriented MTV~\cite{yan2022mtv}, TokenLearner~\cite{ryoo2021tokenlearner}, and the memory-augmented MeMViT~\cite{wu2022memvit} for minutes-scale recognition---the last especially suited to long egocentric streams.

These backbones recur throughout egocentric video-language pretraining: EgoVLP~\cite{lin2022egovlp} and EgoVLPv2 \cite{pramanick2023egovlpv2} build on TimeSformer-style space-time encoders, LaViLa~\cite{zhao2023lavila} learns a contrastive embedding on the same backbone from LLM-generated narrations, and VideoMAE(V2) anchors most Ego4D challenge entries---closing the loop between general transformer architectures and egocentric practice.

\subsection{From Vision Models to Vision-Language Models}
\label{sec:vlm-transition}

The architectures of Section~\ref{sec:transformer} learn visual representations
through supervised or self-supervised objectives, but they do not provide an
explicit natural-language interface. Vision-language models instead align visual
encoders with natural language, yielding open-vocabulary representations that
can transfer to unseen tasks without task-specific labels. This subsection follows that progression as a genealogy: CLIP establishes image-text contrastive pretraining, VideoCLIP extends contrastive pretraining to video, Frozen-in-Time introduces a space-time encoder that jointly handles images and video, and EgoVLP transfers this design to the egocentric domain.

\subsubsection{CLIP}
\label{sec:clip}

\begin{figure*}[t]
\centering
\includegraphics[width=0.9\textwidth]{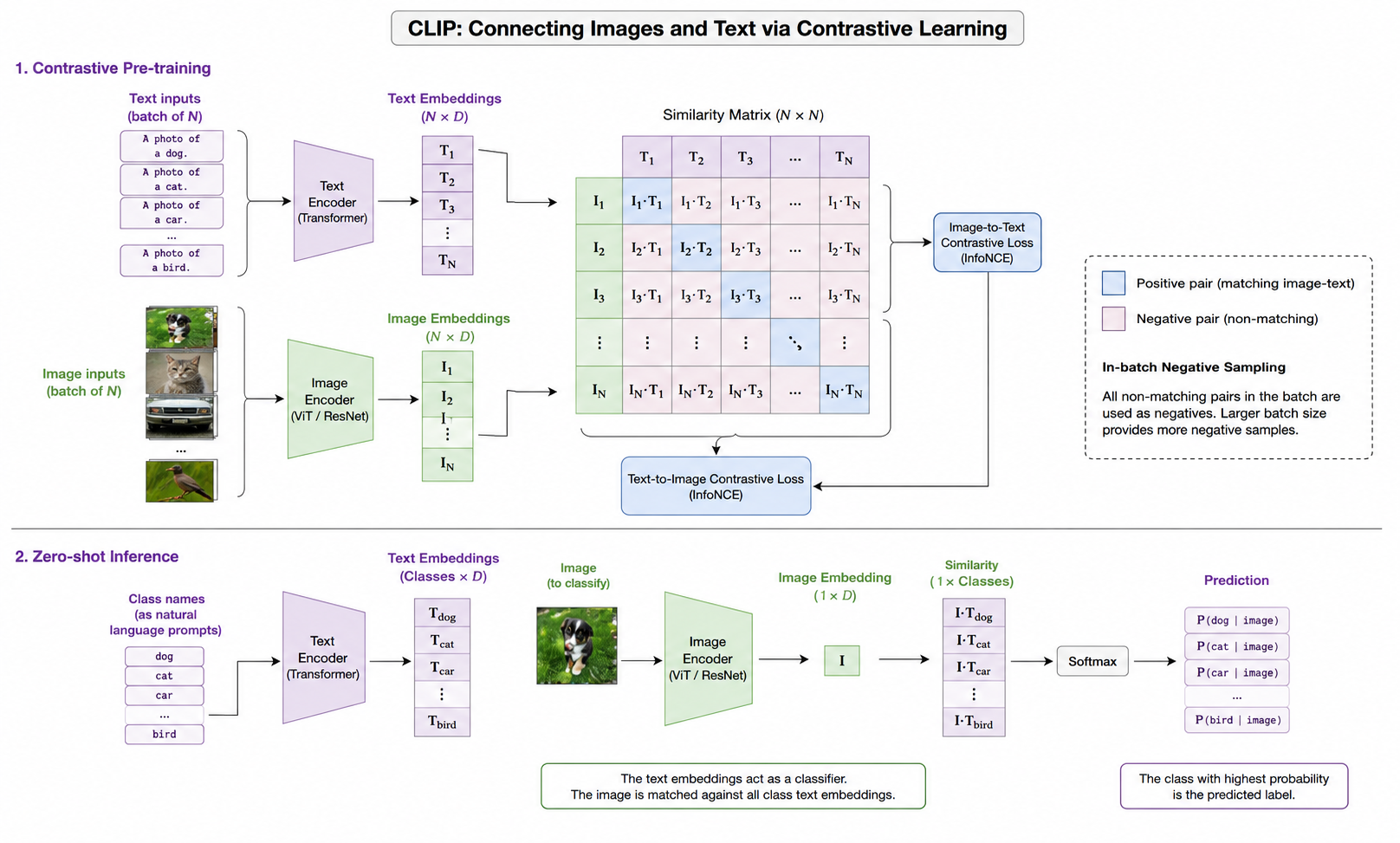}
\caption{The CLIP framework. (Left) Contrastive pretraining jointly aligns an image encoder and a text encoder over batches of image-text pairs; (right) zero-shot prediction constructs a classifier from natural-language prompts. Illustration inspired by~\cite{radford2021clip}.}
\label{fig:clip}
\end{figure*}

CLIP~\cite{radford2021clip} jointly trains an image encoder and a text encoder with a symmetric contrastive objective over 400M web image-text pairs, replacing fixed class labels with natural-language supervision (Fig.~\ref{fig:clip}). After pretraining, language references the learned visual concepts, enabling strong zero-shot transfer (e.g., $76.2\%$ zero-shot ImageNet top-1, matching a supervised ResNet-50)---the open-vocabulary property that fixed-label backbones lack.

\subsubsection{VideoCLIP}
\label{sec:videoclip}

VideoCLIP~\cite{videoclip2021} carries contrastive pretraining to video with a dual encoder trained on HowTo100M, using two ideas: positives from temporally overlapping (not exactly aligned) video-text clips, and retrieval-mined hard negatives. It reaches strong zero-shot text-video retrieval, QA, and action segmentation, in cases beating supervised methods.

\subsubsection{Frozen-in-Time}
\label{sec:frozen}

Frozen-in-Time~\cite{bain2021frozen} pairs a ViT/TimeSformer-style space-time visual encoder with a text encoder, ingesting images and video uniformly (an image is a single-frame video) under a curriculum that gradually adds temporal context. Trained on the modest WebVid-2M plus CC3M, it still reaches state-of-the-art retrieval---and this space-time encoder is the direct architectural ancestor of EgoVLP.

\subsubsection{EgoVLP}
\label{sec:egovlp}

\begin{figure*}[t]
\centering
\includegraphics[width=0.9\textwidth]{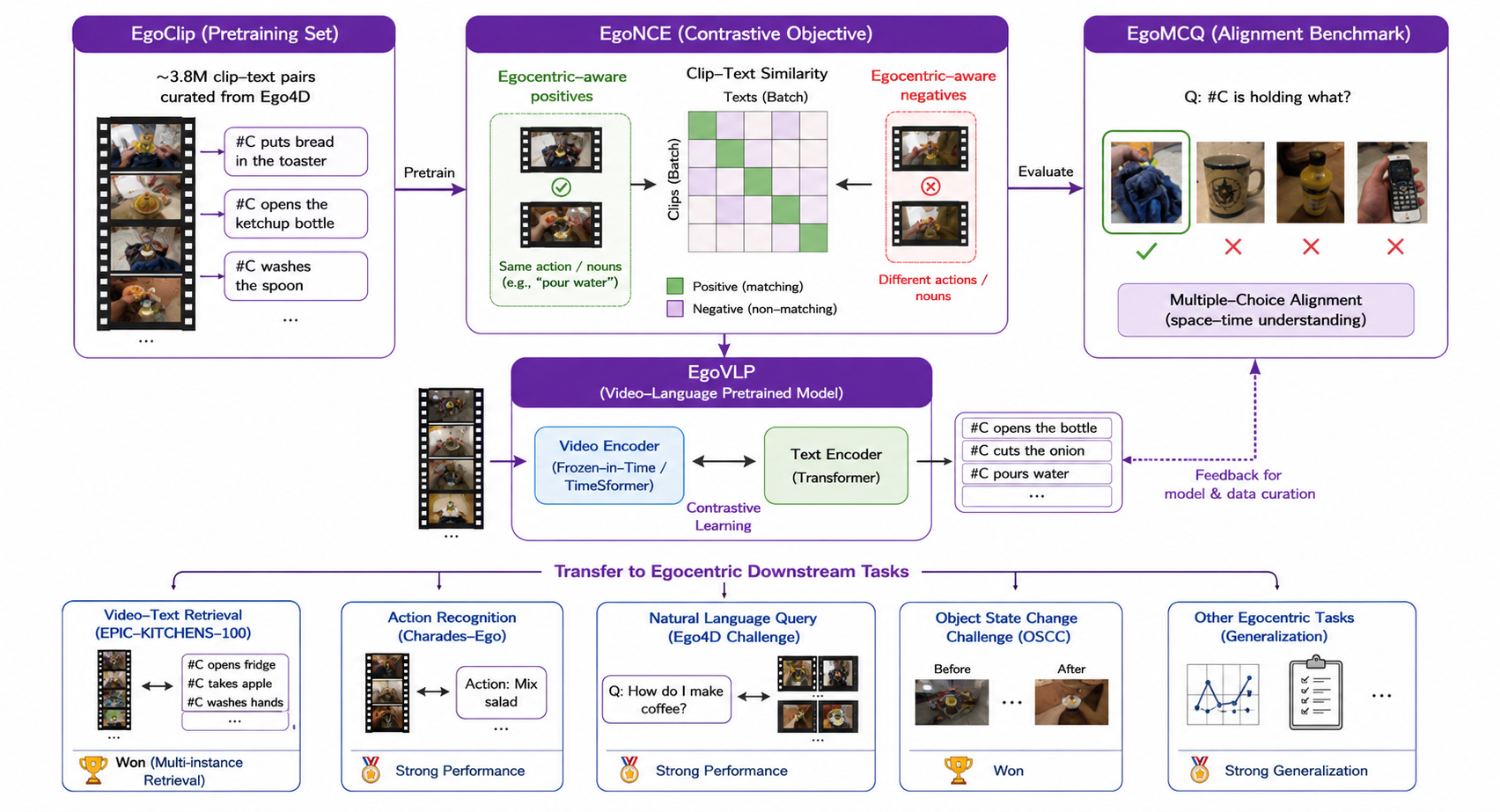}
\caption{The EgoVLP framework: the EgoClip pretraining set curated from Ego4D, a space-time video-language model trained with the EgoNCE objective, and the EgoMCQ development benchmark, with transfer to egocentric downstream tasks. Illustration inspired by~\cite{lin2022egovlp}.}
\label{fig:egovlp}
\end{figure*}

EgoVLP~\cite{lin2022egovlp} transfers this design to first-person video along three axes (Fig.~\ref{fig:egovlp}): EgoClip, a $\sim$3.8M clip-text pretraining set curated from Ego4D; EgoNCE, a contrastive objective that mines egocentric-aware positives (clips sharing actions or nouns) and negatives; and the EgoMCQ alignment benchmark. Built on the Frozen-in-Time / TimeSformer encoder, it transfers strongly across five egocentric tasks and won the EPIC-KITCHENS multi-instance-retrieval and OSCC challenges, establishing the template the rest of this survey builds on.

\subsection{Summary of Model Evolution}
\label{sec:evolution_summary}

The trajectory traced across this section moves from hand-engineered temporal modeling toward language-supervised, open-vocabulary representation. Classical video understanding established three representational vocabularies: spatiotemporal convolutions, from the Sports-1M study~\cite{karpathy2014largescale} through C3D~\cite{tran2015c3d} and its factorized successors~\cite{qiu2017p3d,tran2018r2plus1d} to the Kinetics-pretrained I3D~\cite{carreira2017i3d}; recurrent temporal aggregation over per-frame CNN features~\cite{donahue2015lrcn,ng2015beyond}, refined for first-person video by LSTA~\cite{sudhakaran2019lsta}; and parallel appearance--motion streams~\cite{simonyan2014twostream}, extended to the multimodal EPIC-Fusion Temporal Binding Network~\cite{kazakos2019epicfusion}. These families supplied the backbones and baselines on which egocentric benchmarks such as EPIC-KITCHENS~\cite{damen2022rescaling} and Ego4D~\cite{grauman2022ego4d} still rely.

The transition to attention replaced these inductive biases with learned space-time interactions. ViT~\cite{dosovitskiy2021vit} introduced patch tokenization, TimeSformer~\cite{bertasius2021timesformer} and ViViT~\cite{arnab2021vivit} factorized self-attention across space and time, and VideoMAE~\cite{tong2022videomae} made data-hungry video transformers trainable through self-supervised masked modeling---the property that made these backbones dominant on Ego4D and EPIC-KITCHENS leaderboards. The final shift, from vision models to vision-language models, supervised visual encoders with natural language: CLIP~\cite{radford2021clip} established image-text contrastive pretraining, VideoCLIP~\cite{videoclip2021} and Frozen-in-Time~\cite{bain2021frozen} carried it to video, and EgoVLP~\cite{lin2022egovlp} transferred the design to the egocentric domain via EgoClip and EgoNCE. Each generation thus inherited representational machinery from the previous
one while changing its temporal modeling and supervision strategy, setting the stage for the egocentric vision-language models that the remainder of this survey examines.

\section[Vision-Language Models for Egocentric Video Understanding]
{Vision-Language Models for Egocentric Video\\Understanding}
\label{sec:vlm}

Building on the model evolution reviewed in
Section~\ref{sec:evolution}, this section examines how vision-language models extend conventional video representations toward the language-grounded understanding required for egocentric and embodied AI.

We first diagnose why standard, predominantly exocentric vision-language models underperform on egocentric video (Section~\ref{sec:vlm-fail}), then organize the egocentric vision-language landscape into four broad model groups (Section~\ref{sec:vlm-arch}), and finally synthesize their tradeoffs (Section~\ref{sec:vlm-compare}).

\subsection{Why Standard VLMs Fail in Egocentric Videos}
\label{sec:vlm-fail}

Despite their success on web-scale third-person benchmarks, general-purpose
vision-language models degrade sharply when confronted with first-person video.
The failures cluster into four interrelated categories.

\subsubsection{Weak Interaction Understanding}
\label{sec:weak-interaction}

\begin{figure*}[t]
  \centering
  \includegraphics[width=0.9\textwidth]{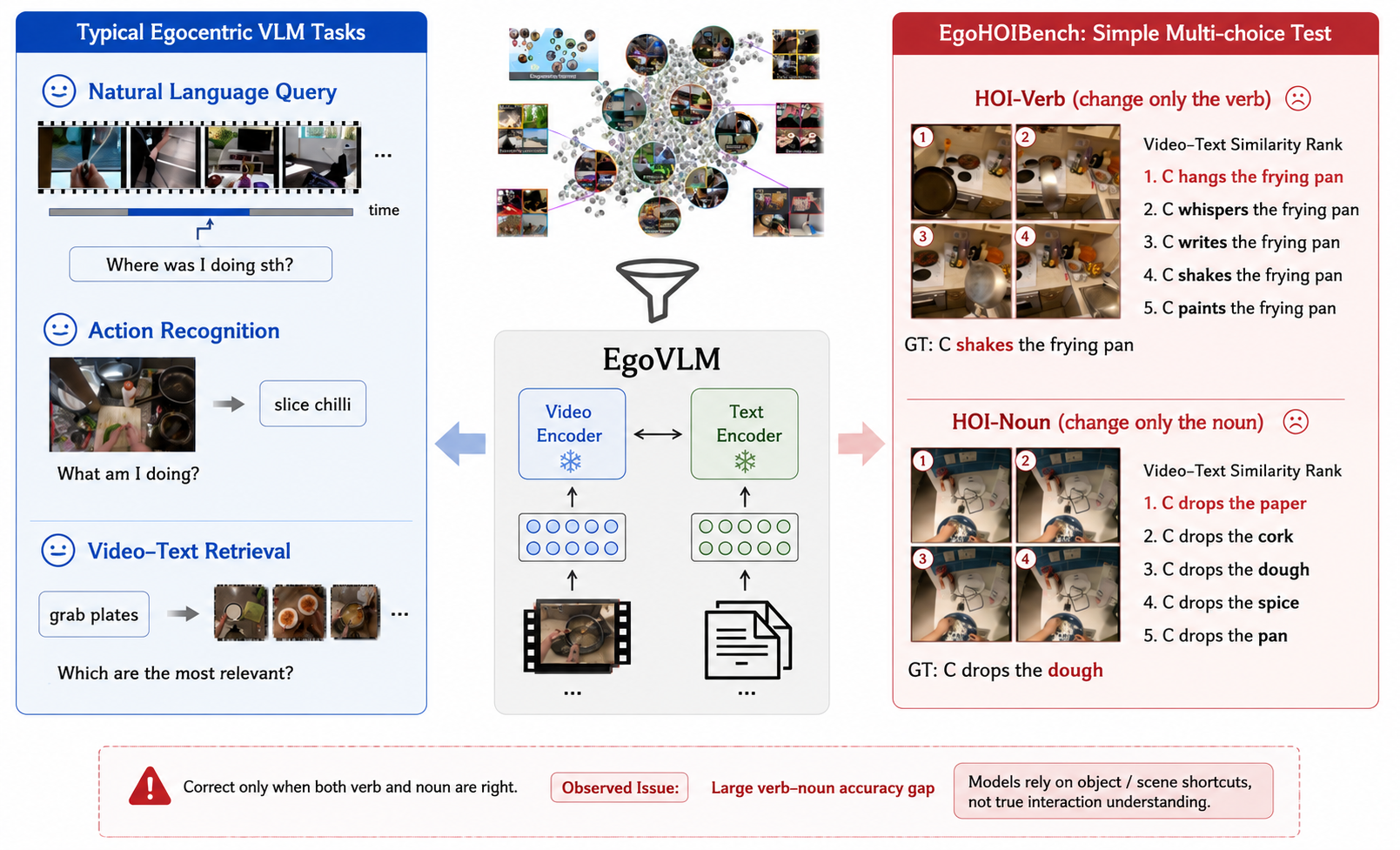}
  \caption{The EgoHOIBench diagnostic. A model must select the correct caption
  from candidates that differ by only a single verb or noun, scoring a trial
  correct only when both are right; egocentric video-language models exhibit a
  large verb--noun accuracy gap, revealing only superficial hand--object
  interaction understanding. Illustration inspired by~\cite{dong2025truly}.}
  \label{fig:egohoibench}
\end{figure*}

The most direct evidence comes from Xu~\textit{et al.}~\cite{dong2025truly},
whose EgoHOIBench asks a model to pick the correct Ego4D caption from candidates
differing by a single verb or noun, scoring a trial correct only when both are
right (Fig.~\ref{fig:egohoibench}). Even models pretrained on millions of
interaction clips fail: LaViLa scores $74.33\%$ on nouns but only $46.61\%$ on
verbs---a $\sim$$28$-point gap traced to object-recognition shortcuts. Their EgoNCE++ objective, which mines HOI-aware hard-negative captions, raises
the LaViLa++ variant's verb accuracy by 34.02 percentage
points~\cite{dong2025truly}. This verb--noun asymmetry is
the empirical core of the survey's thesis: standard models recognize \emph{what}
objects are present but not \emph{how} they are manipulated.

\subsubsection{Poor Temporal Reasoning}

EgoSchema~\cite{mangalam2023egoschema} crystallized the
temporal-reasoning gap: its long-form Ego4D questions carry a median
``temporal certificate'' of roughly $100$ seconds---far longer than those of
prior datasets. In the original evaluation, the tested billion-parameter
models scored below $33\%$ (chance $20\%$), whereas humans reached
approximately $76\%$. Xu~\textit{et al.}~\cite{xu2026temporal} trace this
limitation to objectives that reward answer correctness without temporal
consistency, allowing models to exploit frame-level shortcuts. Their
order-aware reinforcement-learning method, which contrasts ordered and
shuffled frames, lifts a 3B backbone from $20.4\%$ to $49.7\%$ on
EgoSchema and outperforms substantially larger baselines, indicating that
objective design is an important factor beyond model scale.

\subsubsection{Object Ambiguity}
\label{sec:object-ambiguity}

The active object is often small, hand-held, partially occluded, off-center, or
truncated, and must be told apart from many distractors. EgoTracks~\cite{tang2023egotracks}
shows single-object trackers scoring far below their conventional-benchmark
level because objects repeatedly vanish and must be re-detected, and
VISOR~\cite{darkhalil2022visor} reports small objects as a dominant failure mode
even after fine-tuning. Active-object work~\cite{shan2020understanding,thakur2024anacto}
shows that identifying which object is \emph{in contact} needs explicit
hand-contact reasoning generic detectors lack---so for VLMs this surfaces as the
noun-shortcut bias: they name salient objects but cannot bind them to the acting
hand.

\subsubsection{Motion-Induced Noise}
\label{sec:motion-noise}

Continuous ego-motion, blur, and viewpoint instability corrupt the dense
spatiotemporal features VLMs rely on. Chavis~\textit{et al.}~\cite{chavis2025dexfocus}
show a stabilized hand-focused crop improves Ego-Exo4D keystep recognition
($+17$--$22\%$ Top-1), while EgoDistill~\cite{tan2023egodistill} instead exploits
the head-motion signal via IMU to reconstruct video features at $\sim$$200\times$
fewer multiply--adds. Ego-motion is thus at once a nuisance to suppress and a
signal to exploit---a duality exocentric VLMs never learn to handle.

\subsection{Egocentric Vision-Language Architectures}
\label{sec:vlm-arch}

Egocentric vision-language models can be organized into four broad groups
distinguished by their primary objective and output modality.

\subsubsection{Retrieval-Based Methods}
\label{sec:retrieval-methods}

Retrieval-based models learn a joint video--text embedding via contrastive
learning and are evaluated on multi-instance retrieval and multiple-choice
matching. EgoVLP~\cite{lin2022egovlp} pioneered egocentric pretraining with three
components: EgoClip, a $3.8$M clip--text pretraining set curated from Ego4D;
EgoNCE, a contrastive objective that mines action-aware positives and
scene-aware negatives; and EgoMCQ, a $39$K-question development benchmark with
inter-video and intra-video settings. EgoVLP adapts the dual-encoder
Frozen-in-Time architecture~\cite{bain2021frozen} with a TimeSformer video
backbone. EgoVLPv2~\cite{pramanick2023egovlpv2} advances this by inserting
cross-modal fusion directly into the video and language backbones via gated
cross-attention, switchable between dual-encoder and fusion-encoder modes;
this fusion in the backbone is more
parameter-efficient than stacking fusion-specific layers and achieves consistent
improvements across downstream vision-language tasks~\cite{pramanick2023egovlpv2}.
HelpingHands~\cite{zhang2023helpinghands} augments a dual encoder with an
object-aware decoder trained to predict hand and object positions from noisy,
sparse hand--object-detector annotations, improving object grounding while
requiring only RGB at inference and achieving strong zero-shot transfer on
EPIC-KITCHENS-100 multi-instance retrieval and EGTEA~\cite{zhang2023helpinghands}.

\subsubsection{Captioning-Based Methods}
\label{sec:captioning-methods}

Captioning-based methods use natural-language generation as both supervision and
output. LaViLa~\cite{zhao2023lavila} repurposes a pretrained large language model
as a visually-conditioned narrator that densely pseudo-labels Ego4D clips,
together with a rephraser for textual diversity; the contrastively learned embedding gains an absolute
$+10.1\%$ on EGTEA classification and $+5.9\%$ on EPIC-KITCHENS-100 multi-instance
retrieval, and a model trained on only half the Ego4D narrations outperforms a
baseline trained on the full set~\cite{zhao2023lavila}.
Vid2Seq~\cite{yang2023vid2seq}, though not egocentric-specific, established the
dense-captioning paradigm relevant to first-person streams: it augments a
language model with special time tokens to jointly predict event boundaries and
captions in a single sequence, pretrained on the $18$M-video YT-Temporal-1B
corpus using transcribed-speech boundaries as pseudo-events~\cite{yang2023vid2seq}.
These narration-as-supervision approaches directly target the
temporal-synchronization and text-diversity weaknesses of scraped-narration
pretraining.

\subsubsection{Action-Centric Methods}

Action-centric models optimize directly for the verb--noun action structure
central to egocentric tasks. Beyond the recognition and anticipation methods
of Section~\ref{sec:evolution}, recent vision-language work exploits the
verb--noun decomposition explicitly. The EgoNCE++ objective
\cite{dong2025truly} can be viewed as an action-centric pretraining criterion
that targets the verb-understanding deficit by mining hard-negative captions
that alter the verb. EgoHOD~\cite{pei2025egohod} complements this direction
by generating fine-grained hand--object dynamics narrations and training a
lightweight motion adapter, directly targeting interaction-sensitive
representation learning.

\subsubsection{Multimodal and Embodied Extensions}

This group covers multimodal wearable models together with embodied extensions
that use egocentric perception for assistance or control.
ImageBind~\cite{girdhar2023imagebind} is a general multimodal foundation model,
rather than an egocentric-specific architecture, but its joint embedding across
image, text, audio, depth, thermal, and inertial measurements is directly
relevant to multimodal egocentric sensing.

Vinci~\cite{huang2024vinci} is a real-time embodied assistant built on an
egocentric vision-language model for portable devices. It operates continuously
to answer queries about current and historical observations, plan tasks from
past interactions, and generate step-by-step visual demonstrations, with
audio-based hands-free interaction.

EgoMimic~\cite{kareer2024egomimic} scales imitation learning by treating
egocentric human video, captured using Project Aria glasses with 3D hand
tracking, and robot data as complementary embodied demonstrations. It
co-trains a unified policy and reports that one hour of additional human hand
data can provide greater benefit than one hour of additional robot data.

EgoVLA~\cite{villa2025egovla} extends egocentric vision-language modeling to
embodied control. It trains a vision-language-action model on egocentric human
videos to predict future wrist and hand motion and retargets these predictions
through inverse kinematics to a bimanual humanoid. Pretraining on human video
improves both in-domain and out-of-domain generalization on the proposed
12-task Ego Humanoid Manipulation Benchmark, using 100 demonstrations per
task~\cite{villa2025egovla}.

Together, these models represent the transition from multimodal egocentric
perception and assistance toward embodied action.

\subsection{Comparative Analysis of Existing Methods}
\label{sec:vlm-compare}

\begin{table}[!t]
\centering
\caption{Comparison of representative egocentric vision-language models and
relevant multimodal or embodied extensions discussed in
Section~\ref{sec:vlm-arch}. ``EK-100 MIR'' denotes EPIC-KITCHENS-100
multi-instance retrieval; ``cls.'' denotes classification.}
\label{tab:vlm-comparison}
\scriptsize
\renewcommand{\arraystretch}{1.0}
\resizebox{\linewidth}{!}{%
\begin{tabular}{p{2.3cm}p{2.2cm}p{3.2cm}p{2.2cm}p{4.3cm}}
\hline
\textbf{Model} &
\textbf{Category} &
\textbf{Backbone / data} &
\textbf{Key idea} &
\textbf{Reported result} \\
\hline

EgoVLP~\cite{lin2022egovlp} &
Retrieval &
TimeSformer; EgoClip (3.8M) &
EgoNCE action-aware positives and scene-aware negatives &
59.4\% nDCG on EK-100 MIR; 90.7\%/57.2\% on inter-/intra-video EgoMCQ \\

EgoVLPv2~\cite{pramanick2023egovlpv2} &
Retrieval &
TimeSformer-B + RoBERTa; EgoClip &
Gated cross-modal fusion inside the video and language backbones &
Consistent gains on EgoTaskQA, CharadesEgo, and EK-100 MIR \\

HelpingHands~\cite{zhang2023helpinghands} &
Retrieval &
Dual encoder; Ego4D &
Object-aware decoder predicts hand and manipulated-object positions &
Strong zero-shot performance on EK-100 MIR and EGTEA \\

LaViLa~\cite{zhao2023lavila} &
Captioning &
TimeSformer; Ego4D narrations &
LLM-based Narrator and Rephraser generate dense and diverse narrations &
$+10.1\%$ on EGTEA cls.; $+5.9\%$ on EK-100 MIR \\

Vid2Seq~\cite{yang2023vid2seq} &
Captioning &
Language model + time tokens; YT-Temporal-1B &
Time tokens jointly predict event boundaries and captions &
State-of-the-art dense captioning performance; not egocentric-specific \\

EgoHOD~\cite{pei2025egohod} &
Action-centric &
EgoVideo + motion adapter; Ego4D &
HOI-dynamics narrations generated using a detector and an LLM &
$+6.3\%$ on EK-100 MIR; $+16.3\%$ on EGTEA in zero-shot evaluation \\

EgoVLA~\cite{villa2025egovla} &
Embodied extension &
VLA model; egocentric human video &
Wrist and MANO action representation with inverse-kinematics retargeting &
Evaluated on the 12-task Ego Humanoid Manipulation Benchmark \\

ImageBind~\cite{girdhar2023imagebind} &
Multimodal foundation &
Six modalities; image-paired data &
Jointly aligns image, text, audio, depth, thermal, and IMU signals &
Emergent cross-modal retrieval across multiple modalities \\

Vinci~\cite{huang2024vinci} &
Multimodal assistant &
Egocentric VLM; portable-device setting &
Real-time assistance with memory and visual demonstration generation &
Real-time hands-free assistant prototype \\

EgoMimic~\cite{kareer2024egomimic} &
Embodied extension &
Policy model; Aria and robot demonstrations &
Co-trains human and robot data as complementary embodied demonstrations &
One hour of human hand data can provide greater benefit than one hour of robot data \\

\hline
\end{tabular}
}
\end{table}

Table~\ref{tab:vlm-comparison} consolidates the models discussed below.
These model groups trace a clear trajectory from representation learning toward
embodied agency, with distinct tradeoffs. Retrieval-based models---EgoVLP,
EgoVLPv2, and HelpingHands---are the most efficient and excel at alignment tasks
such as EgoMCQ and EPIC-KITCHENS-100 multi-instance retrieval, but their
contrastive objectives are precisely what the diagnostic studies expose as
brittle: by rewarding coarse video--text matching, they develop the noun-biased,
temporally shallow representations that fail EgoHOIBench's verb
substitutions~\cite{dong2025truly} and EgoSchema's long-horizon
reasoning~\cite{mangalam2023egoschema}. Captioning-based models, LaViLa and
Vid2Seq, address the data-quality side by generating dense, synchronized, and
diverse narration, which improves both retrieval and downstream
recognition~\cite{zhao2023lavila}; their generative backbones also enable
open-ended question answering that pure dual encoders cannot support.
Action-centric methods such as EgoNCE++ and EgoHOD target the verb--noun
interaction structure directly. Multimodal models such as ImageBind and Vinci
broaden sensing and assistance, whereas embodied extensions such as EgoMimic
and EgoVLA learn policy or action outputs for deployment.

Backbones and data scale vary accordingly: retrieval and captioning models
converge on TimeSformer or ViT video encoders contrastively trained on
Ego4D-derived corpora such as EgoClip's $3.8$M pairs~\cite{lin2022egovlp}, while
embodied policy models such as EgoMimic and EgoVLA train on combinations of
human video and robot demonstrations and predict action or policy outputs
rather than text alone. Efficiency spans
orders of magnitude, from EgoDistill's roughly $200\times$ reduction in
multiply--adds~\cite{tan2023egodistill} to multi-billion-parameter models that
still underperform humans on EgoSchema~\cite{mangalam2023egoschema}. The
persistent gaps identified in Section~\ref{sec:vlm-fail}---weak interaction
understanding, reliance on spatial shortcuts over temporal reasoning,
active-object ambiguity, and motion-induced feature corruption---remain only
partially addressed: temporal objective redesign closes much of the temporal
gap~\cite{xu2026temporal} and EgoNCE++ much of the interaction
gap~\cite{dong2025truly}, but no single model yet unifies robust fine-grained
interaction understanding, long-horizon temporal reasoning, and real-time
embodied action. Closing that gap is the central open challenge on the path from
hand--object interaction to embodied AI.

\section[Hand--Object Interaction Understanding in Egocentric Videos]
{Hand--Object Interaction Understanding in\\Egocentric Videos}
\label{sec:hoi-understanding}

The first-person camera frames the world through the wearer's hands. Where
third-person video treats hands as one body part among many, egocentric video
places them at the center of nearly every informative moment: the hands are
what move, what occlude, and what carry the wearer's intent into the scene.
This makes hand--object interaction (HOI) understanding the empirical core of
egocentric perception, and the natural hinge between low-level visual parsing
and the language-grounded, embodied reasoning that the rest of this survey
pursues. This section traces that hinge from the bottom up: why hands are
foundational rather than peripheral (Section~\ref{sec:why-hands}), how the
field detects and reconstructs them (Section~\ref{sec:hand-detection}), how it
identifies and reasons about the objects they engage
(Section~\ref{sec:object-centric}), how the interaction unfolds as a temporal
process from reach to release (Section~\ref{sec:fine-grained-hoi}), and where
current understanding still breaks down (Section~\ref{sec:hoi-limitations}).

\subsection{Why Hands Matter in Egocentric Understanding}
\label{sec:why-hands}

In first-person video the hands are not merely visible---they are the
privileged signal. Because the camera is worn and its motion is driven by the
wearer's attention, the hands and the objects they manipulate tend to occupy
the center of the frame, and reading an action is largely a matter of reading
how the hands engage objects~\cite{bandini2023analysis,kwon2021h2o}. This is
why the dominant egocentric benchmarks are organized around manipulation:
EPIC-KITCHENS-100 decomposes every action into a (verb, noun) pair grounded in
hand activity~\cite{damen2022rescaling}, and Ego4D dedicates an entire
benchmark suite, Hands and Objects, to the moments in which hands change the
state of the world~\cite{grauman2022ego4d}.

The dependence runs deeper than annotation convenience: hand understanding
measurably drives downstream performance. AssemblyHands demonstrates this
directly, showing that improving the quality of estimated 3D hand poses yields
a corresponding improvement in egocentric action
recognition~\cite{ohkawa2023assemblyhands}. Hands are therefore best treated
not as a peripheral cue to be detected once, but as a foundational primitive
whose fidelity propagates through the entire perception stack---from
recognition and anticipation up to the vision-language and embodied models of
Sections~\ref{sec:vlm} and~\ref{sec:applications}. The cost of this centrality
is that hands are also the hardest thing in the frame to parse: they are
small, fast, self-similar, frequently self-occluded, and routinely occlude the
very objects that define the action~\cite{bandini2023analysis}. The remainder
of this section is, in effect, the story of how the field has confronted that
difficulty.

\subsection{Hand Detection and Tracking}
\label{sec:hand-detection}

Recovering the hands is the entry point to every downstream interaction task,
and the field has moved through three clearly separable regimes: sparse 2D
keypoints, parametric mesh models, and---most recently---transformer-based
reconstructors that target the egocentric failure modes head-on. We follow
that progression, treating the MANO hand model as the parametric bridge that
connects the keypoint era to the mesh era.

\subsubsection{Keypoint-Based Methods}
\label{sec:keypoint-methods}

Early RGB keypoint detectors beat occlusion with geometry rather than labels: Simon~\textit{et al.}~\cite{simon2017hand} introduced \emph{multiview bootstrapping}, triangulating dome detections into 3D and reprojecting them to relabel occluded views, yielding an occlusion-robust 21-keypoint detector. This underpins OpenPose~\cite{cao2019openpose} and, later, the mobile-real-time MediaPipe Hands~\cite{zhang2020mediapipe}. But sparse 2D keypoints cannot express contact, orientation, or articulation under occlusion---the quantities that matter for manipulation. MANO~\cite{romero2017mano} closed this gap with a low-dimensional parametric hand mesh whose pose space spans a standard grasp taxonomy, becoming the near-universal output representation for mesh-based methods. Accurate 3D supervision then followed from InterHand2.6M~\cite{moon2020interhand} and, in the egocentric regime, AssemblyHands~\cite{ohkawa2023assemblyhands,sener2022assembly101} (3.0M images, 490K egocentric).

\subsubsection{Transformer-Based Hand Modeling}
\label{sec:transformer-hand}

The transformer replaced hand-crafted mesh regressors with attention over vertices and joints: METRO cast mesh recovery as masked sequence modeling for occlusion robustness~\cite{lin2021metro}, Mesh Graphormer added graph convolution for local surface structure~\cite{lin2021meshgraphormer}, and HandOccNet built occlusion handling into the architecture by injecting features into occluded regions before regressing MANO~\cite{park2022handoccnet}.

The current frontier is scale and egocentric realism. HaMeR scales a plain ViT backbone with large in-the-wild data for robustness across viewpoints and occlusions~\cite{pavlakos2024hamer}, and WiLoR adds real-time localization and tracking from monocular video~\cite{potamias2025wilor}. The 2024--2025 wave then targets specifically egocentric failure modes: WildHands corrects near-field perspective distortion using field-of-view cues, topping the ARCTIC egocentric split at a fraction of HaMeR's size~\cite{prakash2024wildhands,fan2023arctic}, while HaWoR~\cite{zhang2025hawor} and Dyn-HaMR~\cite{yu2025dynhamr} recover \emph{world}-frame and 4D interacting-hand motion from a moving camera by coupling SLAM with hand priors. The trajectory is clear---from where the hands are, to their shape, to where and how they move in the world---precisely what embodied agents require.

\begin{figure*}[t]
  \centering
  \includegraphics[width=0.9\textwidth]{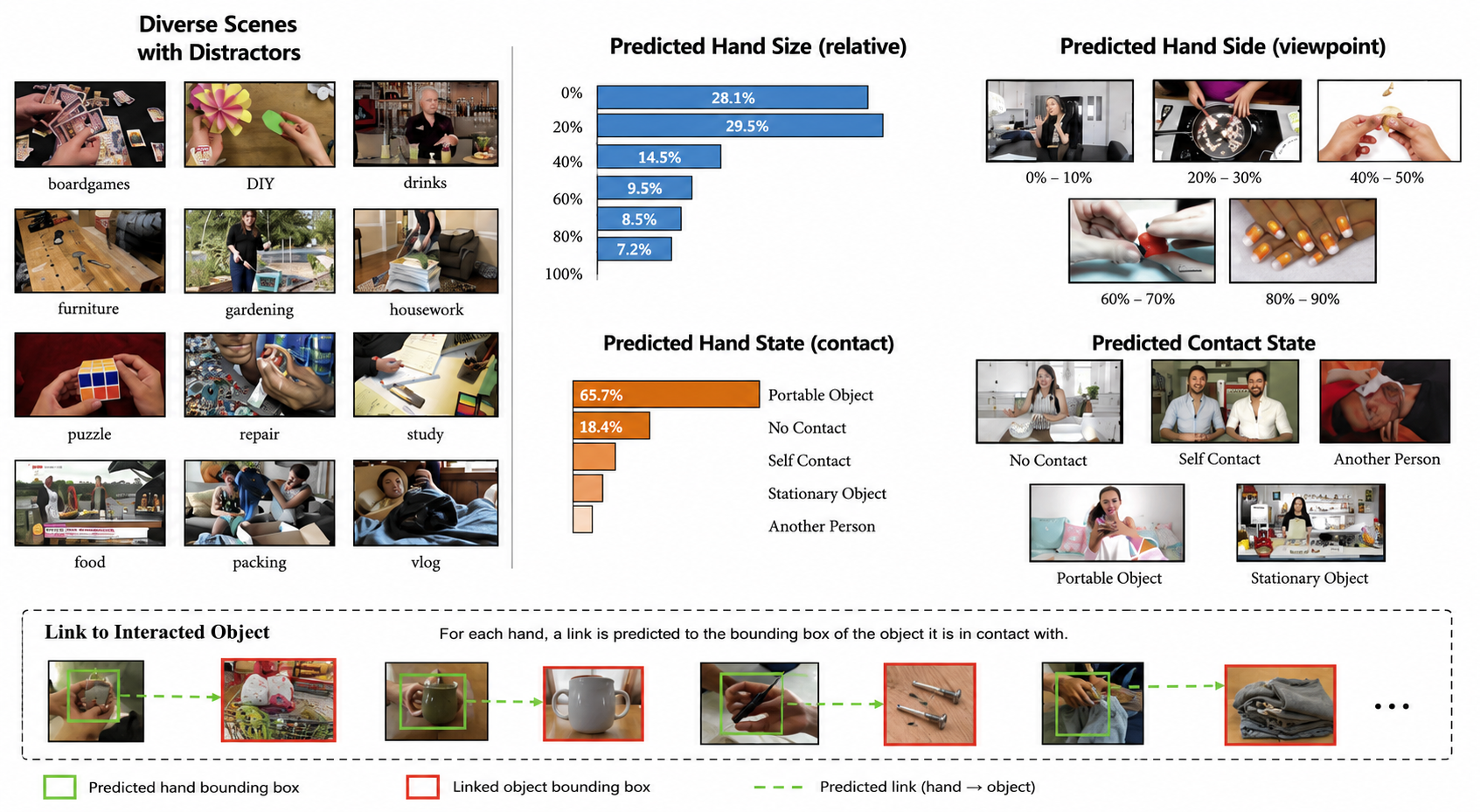}
  \caption{The 100DOH hand-object detector. For each hand it jointly predicts a
  bounding box, side (left/right), contact state (none, self, other-person,
  portable, or non-portable), and a link to the box of the object in contact,
  recovering the active object among many distractors. Illustration inspired
  by~\cite{shan2020understanding}.}
  \label{fig:hand_detector}
\end{figure*}

\subsection{Object-Centric Interaction Understanding}
\label{sec:object-centric}

Recovering the hand is only half the problem; understanding the interaction
requires identifying \emph{which} object is being acted upon and \emph{how} it
affords being acted upon. These are the questions of manipulated-object
detection and affordance understanding.

\subsubsection{Manipulated Object Detection}
\label{sec:manipulated-object}

The defining challenge is that a frame holds many objects but only one or two are \emph{active}. The foundational ``100 Days of Hands'' detector jointly predicts, per hand, its box, side, contact state, and a link to the in-contact object~\cite{shan2020understanding}; Hands23 enriched this with pixel-level segments and a finer touch/hold/use vocabulary~\cite{cheng2023hands23}, and EPIC-KITCHENS VISOR supplied dense egocentric hand/active-object masks under heavy occlusion~\cite{darkhalil2022visor}.

A complementary thread anticipates the active object \emph{before} contact---central to assistive applications. Furnari~\textit{et al.}~\cite{furnari2017next} forecast the next-active object from trajectories; StillFast unified this into an end-to-end short-term anticipation model (next object, verb, time-to-contact) that topped the Ego4D leaderboard~\cite{ragusa2023stillfast}; and ANACTO extends the idea~\cite{thakur2024anacto}. Helping Hands feeds this back into representation learning, adding an object-aware decoder that grounds hands and objects while staying RGB-only at inference~\cite{zhang2023helpinghands}.

\subsubsection{Affordance Understanding}
\label{sec:affordance}

Beyond \emph{which} object lies \emph{how} it can be acted upon---affordance---and egocentric video teaches it at scale. Nagarajan~\textit{et al.}~\cite{nagarajan2019hotspots} learn weakly-supervised interaction \emph{hotspots} predicting where an object would be manipulated even at rest; EGO-TOPO lifts affordances from objects to a topological map of activity zones~\cite{nagarajan2020egotopo}; and AGD20K addresses the supervision bottleneck by transferring from abundant exocentric demonstrations to egocentric views, now the standard weakly-supervised benchmark~\cite{luo2022affordance}.

\subsection{Fine-grained Interaction Understanding}
\label{sec:fine-grained-hoi}

An interaction is not an instant but a process: the hand approaches, makes
contact, closes into a grasp, manipulates the object through a state change,
and releases. Understanding this process at fine temporal granularity is what
separates genuine interaction understanding from object recognition. There is
no single canonical taxonomy of these stages in the literature; we therefore
organize them along the most empirically grounded axis available---Ego4D's
state-change timeline, which annotates every interaction with a
\emph{pre-condition} frame, a \emph{contact} frame, a \emph{point-of-no-return}
(PNR), and a \emph{post-condition} frame~\cite{grauman2022ego4d}---and overlay
the contact-state vocabulary of
100DOH/Hands23~\cite{shan2020understanding,cheng2023hands23} and the grasp
taxonomy of Feix~\textit{et al.}~\cite{feix2016grasp} as the semantic
dimension. The five stages below should be read as this lifecycle, not as a
fixed label set.

\subsubsection{Approach}
\label{sec:approach}
The interaction begins before contact, as the hand reaches toward its target.
This pre-contact phase is the domain of next-active-object anticipation and
hand-trajectory forecasting: predicting which object will be engaged and where
the hand is heading, from motion and scene context
alone~\cite{furnari2017next,ragusa2023stillfast,thakur2024anacto}. Affordance
hotspots provide the spatial prior for this stage, indicating the likely
landing point of the reach before the hand arrives~\cite{nagarajan2019hotspots}.

\subsubsection{Touch}
\label{sec:touch}
Contact onset is the first observable transition of the interaction, marked
explicitly in Ego4D as the \emph{contact} frame and captured in 100DOH as the
shift from a no-contact to an in-contact state~\cite{grauman2022ego4d,shan2020understanding}.
Detecting this transition reliably is non-trivial in the egocentric view,
where the hand frequently occludes the contact point itself; pixel-level
resources such as VISOR were built precisely to study hand--object boundaries
through these contact events~\cite{darkhalil2022visor}.

\subsubsection{Grab}
\label{sec:grab}
Once contact is established, the hand closes into a grasp, and the
\emph{manner} of grasping carries rich information about both object and
intent. The canonical vocabulary is the GRASP taxonomy of Feix~\textit{et
al.}~\cite{feix2016grasp}, which consolidates prior work into 33 distinct
static grasp types organized by opposition, virtual-finger assignment, and the
power--precision spectrum. Hands23 operationalizes this for in-the-wild images
by annotating grasp type alongside contact state~\cite{cheng2023hands23}, and
the MANO pose space was itself constructed to span these grasps, which is why
modern mesh reconstructors can express them~\cite{romero2017mano}.

\subsubsection{Hold and Manipulation}
\label{sec:hold}
The manipulation phase---holding and using the object to effect a change---is
where the interaction does its work. In Ego4D, this phase may include the
point-of-no-return, which marks the frame at which an irreversible state change
begins, while state-change classification categorizes the resulting
change~\cite{grauman2022ego4d}. Here contact is dynamic rather than static,
which dedicated datasets capture explicitly: ARCTIC records dexterous bimanual
manipulation of articulated objects with dense, time-varying 3D
contact~\cite{fan2023arctic}; ContactPose and GRAB provide hand--object and
whole-body contact maps paired with pose~\cite{brahmbhatt2020contactpose,taheri2020grab};
and HOI4D, Assembly101, and HOT3D supply temporally segmented, contact-rich
manipulation sequences in egocentric and multi-view
settings~\cite{liu2022hoi4d,sener2022assembly101,banerjee2024hot3d}.

\subsubsection{Release}
\label{sec:release}
The interaction may terminate as the hand disengages. Ego4D's
\emph{post-condition} frame documents the resulting object state, but it does
not explicitly annotate release in every interaction~\cite{grauman2022ego4d}.
Detecting release nevertheless closes the temporal loop and enables the system
to anticipate the next interaction, making release detection important for
parsing long, continuous first-person activities.

\subsection{Limitations of Current HOI Understanding}
\label{sec:hoi-limitations}

\begin{table}[H]
\centering
\caption{Comparison of representative hand--object interaction methods and resources, spanning hand recovery, manipulated-object reasoning, and interaction datasets (Section~\ref{sec:hoi-understanding}).}
\label{tab:hoi-comparison}

\footnotesize
\renewcommand{\arraystretch}{1.25}

\resizebox{\linewidth}{!}{%
\begin{tabular}{p{2.3cm}p{2.2cm}p{3.2cm}p{2.2cm}p{4.3cm}}
\hline
\textbf{Method / Resource} &
\textbf{Category} &
\textbf{Output / representation} &
\textbf{Egocentric focus} &
\textbf{Key contribution} \\
\hline

100DOH~\cite{shan2020understanding} &
Detection &
Hand box, side, contact state, in-contact object &
Internet-scale &
Joint hand + active-object contact detection \\

Hands23~\cite{cheng2023hands23} &
Detection &
Pixel segments, grasp, contact vocabulary &
In-the-wild &
Richer 2D parsing (touch/hold/use) \\

MANO~\cite{romero2017mano} &
Parametric model &
Low-dimensional pose + shape mesh &
--- &
Near-universal hand output representation \\

HandOccNet~\cite{park2022handoccnet} &
Reconstruction &
MANO mesh &
Occlusion-heavy &
Feature injection into occluded regions \\

HaMeR~\cite{pavlakos2024hamer} &
Reconstruction &
3D hand mesh &
In-the-wild &
ViT-scale, robust across viewpoints \\

WildHands~\cite{prakash2024wildhands} &
Reconstruction &
3D hand pose &
Everyday ego &
FOV cues; best 3D pose on ARCTIC ego split \\

HaWoR~\cite{zhang2025hawor} &
Reconstruction &
World-frame hand motion &
Ego video &
SLAM-decoupled recovery with FOV infilling \\

StillFast~\cite{ragusa2023stillfast} &
Anticipation &
Next active object, verb, time-to-contact &
Ego4D &
Tops short-term anticipation leaderboard \\

VISOR~\cite{darkhalil2022visor} &
Dataset &
Dense hand/object masks + relations &
Ego (EPIC) &
Pixel-level under occlusion and state change \\

ARCTIC~\cite{fan2023arctic} &
Dataset &
Dense time-varying 3D contact &
Ego + exo &
Dexterous bimanual articulated manipulation \\

\hline
\end{tabular}%
}

\end{table}
Table~\ref{tab:hoi-comparison} consolidates the representative methods and
resources surveyed in this section.
Despite this maturity, current HOI understanding remains brittle in ways that
matter for the embodied applications ahead. The most consequential limitation
is conceptual rather than perceptual: even egocentric video--language models trained on millions of interaction clips
show limited fine-grained interaction understanding.
EgoHOIBench shows that these models latch onto nouns and scene context while
failing systematically on fine-grained verb distinctions, mistaking
\emph{what} object is present for \emph{how} it is being
manipulated~\cite{dong2025truly}. This verb--noun asymmetry is the empirical
core of the gap this survey identifies, and it persists precisely because
contrastive objectives reward coarse video--text matching over genuine
interaction reasoning.

The perceptual limitations are equally concrete and are, tellingly, the same
egocentric failure modes that the newest reconstruction methods were built to
attack. Occlusion remains pervasive---hands occlude objects and themselves,
which is why occlusion-robust architectures are a research thread of their
own~\cite{park2022handoccnet}. Perspective distortion of near-field hands
degrades 3D estimation in everyday egocentric images~\cite{prakash2024wildhands},
and hands routinely leave the narrow field of view entirely, forcing methods to
infill motion they cannot observe~\cite{zhang2025hawor,yu2025dynhamr}. A second
limitation is the supervision bottleneck: the highest-quality 3D contact and
pose annotations depend on multi-view capture rigs unavailable in the wild,
and even automated pipelines such as AssemblyHands' presuppose such
rigs~\cite{ohkawa2023assemblyhands}, while affordance learning leans on weak
supervision and cross-view transfer to escape the same
constraint~\cite{luo2022affordance}. Finally, models trained on Western,
kitchen-heavy footage generalize poorly across cultures, environments, and
manipulation styles---a bias that compounds every limitation above.

Taken together, these gaps frame the central problem for the sections that
follow: no current system unifies robust fine-grained interaction
understanding, long-horizon temporal reasoning over the interaction lifecycle,
and the world-frame hand recovery that embodied control demands. Closing that
gap is the path from hand--object interaction to embodied AI.

\section{Spatiotemporal Reasoning in Egocentric Video}
\label{sec:spatiotemporal}

The perception stack of Sections~\ref{sec:hoi-understanding}
and~\ref{sec:evolution} tells us \emph{where} the hands are, \emph{what}
objects they touch, and \emph{when} contact begins and ends. None of that, by
itself, is understanding. Understanding requires reasoning over the
\emph{relations} between those entities and over their evolution in
\emph{time}: that the knife is in the right hand, that the right hand is
cutting the onion that sits on the board, and that this step follows peeling
and precedes frying. This section surveys the modeling layer that sits on top
of perception and performs exactly this reasoning. We argue throughout that
explicit structure---scene graphs, hand--object interaction graphs,
environment topologies, and the order- and memory-aware objectives that
operate over them---provides a promising structured bridge from the low-level hand--object interaction discussed in
Section~\ref{sec:hoi-understanding} to the embodied reasoning of
Section~\ref{sec:applications}. We treat spatial reasoning
(Section~\ref{sec:spatial-reasoning}) and temporal reasoning
(Section~\ref{sec:temporal-reasoning}) in turn, show that their fusion in
interaction-aware temporal models is where the field is converging
(Section~\ref{sec:interaction-aware}), and close with hard evidence that
current vision-language models still largely fail at genuine temporal
reasoning (Section~\ref{sec:temporal-limitations}).

\subsection{Spatial Reasoning}
\label{sec:spatial-reasoning}

Spatial reasoning in video has steadily moved away from dense, monolithic
feature tensors toward explicit relational structure, in which a frame becomes
a set of nodes (hands, objects, regions, places) connected by typed edges
(contact, support, proximity, affordance). This shift matters for egocentric
video in particular, because the first-person signal is fundamentally
relational: an action is defined less by appearance than by which hand engages
which object, and how.

\subsubsection{Object Relations}
\label{sec:object-relations}

The decisive idea is the spatio-temporal scene graph. Action Genome
\cite{ji2020actiongenome} reframed an action as a sequence of per-frame scene
graphs (attention, spatial, and contact predicates) with dense annotations,
enabling strong few-shot recognition. This relational view was absorbed into
video transformers by ORViT~\cite{herzig2022orvit}, which injects object-region
tokens from the earliest layers, and pushed to its limit by
STLT~\cite{radevski2021stlt}, which strips appearance away entirely and reasons
over object categories and bounding-box layouts alone---yet still recovers much
of the discriminative signal for compositional actions. The common ancestor is
Wang and Gupta's space-time region graph~\cite{wang2018stgraph}, the template
egocentric hand--object graphs later specialize.

\subsubsection{Hand--Object Relations}
\label{sec:hoi-relations}

In the egocentric setting the most informative edges are those between the
wearer's hands and the objects they manipulate. Interaction-reasoning networks
make these edges first-class: a transformer interaction unit reasons jointly
about each acting hand, its relation to the other hand, and the engaged
objects, showing that explicit two-hand--object relational modeling is
critical for fine-grained egocentric recognition \cite{hoireasoning2022}.
Egocentric Object Manipulation Graphs \cite{dessalene2020egoomg} go further and
build a graph encoding the contact and \emph{anticipated}-contact state between
each hand and the surrounding objects, then reason over the resulting node-state
sequence to anticipate the next action---an explicit acknowledgement that the
hands are the driving force of state change in manipulation activity. These
methods make concrete the verb-centric structure that, as
Section~\ref{sec:temporal-limitations} will show, appearance-based models
systematically miss.

\subsubsection{Scene Context}
\label{sec:scene-context}

\begin{figure*}[t]
  \centering
  \includegraphics[width=0.9\textwidth]{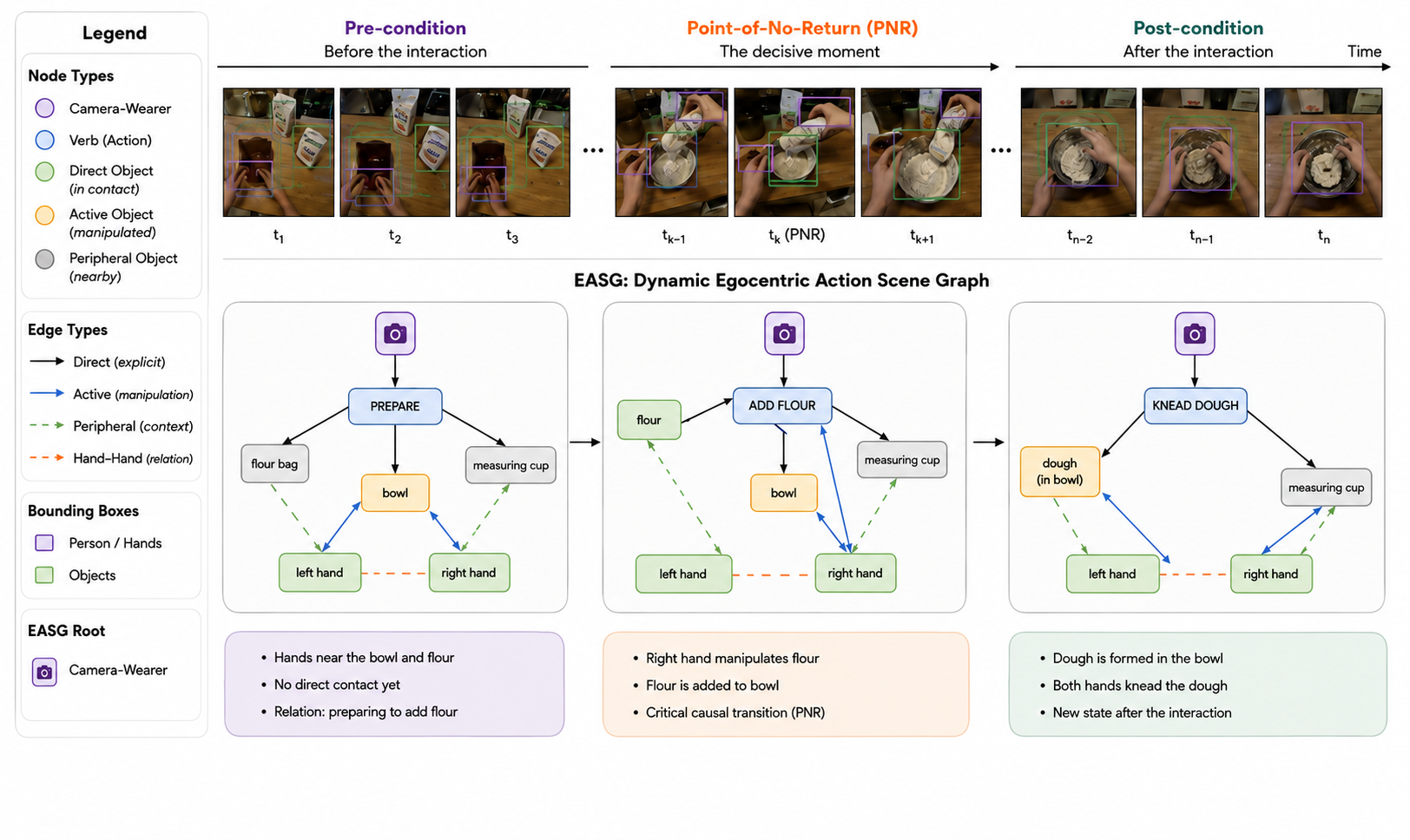}
  \caption{Egocentric Action Scene Graph (EASG). An interaction is represented as
  a temporally evolving graph rooted at the camera-wearer node. Verb (action)
  nodes connect to direct objects, while the manipulated object is linked via
  active edges to the acting hands. Peripheral objects provide contextual
  relations. The graph is grounded by bounding boxes on the pre-condition,
  point-of-no-return (PNR), and post-condition frames, capturing both spatial
  hand--object relations and their evolution over time. Illustration inspired
  by~\cite{rodin2024easg}.}
  \label{fig:easg}
\end{figure*}

Beyond the immediate grasp lies the structure of the surrounding environment
and the longer activity. EGO-TOPO \cite{nagarajan2020egotopo} parses untrimmed
first-person video into a topological graph of activity-centric zones,
modeling a video as a sequence of visits to functional places and using that
graph for affordance discovery and long-term anticipation. Egocentric Action
Scene Graphs (EASG) \cite{rodin2024easg} unify the object- and scene-level
views into a single temporally evolving representation: a dynamic graph with a
camera-wearer node, verb nodes, and direct, active, and peripheral object
nodes, grounded with bounding boxes across the pre-condition, point-of-no-return,
and post-condition frames of each interaction (Fig.~\ref{fig:easg}). Built as a
manual extension of Ego4D, EASG defines a scene-graph-generation task and
yields measurable downstream gains on action anticipation and activity
summarization, and has since been extended into a video question-answering
benchmark for graph-grounded reasoning \cite{rodin2025easgbench}. The
multi-view Home Action Genome \cite{rai2021homage}, which couples a
synchronized ego-view with dense scene-graph and hierarchical activity labels,
provides the egocentric-adjacent bridge back to the Action Genome lineage.

\subsection{Temporal Reasoning}
\label{sec:temporal-reasoning}

If spatial reasoning organizes \emph{what relates to what}, temporal reasoning
organizes \emph{what follows what}. Three problems dominate: learning
representations sensitive to temporal order, carrying information across the
long horizons of first-person video, and localizing the transitions that
separate one action from the next.

\subsubsection{Temporal Ordering}
\label{sec:temporal-ordering}

The foundational insight is that temporal order is itself a free supervisory
signal. Shuffle-and-Learn \cite{misra2016shuffle} trains a network to verify
whether a triplet of frames is correctly ordered, yielding representations
attuned to temporally varying structure without any labels. This idea has
re-emerged at the center of modern egocentric post-training: rather than using
order verification as pretext, recent work converts it into a reward.
Xu~\textit{et al.}~\cite{xu2026temporal} contrast a model's outputs on
temporally ordered versus shuffled frames to derive globally normalized,
verifiable rewards that explicitly favor temporally coherent reasoning, lifting
EgoSchema accuracy from a $20.4\%$ base to $49.7\%$ and suppressing the
frame-level spatial shortcuts that ordinary correctness-based objectives
reward. The lineage from a 2016 pretext task to a 2026 reinforcement-learning
objective is the clearest evidence that order-awareness, far from being solved,
remains the active frontier.

\subsubsection{Long-Term Dependencies}
\label{sec:long-term}

Because egocentric streams run for minutes to hours, dense attention over their
full extent is infeasible, and the field has converged on memory. MeMViT
\cite{wu2022memvit} caches a compressed memory of past clips so that attention
can reference roughly $30\times$ longer temporal context for only about $4.5\%$
additional compute, processing long video online rather than as one giant
input. MC-ViT \cite{balazevic2024mcvit} instead re-purposes a pretrained video
transformer by fine-tuning it to attend to non-parametrically consolidated
memories of past activations, achieving roughly an order of magnitude of memory
compression and setting the state of the art on EgoSchema while outperforming
far larger models. EgoSchema~\cite{mangalam2023egoschema} itself quantifies why this horizon
matters: its temporal-certificate sets have a median length near $100$
seconds. In the original evaluation, the tested billion-parameter models
scored below $33\%$, whereas humans reached approximately $76\%$. Memory, in short, is not an efficiency trick but a
prerequisite for reasoning over the procedural structure of real first-person
activity.

\subsubsection{Action Transitions}
\label{sec:action-transitions}

Reasoning over a long activity requires both temporal grounding and transition
localization. Ego4D \cite{grauman2022ego4d} operationalizes these through two
distinct tasks: natural-language query grounds a textual query to a temporal
window, whereas the Hands-and-Objects benchmark localizes the
point-of-no-return frame and classifies the object state change across the pre-
and post-condition frames. The latter directly represents irreversible action
transitions and connects temporal segmentation to the hand--object interaction
lifecycle of Section~\ref{sec:fine-grained-hoi}.

\subsection{Interaction-Aware Temporal Modeling}
\label{sec:interaction-aware}

\begin{figure*}[t]
  \centering
  \includegraphics[width=0.9\textwidth]{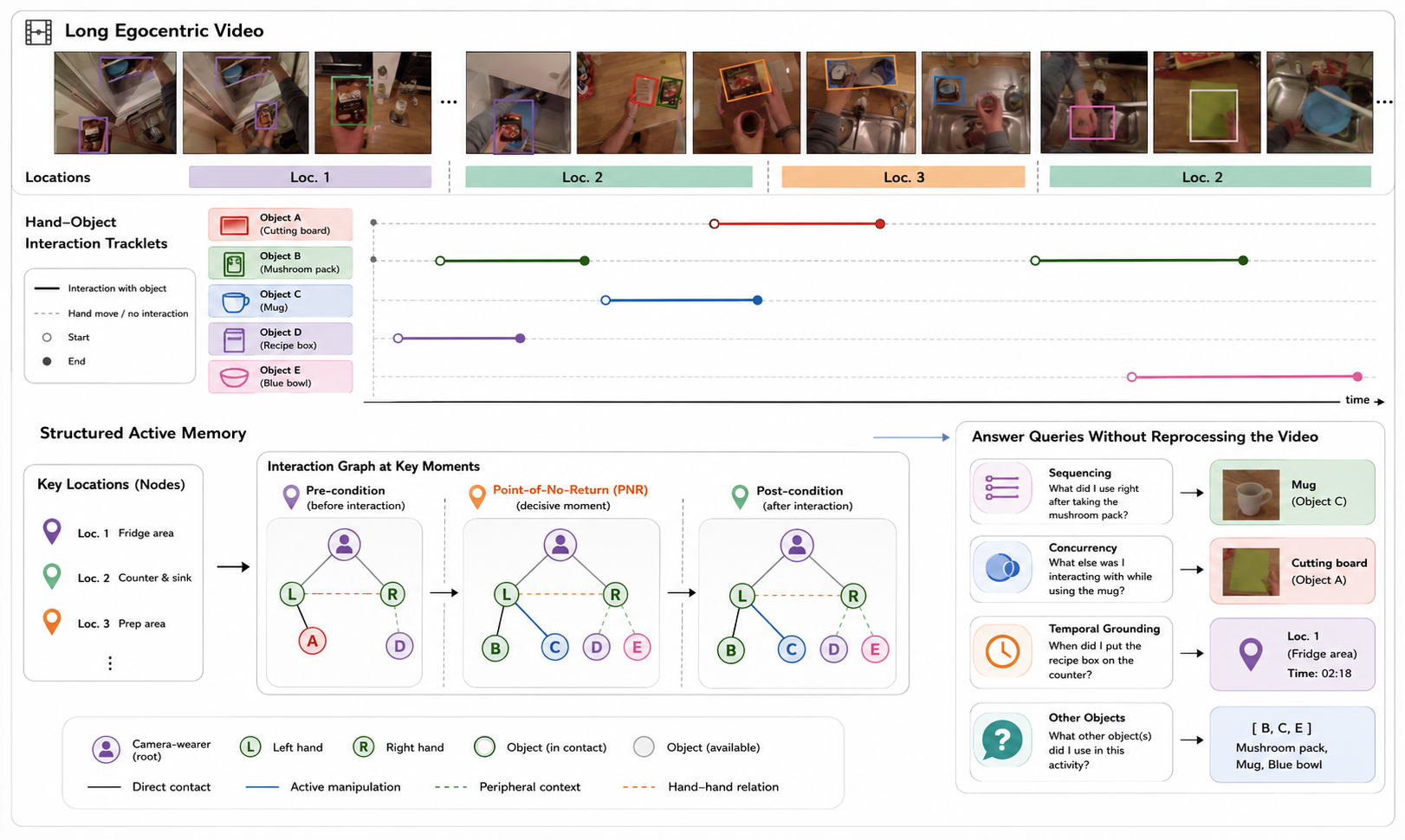}
  \caption{Interaction-aware temporal modeling in egocentric video.
  Hand--object interaction tracklets extracted from long videos are organized
  into a structured active memory indexed by interaction events and locations.
  This representation supports efficient temporal reasoning---including
  sequencing, concurrency, temporal grounding, and retrieval---without
  reprocessing the full video. Interaction-aware representations and
  graph-guided frame selection further improve long-horizon reasoning by using
  hand--object structure as the scaffold for temporal understanding.
  Illustration inspired by~\cite{goletto2024amego}.}
  \label{fig:amego}
\end{figure*}

The two threads above are most powerful when fused: when interaction structure
is used to organize temporal reasoning rather than treated as a separate
perception output. This is the conceptual hinge of the section, and the point
at which graph-based reasoning most clearly carries the paper from perception
toward embodiment. EASG \cite{rodin2024easg} is one instantiation, in that its
graphs are defined over the temporal spine of each interaction. AMEGO
\cite{goletto2024amego} is another and arguably the cleanest: from a single
long egocentric video it constructs a semantic-free active memory of
hand--object interaction tracklets and key locations
(Fig.~\ref{fig:amego}), so that arbitrary queries about sequencing,
concurrency, and temporal grounding can be answered without reprocessing the
video, and it introduces the Active Memories Benchmark of over $20$K queries
on which it substantially outperforms standard video question-answering
baselines. At the representation level, modeling fine-grained hand--object
dynamics during pretraining---through a hand--object detection pipeline, dynamics
narrations, and a dedicated motion adapter---yields large zero-shot gains,
including roughly $+6.3\%$ on EPIC-KITCHENS-100 multi-instance retrieval and
$+16.3\%$ on EGTEA classification \cite{pei2025egohod}, confirming that
interaction structure improves temporal representations and not merely
recognition accuracy. Finally, interaction structure can steer computation
itself: graph-structured frame selection builds an object-centric scene graph
per clip and uses it to choose the few query-relevant keyframes worth attending
to in long-video reasoning \cite{zemskova2026focusgraph}, a direct instance of
graph-guided frame sampling. Across all of these, the pattern is identical---HOI
structure is not the endpoint of perception but the scaffold for temporal
reasoning.

\subsection{Limitations of Existing Temporal Reasoning}
\label{sec:temporal-limitations}

Despite this progress, recent evidence indicates that current vision-language
models often exhibit limited or shortcut-driven temporal reasoning, relying on
spatial and appearance cues that may suffice on existing benchmarks. EgoTempo
\cite{plizzari2025egotempo} makes this concrete and uncomfortable: a text-only
language model with no video input already reaches $31.3\%$ on EgoSchema, and a
model given a \emph{single} frame reaches about $51\%$ on EgoSchema yet only
$9.1\%$ on EgoTempo, whose questions are constructed to be unanswerable without
integrating information across time. Moving from single- to multi-frame input
improves EgoTempo accuracy by roughly $4.3\times$, against only about $1.4\times$
on prior datasets---a gap that exposes how little genuine temporal integration
those prior datasets ever required, and the best model still trails human
accuracy by more than $20$ points. The interaction-level analogue is EgoHOIBench
\cite{dong2025truly}, on which egocentric video-language models trained on
millions of clips collapse under single verb or noun substitutions, revealing a
representation biased toward recognizing objects over understanding
manipulation. Xu~\textit{et al.}~\cite{xu2026temporal} trace the root
cause to objectives that reward answer correctness without rewarding temporal
consistency, allowing models to succeed without reasoning over order at all.
Two conclusions follow for the rest of this survey. First, the apparent
competence of current models is partly an artifact of benchmarks that reward
bag-of-frames recognition, so robust evaluation must isolate temporal and
interaction reasoning explicitly, as EgoTempo and EgoHOIBench begin to do.
Second, the structured, interaction-aware approaches of
Section~\ref{sec:interaction-aware} are not merely one option among many but a
principled response to a diagnosed failure---and closing the gap between
recognition and genuine spatiotemporal reasoning is precisely what the
transition to embodied intelligence in Section~\ref{sec:applications} demands.

\section{Graph-Based and Object-Centric Reasoning}
\label{sec:graph}

Section~\ref{sec:vlm} left us with a verdict: the strongest egocentric
vision-language models recognize \emph{what} is in front of the camera but not
\emph{how} the hands are acting on it. EgoHOIBench~\cite{dong2025truly} pins the
failure precisely---a single verb substitution is enough to fool models trained
on millions of interaction clips---and the cause is structural, not incidental.
A dense grid of patch tokens has no place to store the fact that \emph{this}
hand grasps \emph{that} knife which cuts \emph{this} onion; it can only hope the
relation survives in the soup of attention weights, and it routinely does not.

Graph-based reasoning is a natural structured response, and the field has reached for it
quickly. Make the relation a first-class object: nodes for hands and the objects
they touch, typed edges for contact and manipulation, and a model that reasons
over the graph rather than over the pixels. Our position in this section is that
this response is correct in principle but, for egocentric video specifically,
\emph{still mostly borrowed}. Almost every mature method---the scene-graph
generators, the object-centric encoders, the graph-guided frame samplers---was
built and validated on framed, third-person footage, then transferred to the
first-person setting on the assumption that a graph is a graph. It is not. The
genuinely egocentric graph literature is small, recent, and disproportionately
made of preprints and workshop papers, and where it works it often works by
leaning on a large foundation model to paper over the construction problems that
ego-motion creates. The structure is right; the egocentric foundations under it
are thin. We organize the section to make that gap visible rather than to hide
it: why graphs suit first-person video (Section~\ref{sec:why-graphs}), what the
third-person scene-graph literature offers and where it breaks on transfer
(Section~\ref{sec:scene-graphs}), the hand--object interaction graph as the one
genuinely native egocentric construction (Section~\ref{sec:hoi-graphs}), graph
structure as a controller over which frames a model even sees
(Section~\ref{sec:graph-sampling}) and over the representations it learns
(Section~\ref{sec:graph-repr}), and the open problems that follow
(Section~\ref{sec:graph-challenges}).

\subsection{Why Graphs for Egocentric Video?}
\label{sec:why-graphs}

The argument for graphs is not that they are elegant; it is that a first-person
action is \emph{defined} by a relation and only incidentally by an appearance.
``Cutting an onion'' and ``holding an onion'' are nearly identical pixel
distributions separated entirely by the state of the hand--knife--onion
relationship over time. A model that encodes that relationship implicitly will,
under pressure, fall back on the easier cue---the onion---which is exactly the
noun-shortcut bias the diagnostics of Section~\ref{sec:vlm-fail} exposed. An explicit graph may reduce reliance on this shortcut by representing the interaction relation and the active object as distinct structural elements.

Crucially, the egocentric setting is unusually friendly to this idea, and we
think the field has under-stated how much. Three properties line up. The entity
set is small and stable---a handful of hands and manipulated objects dominate
nearly every informative frame---so a graph over them stays compact where a
third-person crowd scene would explode. The same objects recur across
minutes-long activities, so a structured, queryable graph amortizes across the
whole video instead of being rebuilt frame by frame; this is the insight that
makes AMEGO's interaction-tracklet memory~\cite{goletto2024amego} practical on
hour-long footage where dense processing is hopeless. And the graph coincides
with supervision the community already pays for: EPIC-KITCHENS-100 labels every
action as a (verb, noun) pair~\cite{damen2022rescaling}, and Ego4D's Hands and
Objects benchmark annotates precisely the object state changes that graph edges
are meant to encode~\cite{grauman2022ego4d}. The conceptual machinery is
inherited from the spatio-temporal graph---ST-GCN over body
joints~\cite{yan2018stgcn}, the space-time region graph of Wang and
Gupta~\cite{wang2018stgraph}---but the fit to first-person data is tighter than
it ever was for third-person video. In the egocentric case the graph is close to
the native data model, not one abstraction among many. That, and not aesthetic
preference, is why the rest of this section treats graphs as the principled bet.

\subsection{Scene Graphs in Video Understanding}
\label{sec:scene-graphs}

Most of what the field knows about generating scene graphs from video, it learned
from third-person data, and the lineage is worth tracing because it is precisely
what gets transferred---and what breaks on transfer. The anchor is Action
Genome~\cite{ji2020actiongenome}, which recast an action as a sequence of
per-frame scene graphs with attention, spatial, and contact predicates and
supplied dense annotations at the scale of hundreds of thousands of frames; it is
the benchmark essentially all video scene-graph generation (VidSGG) still reports
on, and that monoculture is itself a problem we return to below.

The methods themselves have improved steadily and genuinely.
STTran~\cite{cong2021sttran} made dynamic scene-graph generation a transformer
problem---spatial encoder for within-frame relations, temporal decoder for
cross-frame ones---and set the baseline everyone still compares against.
OED~\cite{wang2024oed} did away with the brittle detect-then-classify pipeline,
posing the whole task as one end-to-end set prediction over subject--object pair
tokens and dropping the external trackers and handcrafted trajectories that
earlier systems leaned on; this is a real architectural simplification, not a
relabeling. Two moves then push past mere description. SceneSayer's scene-graph
\emph{anticipation}~\cite{peddi2024scenesayer} models how relationships evolve as
a continuous latent process with neural ODEs/SDEs and forecasts future graphs---a
genuinely forward-looking innovation that the egocentric literature has barely
touched. HyperGLM~\cite{nguyen2025hyperglm} abandons the pairwise-edge assumption
altogether for a hypergraph of higher-order interactions, fuses a spatial entity
graph with a procedural causal one, and feeds the result into an LLM; its
accompanying dataset deliberately spans third-person, egocentric, and drone views,
which is the first serious acknowledgement from this community that viewpoint
matters. A parallel and overdue correction targets the field's severe long-tail
bias: VISA~\cite{li2025visa} debiases visually and semantically and posts
double-digit relative gains on rare predicates, conceding what the
recall@K-on-Action-Genome leaderboard had been hiding---that these models were
quietly ignoring most of the relationship vocabulary. A full account of the area
is available in survey form~\cite{li2024sggsurvey}.

Overall, this literature is strong but remains predominantly third-person, and
its transfer to egocentric video is non-trivial. The framed,
stable-camera assumption that lets STTran link objects cleanly across frames is
exactly what head motion, truncation, and persistent hand occlusion destroy
(Section~\ref{sec:challenges}). One of the few explicitly egocentric video scene-graph-generation pipelines is SAMJAM~\cite{li2025samjam}, and it is telling \emph{how} it
works: rather than train a VidSGG model on first-person data---which barely
exists at scale---it stitches SAM2's mask tracking to a Gemini VLM that proposes
the graphs zero-shot, then binds predicted objects to tracked masks for temporal
consistency on EPIC-KITCHENS. It is a clever proof of concept and we cite it as
such, but it is also a tell: the egocentric community is reaching for frozen
foundation models because the native data and methods to do this properly are not
yet there. That absence, not SAMJAM's results, is the real state of the art.

\subsection{Hand--Object Interaction Graphs}
\label{sec:hoi-graphs}

If anything in this section is genuinely egocentric rather than borrowed, it is
the hand--object interaction (HOI) graph: a scene graph whose privileged nodes are
the wearer's hands and whose load-bearing edges are the contact and manipulation
relations binding hands to the objects they act on. This is the structure
Section~\ref{sec:hoi-understanding} identified as the empirical core of
first-person perception, and it is where the egocentric literature has actually
contributed rather than imported.

\subsubsection{Graph Construction}
\label{sec:graph-construction}

Construction comes down to three decisions---what is a node, what is an edge, how
are both obtained---and the egocentric field has converged on a sensible answer.
Hands and active objects are nodes; edges carry the contact vocabulary
(no contact, self-contact, contact with another person, portable-object contact,
and non-portable-object contact) established by the 100DOH
detector~\cite{shan2020understanding} and Hands23~\cite{cheng2023hands23} refined. Two construction
philosophies then diverge, and the contrast is instructive. EGO-OMG~\cite{dessalene2020egoomg}
and EASG~\cite{rodin2024easg} build \emph{semantically rich} graphs: EGO-OMG
tracks contact and anticipated-contact state per hand, and EASG defines a full
node taxonomy---camera-wearer, verb, and direct/active/peripheral object
nodes---grounded with boxes and hand-extended from Ego4D, later turned into a
QA benchmark for graph-grounded reasoning~\cite{rodin2025easgbench}
(Fig.~\ref{fig:easg}). The strength is expressiveness; the cost, which we should
state plainly, is that this richness depends on dense manual annotation that does
not exist at scale for egocentric video. AMEGO~\cite{goletto2024amego} takes the
opposite bet and, in our reading, the more scalable one: a \emph{semantic-free}
graph of interaction tracklets and key locations that needs no relationship labels,
scales to hour-long video, and is \emph{queried} rather than classified
(Fig.~\ref{fig:amego}). The tension between annotation-hungry expressiveness and
annotation-free scalability is the defining design axis of egocentric HOI graphs,
and it is unresolved.

\subsubsection{Graph Dynamics Over Time}
\label{sec:graph-dynamics}

A static graph is a snapshot; the understanding lives in how it changes. The
field has a ready-made temporal spine for this---Ego4D's interaction lifecycle of
pre-condition, contact, point-of-no-return, and post-condition
frames~\cite{grauman2022ego4d}, developed in Section~\ref{sec:fine-grained-hoi}---and
the strongest egocentric methods exploit it. EASG pins its graphs directly to the
PRE/PNR/POST frames~\cite{rodin2024easg}; EGO-OMG reasons over the resulting
node-state sequence to anticipate the next action~\cite{dessalene2020egoomg};
AMEGO encodes dynamics as the birth, persistence, and death of tracklets so that
sequencing and concurrency queries become operations on
lifetimes~\cite{goletto2024amego}. But notice what egocentric work does \emph{not}
yet do: it tracks how graphs change, it does not \emph{forecast} it. The
third-person side already models edge evolution as a continuous latent process and
predicts future graphs outright~\cite{peddi2024scenesayer}. Porting that capability
to HOI graphs---forecasting the next contact edge rather than detecting it after
the fact---is, in our view, one of the clearest near-term opportunities the
egocentric field is currently leaving on the table.

\subsubsection{Graph-based Interaction Modeling}
\label{sec:graph-interaction-modeling}

Constructing the graph is half the job; the model still has to reason over it, and
the evidence here directly substantiates the section's thesis. Interaction-reasoning
networks make hand--object edges first-class via a transformer unit that jointly
relates each acting hand, the other hand, and the engaged objects, and show that
this explicit two-hand--object modeling is what fine-grained egocentric recognition
actually needs~\cite{hoireasoning2022}. ORViT~\cite{herzig2022orvit} generalizes
the lesson---object-region structure belongs \emph{inside} the backbone, injected
from the earliest layers, not appended afterward---and improves EPIC-KITCHENS-100
accordingly. The sharpest evidence is STLT~\cite{radevski2021stlt}, which throws
away appearance entirely and reasons over object categories and bounding-box
layouts alone, yet still recovers most of the discriminative signal for
compositional actions. Read against Section~\ref{sec:vlm-fail}, this result provides evidence that
explicit layout dynamics carry useful information for compositional actions and
that dense appearance-based models may underuse such relational structure. This is the concrete form of our claim---dense models ask \emph{what is present}, graph models are built to ask \emph{how things
relate}, and on egocentric video the second question is the one that matters.

\subsection{Graph-guided Frame Sampling}
\label{sec:graph-sampling}

Graphs can also decide which frames a model ever processes---and on hour-long
egocentric video this is not a side issue but a hard ceiling on quality. Every
long-video VLM must discard most of its input, because accuracy degrades and
latency grows with the frame budget, so the sampler quietly bounds everything
downstream of it. The egocentric case is the worst case for naive sampling, which
is exactly why it deserves attention here.

\subsubsection{Random vs.\ Interaction-aware Sampling}
\label{sec:random-vs-interaction}

Uniform sampling---every $k$-th frame---is a common default but can be poorly
matched to the sparse and irregular interaction events found in first-person
video. The informative moments (contact onset, point of no return,
state change) are sparse and irregularly spaced, so uniform sampling spends its
budget as readily on a motion-blurred head turn as on a grasp. The first real
correction was to make sampling \emph{query-aware}: SeViLA~\cite{yu2023sevila}
turns an image-language model into a localizer that picks language-relevant
keyframes and self-supervises without moment labels, and
Frame-Voyager~\cite{yu2025framevoyager} goes further by ranking whole frame
\emph{combinations} according to the loss they induce in a downstream video-LLM,
capturing inter-frame dependencies that per-frame scoring throws away. The finding the field has now established---and that
the egocentric setting makes especially stark---is that on first-person video,
\emph{which} frames the model sees governs accuracy as much as how it processes
them, because relevance is so unevenly distributed in time. The honest caveat:
SeViLA and Frame-Voyager are general-video methods; the egocentric community
inherits them rather than having produced them.

\subsubsection{Segment-aware Sampling}
\label{sec:segment-aware}

Operating on graph-structured \emph{segments} rather than isolated frames is where
this turns principled. STEP~\cite{qiu2025step} induces a spatio-temporal scene
graph from raw video and uses it to self-generate multi-step reasoning supervision,
letting structure decide what a video-LLM learns to attend to. The cleanest
egocentric instance is FocusGraph~\cite{zemskova2026focusgraph}: a trainable
scene-caption selector first picks query-relevant \emph{clips} by reasoning over
compact, graph-based \emph{textual} captions---a structured scene description,
not a stack of low-resolution frames---after which a training-free patchwise
sparse-flow step retains keyframes within those clips. The payoff is concrete:
state-of-the-art accuracy on egocentric long-video QA (FindingDory, and the
Ego4D-sourced HourVideo) at under one token per frame and a fraction of the
latency of agent-based selectors. GraphVideoAgent~\cite{hu2025graphvideoagent}
makes the same bet with a dynamic entity-relation graph memory steering iterative
selection, with consistent EgoSchema gains. The recurring pattern is the one from
Section~\ref{sec:hoi-graphs}: interaction structure is not the output of
perception but the scaffold that organizes it---here, the scaffold that decides
where to look. We note, though, that FocusGraph is a 2026 preprint leaning on a
7B VLM to write its graph captions; the result is promising and the dependence on
a heavy frozen model is real.

\subsection{Graph-enhanced Representation Learning}
\label{sec:graph-repr}

The most durable use of a graph is not at inference at all but during
training---folding relational structure back into the learned features. The
egocentric high point here is EgoHOD~\cite{pei2025egohod}, and it is worth being
precise about why it matters: it builds a hand--object detection pipeline,
auto-generates fine-grained interaction-\emph{dynamics} narrations with a language
model, and trains a dedicated motion adapter on them, yielding zero-shot gains on
the order of $+6.3\%$ on EPIC-KITCHENS-100 multi-instance retrieval and $+16.3\%$
on EGTEA classification. The significance is not the headline numbers but what
they isolate: encoding interaction structure during pretraining improves the
\emph{representation}, not just a downstream classifier, which is the cleanest
available evidence that the structure carries information the dense objective was
discarding. The more general object-centric machinery points the same direction
without quite landing in the egocentric setting: slot-based encoders such as
SAVi~\cite{kipf2022savi} bind scene content to a few object slots, and
ObjectPrompt~\cite{zhang2024objectprompt} pulls object-centric features from a VLM
without finetuning and retrieves the relevant objects for long-term anticipation
on Ego4D and EGTEA Gaze+---complementing ORViT's in-backbone object
tokens~\cite{herzig2022orvit}. Graph-contrastive self-supervision over frame- or
region-level relational graphs~\cite{li2021weaklyhoi} offers a route to
structure-aware features without dense labels. The pattern, and the limitation, is
by now familiar: with the single exception of EgoHOD, the representation-learning
toolkit is general-video stock, validated on egocentric benchmarks rather than
designed for the first-person regime---and EgoHOD's outsized gains are the best
indication of how much is being left unclaimed by not designing for it from the
start.

\subsection{Open Challenges in Graph-based Reasoning}
\label{sec:graph-challenges}

The thesis of this section has a corollary: if egocentric graph reasoning is the
right idea executed on borrowed and immature foundations, then the open problems
are not refinements but the load-bearing work that still has to be done. Four
stand out.

\emph{Graphs do not survive ego-motion.} The detectors and trackers that populate
graph nodes were tuned on framed third-person video, and the head motion, blur,
and truncation of first-person capture (Section~\ref{sec:challenges}) corrupt node
identity and edge assignment directly. That both SAMJAM~\cite{li2025samjam} and
FocusGraph~\cite{zemskova2026focusgraph} resort to heavy foundation models or
optical-flow heuristics just to hold their graphs together across motion is the
symptom; robust, lightweight egocentric graph construction is the unmet need.

\emph{Length breaks the graph.} A graph that adds a node per object instance and an
edge per relation per frame is intractable over the minutes-to-hours horizons of
real egocentric activity. AMEGO's semantic-free tracklets~\cite{goletto2024amego}
and FocusGraph's compact textual graphs~\cite{zemskova2026focusgraph} are the only
serious answers so far, and neither offers a general account of how to compress,
forget, and re-instantiate graph structure over long horizons.

\emph{The annotations are not there.} The expressive graphs the field wants
depend on dense supervision---Action Genome's~\cite{ji2020actiongenome}, EASG's~\cite{rodin2024easg}---that is expensive in third-person video and barely exists for
egocentric. This single fact explains the recent lurch toward zero-shot,
foundation-model-driven construction, and whether weak and zero-shot supervision can match annotation-based quality remains unresolved.

\emph{The symbol--vector gap is still open.} A graph is discrete and symbolic; a
VLM reasons in continuous embeddings, and bolting one onto the other is unsolved.
HyperGLM's hypergraph-into-an-LLM~\cite{nguyen2025hyperglm} and STEP's graph-guided
self-training~\cite{qiu2025step} are the first credible attempts, but no method yet
unifies robust egocentric graph construction, long-horizon graph reasoning, and the
open-ended language interface of a VLM in one system. Closing that gap is, we argue,
the decisive step from hand--object interaction to the embodied intelligence of
Section~\ref{sec:applications}---and it is the step the egocentric field has not yet
taken.

\section{Prompting and Semantic Alignment in Vision-Language Models}
\label{sec:prompting-alignment}

The transition from task-specific egocentric models to general-purpose
vision-language models (VLMs) shifts the central engineering problem. It is
no longer ``how do we design a network for action recognition,'' but ``how do
we phrase a query, and how do we guarantee that the phrasing and the video
actually refer to the same thing.'' These two questions---how to prompt, and
how to align---are inseparable in the egocentric setting, where the visual
signal is cluttered, the camera is constantly moving, and the textual
supervision is short, noisy, and frequently ambiguous. This section traces
prompting from its hand-crafted origins to interaction-aware designs, and examines why semantic alignment, although extensively studied in third-person vision, remains particularly challenging for first-person video.

\subsection{Prompt Engineering in Vision-Language Models}
\label{sec:prompt-engineering}

Prompting begins with CLIP \cite{radford2021clip}. Because CLIP synthesizes a
classifier directly from natural language, the wording of the class
description is not cosmetic---it is the classifier. Radford \textit{et al.}
showed that the template ``a photo of a \{label\}'' and, more strikingly,
ensembling 80 context prompts improved ImageNet accuracy by several points
over a single bare label \cite{radford2021clip}. This is the uncomfortable
lesson that motivated everything after it: a model can be frozen and powerful,
yet its usable accuracy hinges on a sentence a human guessed.

Hand-tuning that sentence does not scale, so the field replaced discrete words
with learnable vectors. CoOp \cite{zhou2022coop} introduced Context
Optimization, modeling a prompt's context tokens as continuous vectors trained
end-to-end while CLIP stays frozen; with only sixteen labeled images per class
it beat carefully engineered prompts by roughly fifteen percent on average
across eleven datasets \cite{zhou2022coop}. The weakness was equally clear: a
single static prompt overfits the classes it was trained on. CoCoOp
\cite{zhou2022cocoop} answered this with a lightweight Meta-Net that generates
an \emph{instance-conditional} token per image, so the prompt adapts to what is
actually in front of the camera---a small change that substantially improved
generalization to unseen classes and domain shift. The same logic was then
pushed into the vision branch by VPT \cite{jia2022vpt}, which prepends a handful
of learnable tokens to a frozen ViT and tunes under one percent of parameters,
and into both branches jointly by MaPLe \cite{khattak2023maple}, which couples
vision and language prompts so the two modalities are adapted in step rather
than in isolation.

Video forced a further generalization, because a prompt now has to address time
as well as appearance. ActionCLIP \cite{wang2021actionclip} reframed action
recognition as a ``pre-train, prompt, fine-tune'' matching problem using
textual label prompts, and Ju \textit{et al.} \cite{ju2022prompting} showed that
optimizing a few continuous prompt vectors is enough to adapt a frozen
image-text model across recognition, localization, and retrieval. Vita-CLIP
\cite{wasim2023vitaclip} took the idea to its natural conclusion with a unified
multimodal scheme: global video-level prompts model the data distribution,
local frame-level prompts provide per-frame conditioning, and a summary prompt
condenses the clip, with text-side prompts added in parallel---letting a single
frozen-backbone model remain competitive in the supervised regime while keeping
the zero-shot capability that full fine-tuning destroys. The trajectory across
this subsection is consistent and worth stating plainly: prompting moved from
\emph{words humans choose}, to \emph{text vectors models learn}, to
\emph{visual and temporal vectors models learn jointly}. A broader catalogue of
this lineage is given in the recent VLM survey of Zhang \textit{et al.}
\cite{zhang2024vlmsurvey}.

\subsection{Video--Text Semantic Alignment}
\label{sec:video-text-alignment}

\begin{table}[H]
\centering
\caption{Egocentric video--text semantic alignment approaches
(Section~\ref{sec:video-text-alignment}) and the weakness each targets.}
\label{tab:alignment-comparison}
\footnotesize
\setlength{\tabcolsep}{4pt}
\renewcommand{\arraystretch}{1.3}
\begin{tabular}{p{1.75cm}p{2.65cm}p{2.65cm}}
\hline
\textbf{Method} & \textbf{Alignment strategy} & \textbf{Target weakness / finding} \\
\hline
EgoVLP~\cite{lin2022egovlp} & EgoNCE mines ego-aware positives/negatives & Naive InfoNCE ignores shared verb/noun semantics \\
EgoVLPv2~\cite{pramanick2023egovlpv2} & Cross-modal fusion in the backbone & Late fusion limits video--text interaction \\
LaViLa~\cite{zhao2023lavila} & LLM re-narration (Narrator + Rephraser) & Sparse, unevenly timed Ego4D narrations \\
EgoNCE++~\cite{dong2025truly} &  HOI-aware hard-negative captions & Object-over-action (verb) bias \\
EgoHOD~\cite{pei2025egohod} & HOI-dynamics narrations + motion adapter & Condensed narrations lack fine detail \\
Wray~\textit{et al.}~\cite{wray2021semantic} & Semantic (many-to-many) relevance evaluation & One-caption-one-video assumption is false \\
\hline
\end{tabular}
\end{table}

Prompting assumes the text and the video can be compared in a shared space.
Building that space is the alignment problem. Table~\ref{tab:alignment-comparison}
summarizes the main egocentric alignment approaches and the weaknesses they
target. In the general domain it is
solved by large-scale contrastive pretraining: VideoCLIP \cite{videoclip2021}
contrasts temporally overlapping video-text pairs against \emph{hard negatives}
retrieved by nearest-neighbor search, and Frozen-in-Time \cite{bain2021frozen}
trains a joint image-and-video dual encoder end-to-end for retrieval. Both work
well precisely because their text---scraped captions and ASR---is loosely but
broadly descriptive.

Egocentric video breaks that assumption, and the alignment literature is
largely a sequence of responses to how it breaks. EgoVLP \cite{lin2022egovlp}
made the first systematic attempt, contributing three pieces that are now
standard reference points: EgoClip, a 3.8M clip-text pretraining set curated
from Ego4D; EgoNCE, a contrastive objective that mines egocentric-aware
positives (clips sharing verb or noun semantics) and negatives rather than
treating every other clip as negative; and EgoMCQ, a multiple-choice benchmark
built specifically to measure video-text alignment. EgoVLPv2
\cite{pramanick2023egovlpv2} then moved cross-modal fusion out of the
fine-tuning head and into the backbone itself, so video and language interact
during pretraining instead of only at the end.

The most influential recent direction attacks the text side directly. LaViLa
\cite{zhao2023lavila} observes that Ego4D narrations are sparse and unevenly
distributed in time, and repurposes a pretrained large language model as a
visually-conditioned ``Narrator'' (with a ``Rephraser'' for diversity) to
densely re-narrate long videos before contrastive training. The payoff is
concrete---absolute gains of $10.1\%$ on EGTEA classification and $5.9\%$ on
EPIC-KITCHENS-100 multi-instance retrieval \cite{zhao2023lavila}---and it
established a template the field still follows: when the supervision is weak,
generate better supervision. Underlying all of this is a conceptual point made
sharply by Wray \textit{et al.} \cite{wray2021semantic}: the standard
instance-based assumption that one caption matches exactly one video is simply
false. Many videos can be equally valid answers to one description and vice
versa, so retrieval should be evaluated by \emph{semantic} relevance rather
than instance identity. This reframing matters far beyond evaluation, because
the many-to-many structure it describes is the root of the ambiguity discussed
next.

\subsection{Semantic Ambiguity in Egocentric Videos}
\label{sec:semantic-ambiguity}

Egocentric narrations are short, telegraphic, and written by annotators who saw
the action live; ``\#C C puts it down'' is typical. This produces three
compounding problems: temporal misalignment between narration and the moment it
describes, vocabulary inconsistency across annotators, and genuine
verb--noun polysemy where the same words cover visually distinct actions. The
result is that a model can score well by exploiting the easy, stable part of
the signal---the objects---while ignoring the hard, time-dependent part---the
interaction.

Three recent diagnostic studies converge on exactly this failure from
independent angles, and together they make the case that egocentric alignment
is not yet solved. Xu \textit{et al.} \cite{dong2025truly}, with their
EgoHOIBench benchmark, show that state-of-the-art EgoVLMs are easily flipped by
substituting a single verb or noun in an interaction description, and attribute
the gap to insufficient fine-grained supervision and a marked difficulty in
recognizing verbs relative to nouns---a quantified object-over-action bias.
Their proposed EgoNCE++, an asymmetric contrastive objective that mines
HOI-aware negative captions on the video-to-text side while preserving an
object-centric text-to-video space, improves three EgoVLMs across seven
downstream tasks. From the multimodal-LLM direction, EgoTempo
\cite{plizzari2025egotempo} demonstrates that on existing egocentric QA
benchmarks strong models score ``remarkably high'' using only the text of the
question or a single frame---direct evidence that they lean on language and
object priors rather than temporal content---and introduces a benchmark whose
questions cannot be answered without integrating the whole video. EgoSchema~\cite{mangalam2023egoschema} formalizes the long-form version
of the same problem through its notion of a \emph{temporal certificate}:
its questions have a median certificate near 100 seconds. In the original
evaluation, the tested billion-parameter models scored below $33\%$,
whereas humans reached approximately $76\%$.
 Taken together, these results indicate that current VLM predictions can rely
heavily on object identity and language priors rather than on fine-grained
interaction and temporal understanding.

\subsection{Interaction-aware Prompting and Conditioning}
\label{sec:interaction-aware-prompting}

If the diagnosis is that models attend to objects and underuse interactions, the
remedy is to make interaction structure an explicit conditioning signal.
Interaction-aware prompting and conditioning inject hands, contact, and
manipulated-object structure into the visual input, textual supervision, or
auxiliary training objectives, rather than relying on a generic prompt to
recover it.

The clearest instance on the supervision side is Helping Hands
\cite{zhang2023helpinghands}, which attaches an object-aware decoder that is
trained to predict hand positions, manipulated-object positions, and object
labels from the paired caption, while remaining RGB-only at inference; this
explicit interaction signal improves zero-shot egocentric retrieval and
classification. EgoHOD \cite{pei2025egohod} pushes the same idea into the text
itself: it couples a hand-object detector with an LLM to generate narrations
that explicitly describe fine-grained hand-object dynamics---rather than the
heavily condensed original narrations---and trains the EgoVideo model with a
lightweight motion adapter on top, reporting zero-shot gains of $6.3\%$ on
EK-100 multi-instance retrieval and $16.3\%$ on EGTEA classification
\cite{pei2025egohod}. On the visual-conditioning side, POV \cite{xu2023pov} uses
frame-level interactive masking prompts that steer the model toward the
hand-object region to transfer from third- to first-person views, and
action-conditioned prompting \cite{jia2024actionprompt} uses an LLM to expand
each action into a descriptive, structured text prompt and aligns those
knowledge concepts with the video for open-vocabulary recognition. A related
line grounds detected objects through a VLM as a reasoning anchor for
discovering novel actions \cite{kundu2024discovering}, and HOI-Ref
\cite{bansal2024hoiref} adapts VLM prompts to refer explicitly to hands and
objects in egocentric images.

The common thread is a reversal of the default assumption. Standard prompting
asks the model to find the interaction; interaction-aware prompting tells the
model where and what the interaction is, and lets it reason from there. Given
the evidence in Section~\ref{sec:semantic-ambiguity} that the interaction is
exactly what generic models miss, this is not a refinement but a necessary
correction---and, as the embodied applications in later sections demand
faithful modeling of manipulation rather than scene gist, it is likely to
become the dominant prompting paradigm for first-person video.

\section{Applications Toward Embodied Intelligence}
\label{sec:applications}

The methods surveyed so far---hand--object reasoning, spatial--temporal
modeling, graph-based representations, and semantic alignment---are not ends in
themselves. Their value is ultimately measured by what they enable an embodied
agent to \emph{do}: assist a person in real time, understand a skilled
activity, run on a wearable device, or transfer a manipulation skill to a robot.
A striking convergence has emerged across these goals. The same first-person
sensing stack---Ego4D~\cite{grauman2022ego4d},
Ego-Exo4D~\cite{grauman2024egoexo4d}, and Project Aria smart
glasses~\cite{engel2023projectaria}---now feeds both human-facing assistants and
robot-learning pipelines. This shared data substrate lets us view egocentric
perception and embodied action as a single pipeline (capture
$\rightarrow$ understanding $\rightarrow$ assistance or policy) rather than four
disconnected application areas. The recurring obstacle that binds them is the
gap between what a head-mounted camera observes a human hand doing and what a
downstream agent---a robot gripper, or an assistant inferring intent---can act
on. We organize this section around that pipeline and return to the gap in
Section~\ref{sec:embodied-challenges}.

\subsection{Assistive Systems}
\label{sec:assistive}

The most direct application of egocentric VLMs is the real-time assistant: a
system that watches the user's first-person view and offers timely guidance.
Vinci~\cite{huang2024vinci} is a representative real-time assistant prototype,
built on an egocentric video-language backbone and designed for portable
devices. It processes a live stream, supports hands-free audio interaction, maintains a
memory module for historical context, and can synthesize step-by-step how-to
demonstrations. EgoLife~\cite{yang2025egolife} pushes the assistant
toward week-scale, always-on use: six participants wore Aria glasses for a week,
yielding a 300-hour multimodal dataset and the EgoLifeQA benchmark for
long-context queries such as event recall and habit insight. Its EgoButler
system couples an omni-modal VLM with retrieval-augmented memory (EgoRAG) to
answer questions over ultra-long histories---a concrete blueprint for cognitive
and accessibility assistance.

A second, complementary thread treats assistance as \emph{error correction}
rather than narration. HoloAssist~\cite{wang2023holoassist} captures
$166$ hours of two-party interaction---a remote instructor watching a
performer's live egocentric feed and intervening verbally---over $350$
instructor--performer pairs and $20$ manipulation tasks, with annotations for
mistakes and intervention type. This distinguishes passive guidance from
\emph{active} coaching, an important axis for AR tutors. The 2024 wave of
procedural mistake-detection work sharpened the problem.
PREGO~\cite{flaborea2024prego} formulates \emph{online}, open-set
mistake detection by comparing the recognized current action against an
anticipated one, introducing the Assembly101-O and Epic-tent-O benchmarks.
EgoPER~\cite{lee2024egoper} contributes a richer error taxonomy
(omission, addition, modification, slip, correction) grounded in task graphs,
while CaptainCook4D~\cite{peddi2024captaincook4d} provides recipe-execution
videos with deliberately induced errors. Together these define what a
``mistake'' formally is and expose how brittle current models remain on
as-it-happens detection---the capability an interactive assistant most needs.

\subsection{Human Activity Understanding}
\label{sec:activity}

Assistance presupposes understanding not just \emph{what} a person does but \emph{how well}. Ego-Exo4D~\cite{grauman2024egoexo4d} is the foundational resource, pairing first- and third-person skilled-activity video with expert commentary and benchmarks for keystep recognition, proficiency estimation, and 3D ego-pose; its proficiency task---ranking novice-to-expert and localizing ``good'' actions and ``tips''---is markedly harder than action labels, and submissions remain short of human agreement. Procedural understanding has deeper roots in Assembly101~\cite{sener2022assembly101} and AssemblyHands~\cite{ohkawa2023assemblyhands}, and a recurring finding is that \emph{where} the model looks matters: a stabilized hand-centric crop yields $+22\%$ Top-1 on Ego-Exo4D keystep recognition~\cite{chavis2025dexfocus}, reinforcing the survey-wide theme that hands are the most informative region.

\subsection{AR/VR and Wearable AI}
\label{sec:arvr}

Deploying egocentric AI capabilities depends on wearable platforms that combine
multimodal sensing with efficient machine-perception tools. Project
Aria~\cite{engel2023projectaria} provides a research platform for capturing
multimodal egocentric data and supports associated perception capabilities.
Aria Gen 2~\cite{kong2025ariagen2} extends this platform with richer sensing,
on-device computation, and real-time and offline perception outputs, including
SLAM and hand tracking. On the perception side,
HOT3D~\cite{banerjee2024hot3d} demonstrates that multi-view egocentric capture
can improve 3D hand and 6-DoF object tracking and support transfer across
different headset configurations.

Despite these advances, wearable deployment exposes a substantial reasoning and
efficiency gap. In the original EgoSchema
evaluation~\cite{mangalam2023egoschema}, the tested billion-parameter models
scored below $33\%$, whereas humans achieved approximately $76\%$. Closing
long-horizon reasoning gaps under the limited power, memory, and latency budgets
of wearable devices, rather than relying on server-scale computation, therefore
remains a central challenge for always-on AR/VR assistants.

\subsection{Human--Robot Collaboration}
\label{sec:hrc}

The most ambitious use of egocentric video is teaching robots to manipulate by
watching people. This line is best read as a spectrum defined by how much
\emph{robot} data each method requires.

Representation transfer sits at one end.
R3M~\cite{nair2022r3m} pretrains a visual representation on Ego4D via
time-contrastive and video--language objectives, improving downstream
manipulation success by over $20\%$ relative to training from scratch and over
$10\%$ relative to CLIP/MoCo features---the seminal evidence that human
egocentric video yields transferable manipulation priors.

Generalist vision--language--action (VLA) models occupy the
robot-data-heavy middle. RT-2~\cite{zitkovich2023rt2} introduced the VLA
formulation, representing actions as text tokens and co-training a VLM on robot
trajectories and web data; OpenVLA~\cite{kim2024openvla} made the
approach open and efficient, with a $7$B model trained on $970$K demonstrations
that surpasses the $55$B RT-2-X by $16.5\%$ absolute success across $29$ tasks;
and $\pi_0$~\cite{black2024pi0} pairs a VLM with a flow-matching action
expert for high-frequency dexterous control. These set the baselines against
which egocentric-trained policies are measured.

Egocentric-driven policies reduce the robot-data requirement by treating
human first-person video as a first-class training source.
EgoMimic~\cite{kareer2024egomimic} co-trains a single policy on Aria-captured
human demonstrations and robot demonstrations, using a low-cost manipulator that
minimizes the kinematic gap to human hands. EgoVLA~\cite{villa2025egovla}
trains a VLA on egocentric human video in a unified action space (wrist pose plus
MANO hand parameters), converting human actions to robot actions through inverse
kinematics and retargeting, then fine-tuning on a few robot demonstrations. At
the far end, EgoZero~\cite{liu2025egozero} learns from Aria glasses with
\emph{zero} robot data, reporting $70\%$ zero-shot success across seven tasks
with only twenty minutes of data collection per task via a morphology-agnostic
state representation. Where human demonstrations are scarce,
DexMimicGen~\cite{jiang2025dexmimicgen} amplifies them, synthesizing
$21$K bimanual trajectories from $60$ source demonstrations. Read together, these
works trace a clear trajectory toward learning dexterous manipulation primarily
from human video, with robot data as a thin alignment layer.

\subsection{Open Challenges for Embodied Deployment}
\label{sec:embodied-challenges}

Despite rapid progress, several obstacles recur across all four application
areas. The foremost is the embodiment gap: a human hand and a robot
gripper differ in morphology, degrees of freedom, and contact dynamics, so even
perfectly understood human demonstrations do not map cleanly onto robot actions.
Closely related is the visual domain gap---differences in camera height,
viewpoint, and field of view between a head-mounted human perspective and a
robot's sensors---which limits naive representation transfer. For real-time
assistants and on-device wearables, latency and compute remain hard
constraints: the long-form reasoning gap exposed by
EgoSchema~\cite{mangalam2023egoschema} must be closed within the power budget of
glasses, not a datacenter. Skewed and subjective labels, evident in the
proficiency setting of Ego-Exo4D~\cite{grauman2024egoexo4d}, complicate both
training and evaluation of skill-aware systems. Finally, always-on egocentric
capture raises privacy concerns that the field's own forward-looking
analyses flag as a precondition for real-world adoption~\cite{plizzari2024outlook}.
Progress on embodied intelligence will depend less on any single benchmark than
on jointly addressing this cluster of perception, transfer, efficiency, and
trust problems.

\section{Open Challenges and Future Directions}
\label{sec:open-challenges}

Every section of this survey has ended at the same wall. Section~\ref{sec:vlm}
showed that the strongest egocentric vision-language models recognize
\emph{what} is in the frame but not \emph{how} the hands act on it;
Section~\ref{sec:spatiotemporal} showed that they exploit spatial shortcuts in
place of genuine temporal reasoning; Section~\ref{sec:graph} showed that the
structural fix---graphs---is still built on borrowed, third-person foundations;
and Section~\ref{sec:applications} showed that even when perception works, the
gap to embodied action remains open. These are not six unrelated complaints.
They are facets of a single deficit: current models perceive first-person video
as a bag of frames and objects, when the signal is fundamentally about
\emph{relations unfolding in time}. This section lays out where the field must
go, organized so that each challenge names both the diagnosed failure and the
most promising line of attack. We are deliberately specific about what is solved,
what is merely patched, and what remains genuinely open. Table~\ref{tab:challenges}
summarizes the six challenges below, each pairing a diagnosed failure with its
supporting evidence and the most promising direction.

\begin{table}[H]
\centering
\caption{Summary of open challenges and future directions
(Section~\ref{sec:open-challenges}). Each row pairs a diagnosed failure with its
supporting evidence and the most promising line of attack.}
\label{tab:challenges}
\footnotesize
\renewcommand{\arraystretch}{1.35}
\resizebox{\linewidth}{!}{%
\begin{tabular}{p{2.5cm}p{3.5cm}p{4.3cm}p{4.2cm}}
\hline
\textbf{Challenge} & \textbf{Diagnosed failure} & \textbf{Evidence} & \textbf{Promising direction} \\
\hline
Temporal reasoning (Sec.~\ref{sec:future-temporal}) & Models exploit spatial/appearance shortcuts rather than order & Text-only 31.3\% and single-frame $\sim$51\% on EgoSchema; single-frame collapses to 9.1\% on EgoTempo~\cite{plizzari2025egotempo} & TGPO lifts a 3B model to 49.7\% on
EgoSchema~\cite{xu2026temporal}; EgoVLM explores complementary
GRPO-based policy optimization~\cite{vinod2025egovlm} \\
Interaction understanding (Sec.~\ref{sec:future-interaction}) & Verb--noun asymmetry: nouns recognized, verbs missed & LaViLa scores 74.33\% on nouns vs.\ 46.61\% on verbs (EgoHOIBench)~\cite{dong2025truly} & Generate interaction supervision: EgoNCE++ ($+34.02$ percentage points in LaViLa++ verb accuracy) and EgoHOD~\cite{dong2025truly,pei2025egohod} \\
Graph-enhanced VLMs (Sec.~\ref{sec:future-graph}) & Dense tokens store no explicit relation; symbol--vector gap & Layout-only STLT still recovers the verb without appearance~\cite{radevski2021stlt} & Hypergraph-into-LLM, graph-guided self-training, scene-graph forecasting~\cite{nguyen2025hyperglm,qiu2025step,peddi2024scenesayer} \\
Efficient frame sampling (Sec.~\ref{sec:future-sampling}) & Uniform sampling wastes budget on uninformative frames & Frame selection substantially affects long-video VLM accuracy and
efficiency~\cite{yu2023sevila,yu2025framevoyager,zemskova2026focusgraph} & Query- and structure-aware selection: SeViLA, Frame-Voyager, FocusGraph~\cite{yu2023sevila,yu2025framevoyager,zemskova2026focusgraph} \\
Multimodal learning (Sec.~\ref{sec:future-multimodal}) & Video-only models ignore IMU, gaze, audio, and SLAM & IMU signals provide motion cues when visual frames are degraded by
ego-motion~\cite{tan2023egodistill} & Adaptive fusion: ImageBind, EgoDistill, EgoLife~\cite{girdhar2023imagebind,tan2023egodistill,yang2025egolife} \\
Embodied deployment (Sec.~\ref{sec:future-embodied}) & Embodiment and visual domain gaps; on-device compute limits & Original EgoSchema evaluation: tested models $<$33\% vs.\
$\sim$76\% human~\cite{mangalam2023egoschema}; wearable deployment additionally imposes strict compute
constraints~\cite{engel2023projectaria} & Egocentric world models; privacy and trust benchmarks~\cite{tu2025playerone,bai2025wholebody,egoillusion2025} \\
\hline
\end{tabular}
}
\end{table}

\subsection{Better Temporal Reasoning}
\label{sec:future-temporal}

One of the clearest diagnostic results is that a language model with no video
input scores $31.3\%$ on EgoSchema, while a model given one frame reaches about
$51\%$~\cite{plizzari2025egotempo}. These findings indicate that EgoSchema
permits substantial performance through non-temporal shortcuts and that leading
models may rely less on temporal integration than their aggregate scores
suggest.
EgoTempo~\cite{plizzari2025egotempo} makes the point unavoidable: the same
single-frame model that reaches $51\%$ on EgoSchema collapses to $9.1\%$ on
questions constructed to be unanswerable without integrating the whole video.
EgoSchema~\cite{mangalam2023egoschema} had already quantified the horizon with
its temporal-certificate sets---a median near $100$ seconds, far longer than any
prior dataset---and the sub-$33\%$ accuracy of the models tested in the original
EgoSchema evaluation against roughly $76\%$ for humans is the size of the gap that remains.

The promising result is that objective design appears to be an important factor
beyond model scale. The
clearest evidence is that order-aware reinforcement learning, which contrasts a
model's outputs on temporally ordered versus shuffled frames to reward
temporally coherent reasoning, lifts a $3$-billion-parameter backbone to $49.7\%$
on EgoSchema---outperforming far larger baselines~\cite{xu2026temporal}. The
lineage here, from Shuffle-and-Learn's 2016 pretext task to a 2026
reinforcement-learning reward (Section~\ref{sec:temporal-ordering}), is the
clearest signal that order-awareness is the live frontier rather than a solved
detail. We expect the next advances to come from verifiable-reward
post-training---GRPO-style optimization applied directly to egocentric
understanding, as early work without any supervised fine-tuning already
suggests~\cite{vinod2025egovlm}---coupled with the long-context memory
mechanisms of Section~\ref{sec:long-term}. The open question is no longer
\emph{whether} models can be made to reason over time, but whether
certificate-long horizons can be reached under the compute budget of a wearable
rather than a server.

\subsection{Interaction-aware Video Understanding}
\label{sec:future-interaction}

If temporal reasoning is the survey's headline failure, the verb--noun asymmetry
is its root cause. EgoHOIBench~\cite{dong2025truly} reduced the problem to a
single, reproducible number: substitute one verb in a caption and
state-of-the-art egocentric models trained on millions of interaction clips
flip their answer, with LaViLa scoring $74.33\%$ on nouns but only $46.61\%$ on
verbs. A model that knows the onion but not the cutting has not understood the
interaction; it has recognized the scene. This is exactly the deficit that
defines first-person video, because an action there is a relation between hand
and object, not an appearance (Section~\ref{sec:why-hands}).

The corrective is to make the interaction an explicit training signal rather
than something the model is left to infer. The two most convincing
demonstrations both work by generating the supervision that scraped narrations
lack: the EgoNCE++ objective mines hard-negative captions that alter the verb,
raising the LaViLa++ variant's verb accuracy by 34.02 percentage points ~\cite{dong2025truly}, while EgoHOD
auto-generates fine-grained hand--object dynamics narrations and trains a
dedicated motion adapter on them, yielding zero-shot gains of roughly $+6.3\%$
on EPIC-KITCHENS-100 multi-instance retrieval and $+16.3\%$ on
EGTEA~\cite{pei2025egohod}. The principle is now established---\emph{when the
supervision is weak, generate better supervision}---and it is the template
interaction-aware prompting (Section~\ref{sec:interaction-aware-prompting})
extends. The open challenge is scale: these gains depend on detector-driven or
LLM-generated annotations whose fidelity is itself limited by the egocentric
failure modes of Section~\ref{sec:hoi-limitations}. Closing the verb gap at
Ego4D scale, without a dense-annotation crutch, is the work that remains.

\subsection{Graph-enhanced Vision-Language Models}
\label{sec:future-graph}

Section~\ref{sec:graph} argued that graphs are a promising response to the
noun-shortcut bias because they represent interactions and active objects as
explicit relational elements, potentially reducing reliance on appearance
shortcuts. The egocentric setting is particularly suitable for this approach
because its entity set is relatively small, stable, and recurrent across
minutes-long activities. Supporting evidence comes from models that reason over
object categories and bounding-box layouts without appearance features while
retaining substantial discriminative information for compositional
actions~\cite{radevski2021stlt}. This suggests that explicit layout dynamics
capture relational information that dense models may otherwise underuse.

Three problems stand between this promise and a working system. First,
\emph{the symbol--vector gap}: a graph is discrete and symbolic, a VLM reasons
in continuous embeddings, and uniting them is unsolved. The first credible
attempts inject a hypergraph of higher-order interactions into an
LLM~\cite{nguyen2025hyperglm} and use a spatio-temporal scene graph to
self-generate multi-step reasoning supervision~\cite{qiu2025step}, but neither
unifies robust egocentric graph construction with the open-ended language
interface of a VLM. Second, \emph{graphs do not survive ego-motion}: the
detectors that populate nodes were tuned on stable third-person footage, and the
fact that the cleanest egocentric instances lean on frozen foundation models
just to hold their graphs together across head
motion~\cite{li2025samjam,zemskova2026focusgraph} is the symptom of a missing
native solution. Third, \emph{the field tracks graph change but does not yet
forecast it}---the third-person literature already models edge evolution as a
continuous latent process and predicts future graphs
outright~\cite{peddi2024scenesayer}, and porting that capability to hand--object
graphs, forecasting the next contact edge rather than detecting it after the
fact, is the clearest near-term opportunity the egocentric field is leaving on
the table.

\subsection{Efficient Frame Sampling}
\label{sec:future-sampling}

On hour-long egocentric video, the sampler is not a preprocessing detail---it is
a hard ceiling on quality, because every long-video model must discard most of
its input, and uniform sampling spends its budget as readily on a motion-blurred
head turn as on a grasp. The informative moments of first-person activity,
contact onset and state change, are sparse and irregularly spaced
(Section~\ref{sec:fine-grained-hoi}), which makes the egocentric case the worst
case for naive sampling and the most rewarding for a learned alternative. The
field has established that on first-person video, \emph{which} frames the model
sees governs accuracy as much as how it processes them.

The trajectory is from query-aware to structure-aware selection. SeViLA turns an
image-language model into a keyframe localizer~\cite{yu2023sevila}, and
Frame-Voyager ranks whole frame combinations by the loss they induce
downstream~\cite{yu2025framevoyager}; the more recent move is to let interaction
structure decide where to look, as in the graph-caption-driven clip selection of
FocusGraph, which reaches state-of-the-art egocentric long-video QA at under one
token per frame~\cite{zemskova2026focusgraph}. The recurring pattern is the one
from Section~\ref{sec:interaction-aware}: HOI structure is not the output of
perception but the scaffold that organizes it---here, the scaffold that decides
what is worth attending to. The honest caveat is that the strongest selectors
are still general-video methods the egocentric community inherits, and the most
capable egocentric instance is a 2026 preprint leaning on a heavy frozen
VLM~\cite{zemskova2026focusgraph}. A lightweight, native, interaction-aware
sampler that runs within a wearable's budget does not yet exist, and it is a
prerequisite for everything downstream of it.

\subsection{Multimodal Egocentric Learning}
\label{sec:future-multimodal}

A head-mounted device is not a camera; it is a sensor suite. The wearable
platforms that now anchor the field---Project Aria~\cite{engel2023projectaria}
and its on-device successor~\cite{kong2025ariagen2}---stream RGB
alongside eye gaze, inertial measurements, audio, and SLAM, yet the dominant
models still treat egocentric understanding as a video-only problem and leave
most of that signal unused. This is a missed opportunity precisely where the
visual channel is weakest: inertial signals remain available when ego-motion degrades dense visual
features, and the very head motion that blurs a frame is itself an
attention signal about where the wearer is looking.

The foundations for fusing these streams are in place. ImageBind binds image,
text, audio, depth, thermal, and inertial measurements into one embedding space
using only image-paired data, enabling emergent cross-modal retrieval directly
applicable to egocentric rigs~\cite{girdhar2023imagebind}, and EgoDistill turns
the head-motion signal into an efficiency lever, reconstructing heavy video
features from sparse frames plus lightweight inertial readings at roughly $200\times$
fewer GFLOPs~\cite{tan2023egodistill}. The frontier is week-scale, always-on,
genuinely multimodal assistance: EgoLife wires six participants into Aria
glasses for a week and couples an omni-modal model with retrieval-augmented
memory to answer long-context queries over the resulting $300$-hour
stream~\cite{yang2025egolife}. The open problem is no longer whether the
modalities can be aligned, but how to fuse them adaptively---trusting the IMU
when the camera is blurred, the gaze when the scene is cluttered---under
real-time, on-device constraints rather than offline on a server.

\subsection{Toward Embodied Intelligence}
\label{sec:future-embodied}

Every thread in this survey converges on a single ambition: teaching machines to
act by watching people act. Section~\ref{sec:applications} traced the spectrum
of how much robot data this requires, from R3M's Ego4D-pretrained
representations that lift manipulation success by over
$20\%$~\cite{nair2022r3m}, through the robot-data-heavy generalist VLAs where
OpenVLA's $7$B model surpasses the $55$B RT-2-X by $16.5\%$ across $29$
tasks~\cite{kim2024openvla} and $\pi_0$ adds a flow-matching action expert for
dexterous control~\cite{black2024pi0}, to the egocentric-driven policies that
treat human first-person video as a first-class training source. The endpoint of
that spectrum is striking: EgoZero learns from smart glasses with \emph{zero}
robot data, reporting $70\%$ zero-shot success across seven manipulation
tasks~\cite{liu2025egozero}, while EgoMimic~\cite{kareer2024egomimic} and
EgoVLA~\cite{villa2025egovla} show that egocentric human demonstrations can
complement robot data and reduce the amount of robot-specific supervision
required.

What stands in the way is the cluster of gaps named in
Section~\ref{sec:embodied-challenges}, and they are sobering precisely because
they are not perceptual. The \emph{embodiment gap}---a human hand and a robot
gripper differ in morphology, degrees of freedom, and contact dynamics---means
even perfectly understood demonstrations do not map cleanly onto actions. The
\emph{visual domain gap} between a head-mounted human view and a robot's sensors
limits naive transfer. And the long-form reasoning gap exposed by
EgoSchema~\cite{mangalam2023egoschema} must be closed within the power budget of
glasses, not a datacenter. Two directions look most likely to move the field.
The first is \emph{egocentric world models}: simulators that predict future
first-person observations from action, which turn passive video into a
controllable environment for planning---an idea moving quickly from concept to
realistic generation~\cite{tu2025playerone,bai2025wholebody}. The second is
\emph{reliability and trust}: as always-on capture moves toward real deployment,
the privacy concerns that the field's own forward-looking analyses flag as a
precondition for adoption~\cite{plizzari2024outlook} become as load-bearing as
accuracy, and emerging benchmarks that probe egocentric hallucination begin to
measure whether a model's confident answer can be believed at
all~\cite{egoillusion2025}. Progress toward embodied intelligence will be
measured less by any single leaderboard than by jointly closing this cluster of
perception, transfer, efficiency, and trust problems---the final step on the
path from hand--object interaction to embodied AI that has organized this entire
survey.

\section{conclusion}
\label{sec:conclusion}
Egocentric video provides a distinctive window into human attention, object manipulation, and purposeful interaction with the surrounding environment. This survey has traced the development of egocentric video understanding from conventional recognition architectures to contemporary vision--language models, covering the field's principal tasks, datasets, perceptual challenges, methodological advances, and emerging applications. Particular attention has been given to hand--object interaction, spatiotemporal reasoning, frame and clip selection, multimodal learning, prompting, and semantic alignment. Within this broader landscape, graph-based and object-centric approaches have been examined as promising means of explicitly representing the spatial, temporal, and semantic relations that characterize first-person activities.

Across the reviewed literature, a consistent limitation emerges: current models generally identify visible objects more reliably than they understand actions, interaction stages, and evolving user intent. Their apparent competence can therefore depend heavily on appearance-based shortcuts or isolated frames rather than genuine integration of relations unfolding over time. Vision--language pretraining has substantially expanded the semantic capabilities of egocentric systems, but language supervision and model scale alone are unlikely to resolve these weaknesses. Further progress will require temporally grounded objectives, interaction-aware supervision, efficient and informative frame selection, adaptive multimodal fusion, and representations that preserve explicit relational structure. Graph-based reasoning is especially valuable in this regard, not as a replacement for dense video--language representations, but as a complementary mechanism for organizing hands, objects, actions, and scene context.

Future advances must also be evaluated beyond aggregate benchmark performance. More rigorous assessment of temporal reasoning, compositional generalization, cross-domain transfer, computational efficiency, privacy, and reliability will be essential for real-world deployment. Overall, the field is moving from recognizing the contents of first-person video toward understanding how interactions develop and how such understanding can support assistance and embodied action. Achieving this transition will depend on models that interpret egocentric experience robustly, efficiently, and trustworthily, thereby strengthening the connection between human activity understanding and embodied artificial intelligence.

\section*{Declaration of generative AI and AI-assisted technologies in the manuscript preparation process}

During the preparation of this work, the authors used OpenAI's ChatGPT to assist with language editing and improve the clarity and readability of the manuscript. After using this tool, the authors reviewed and edited the content as needed and take full responsibility for the content of the publication.

\subsection*{Funding}
The authors received no specific funding for this work.

\subsection*{Conflicts of Interest}
The authors declare that they have no conflicts of interest.

\subsection*{Data Availability}
No new data were created or analyzed in this review article. Data sharing is
not applicable to this article.

\printbibliography

@inproceedings{grauman2022ego4d,
  author = {K. Grauman and A. Westbury and E. Byrne and Z. Chavis and A. Furnari and R. Girdhar and J. Hamburger and H. Jiang and M. Liu and X. Liu and M. Martin and T. Nagarajan and I. Radosavovic and S. K. Ramakrishnan and F. Ryan and others},
  title = {Ego4D: Around the world in 3,000 hours of egocentric video},
  booktitle = {Proc. IEEE/CVF Conf. Comput. Vis. Pattern Recognit. (CVPR)},
  pages = {18995--19012},
  month = jun,
  year = {2022},
}

@article{damen2022rescaling,
  author = {D. Damen and H. Doughty and G. M. Farinella and A. Furnari and J. Ma and E. Kazakos and D. Moltisanti and J. Munro and T. Perrett and W. Price and M. Wray},
  title = {Rescaling egocentric vision: Collection, pipeline and challenges for EPIC-KITCHENS-100},
  journal = {Int. J. Comput. Vis. (IJCV)},
  volume = {130},
  number = {1},
  pages = {33--55},
  month = jan,
  year = {2022},
}

@inproceedings{grauman2024egoexo4d,
  author = {K. Grauman and others},
  title = {Ego-Exo4D: Understanding skilled human activity from first- and third-person perspectives},
  booktitle = {Proc. IEEE/CVF Conf. Comput. Vis. Pattern Recognit. (CVPR)},
  month = jun,
  year = {2024},
}

@inproceedings{radford2021clip,
  author = {A. Radford and J. W. Kim and C. Hallacy and A. Ramesh and G. Goh and S. Agarwal and G. Sastry and A. Askell and P. Mishkin and J. Clark and G. Krueger and I. Sutskever},
  title = {Learning transferable visual models from natural language supervision},
  booktitle = {Proc. Int. Conf. Mach. Learn. (ICML)},
  pages = {8748--8763},
  year = {2021},
}

@inproceedings{lin2022egovlp,
  author = {K. Q. Lin and A. J. Wang and M. Soldan and M. Wray and R. Yan and E. Z. Xu and D. Gao and R. Tu and W. Zhao and W. Kong and C. Cai and H. Wang and D. Damen and B. Ghanem and W. Liu and M. Z. Shou},
  title = {Egocentric video-language pretraining},
  booktitle = {Adv. Neural Inf. Process. Syst. (NeurIPS)},
  volume = {35},
  pages = {7575--7586},
  year = {2022},
}

@article{plizzari2024outlook,
  author = {C. Plizzari and others},
  title = {An outlook into the future of egocentric vision},
  journal = {Int. J. Comput. Vis.},
  volume = {132},
  pages = {4880--4936},
  year = {2024},
}

@misc{huang2024vinci,
  author = {Y. Huang and others},
  title = {Vinci: A real-time embodied smart assistant based on egocentric vision-language model},
  year = {2024},
  eprint = {2412.21080},
  eprinttype = {arxiv},
}

@misc{villa2025egovla,
  author = {R. Yang and others},
  title = {EgoVLA: Learning vision-language-action models from egocentric human videos},
  year = {2025},
  eprint = {2507.12440},
  eprinttype = {arxiv},
}

@inproceedings{dong2025truly,
  author = {B. Xu and Z. Wang and Y. Du and Z. Song and S. Zheng and Q. Jin},
  title = {Do egocentric video-language models truly understand hand-object interactions?},
  booktitle = {Proc. Int. Conf. Learn. Represent. (ICLR)},
  year = {2025},
}

@misc{xu2026temporal,
  author = {Z. Xu and T. Qin and B. Jin and Z. Lai and M. Cao and L. Huang and P. Zhang},
  title = {Incentivizing temporal-awareness in egocentric video understanding models},
  year = {2026},
  eprint = {2603.27184},
  eprinttype = {arxiv},
}

@inproceedings{kareer2024egomimic,
  author = {S. Kareer and D. Patel and R. Punamiya and P. Mathur and S. Cheng and C. Wang and J. Hoffman and D. Xu},
  title = {EgoMimic: Scaling imitation learning via egocentric video},
  booktitle = {Proc. IEEE Int. Conf. Robot. Autom. (ICRA)},
  year = {2025},
}

@article{li2025challenges,
  author = {X. Li and H. Qiu and L. Wang and H. Zhang and C. Qi and L. Han and H. Xiong and H. Li},
  title = {Challenges and trends in egocentric vision: A survey},
  journal = {Mach. Intell. Res.},
  volume = {23},
  number = {1},
  pages = {1--33},
  year = {2026},
}

@article{kanade2012firstperson,
  author = {T. Kanade and M. Hebert},
  title = {First-person vision},
  journal = {Proc. IEEE},
  volume = {100},
  number = {8},
  pages = {2442--2453},
  month = aug,
  year = {2012},
}

@article{rodin2021predicting,
  author = {I. Rodin and A. Furnari and D. Mavroeidis and G. M. Farinella},
  title = {Predicting the future from first person (egocentric) vision: A survey},
  journal = {Comput. Vis. Image Underst. (CVIU)},
  volume = {211},
  pages = {103252},
  year = {2021},
}

@article{bandini2023analysis,
  author = {A. Bandini and J. Zariffa},
  title = {Analysis of the hands in egocentric vision: A survey},
  journal = {IEEE Trans. Pattern Anal. Mach. Intell. (TPAMI)},
  volume = {45},
  number = {6},
  pages = {6846--6866},
  month = jun,
  year = {2023},
}

@misc{engel2023projectaria,
  author = {J. Engel and others},
  title = {Project Aria: A new tool for egocentric multi-modal AI research},
  year = {2023},
  eprint = {2308.13561},
  eprinttype = {arxiv},
}

@inproceedings{wang2016tsn,
  author = {L. Wang and Y. Xiong and Z. Wang and Y. Qiao and D. Lin and X. Tang and L. Van Gool},
  title = {Temporal segment networks: Towards good practices for deep action recognition},
  booktitle = {Proc. Eur. Conf. Comput. Vis. (ECCV)},
  pages = {20--36},
  year = {2016},
}

@inproceedings{carreira2017i3d,
  author = {J. Carreira and A. Zisserman},
  title = {Quo vadis, action recognition? A new model and the Kinetics dataset},
  booktitle = {Proc. IEEE Conf. Comput. Vis. Pattern Recognit. (CVPR)},
  pages = {6299--6308},
  year = {2017},
}

@inproceedings{feichtenhofer2019slowfast,
  author = {C. Feichtenhofer and H. Fan and J. Malik and K. He},
  title = {SlowFast networks for video recognition},
  booktitle = {Proc. IEEE/CVF Int. Conf. Comput. Vis. (ICCV)},
  pages = {6202--6211},
  year = {2019},
}

@inproceedings{fan2021mvit,
  author = {H. Fan and B. Xiong and K. Mangalam and Y. Li and Z. Yan and J. Malik and C. Feichtenhofer},
  title = {Multiscale vision transformers},
  booktitle = {Proc. IEEE/CVF Int. Conf. Comput. Vis. (ICCV)},
  pages = {6824--6835},
  year = {2021},
}

@inproceedings{liu2022videoswin,
  author = {Z. Liu and J. Ning and Y. Cao and Y. Wei and Z. Zhang and S. Lin and H. Hu},
  title = {Video Swin Transformer},
  booktitle = {Proc. IEEE/CVF Conf. Comput. Vis. Pattern Recognit. (CVPR)},
  pages = {3202--3211},
  year = {2022},
}

@article{furnari2020rulstm,
  author = {A. Furnari and G. M. Farinella},
  title = {Rolling-unrolling LSTMs for action anticipation from first-person video},
  journal = {IEEE Trans. Pattern Anal. Mach. Intell. (TPAMI)},
  volume = {43},
  number = {11},
  pages = {4021--4036},
  month = nov,
  year = {2021},
}

@inproceedings{girdhar2021avt,
  author = {R. Girdhar and K. Grauman},
  title = {Anticipative video transformer},
  booktitle = {Proc. IEEE/CVF Int. Conf. Comput. Vis. (ICCV)},
  pages = {13505--13515},
  year = {2021},
}

@inproceedings{krishna2017densecap,
  author = {R. Krishna and K. Hata and F. Ren and L. Fei-Fei and J. C. Niebles},
  title = {Dense-captioning events in videos},
  booktitle = {Proc. IEEE Int. Conf. Comput. Vis. (ICCV)},
  pages = {706--715},
  year = {2017},
}

@inproceedings{yang2023vid2seq,
  author = {A. Yang and A. Nagrani and P. H. Seo and A. Miech and J. Pont-Tuset and I. Laptev and J. Sivic and C. Schmid},
  title = {Vid2Seq: Large-scale pretraining of a visual language model for dense video captioning},
  booktitle = {Proc. IEEE/CVF Conf. Comput. Vis. Pattern Recognit. (CVPR)},
  pages = {10714--10726},
  year = {2023},
}

@inproceedings{zhao2023lavila,
  author = {Y. Zhao and I. Misra and P. Kr\"ahenb\"uhl and R. Girdhar},
  title = {Learning video representations from large language models},
  booktitle = {Proc. IEEE/CVF Conf. Comput. Vis. Pattern Recognit. (CVPR)},
  pages = {6586--6597},
  year = {2023},
}

@inproceedings{bain2021frozen,
  author = {M. Bain and A. Nagrani and G. Varol and A. Zisserman},
  title = {Frozen in time: A joint video and image encoder for end-to-end retrieval},
  booktitle = {Proc. IEEE/CVF Int. Conf. Comput. Vis. (ICCV)},
  pages = {1728--1738},
  year = {2021},
}

@inproceedings{pramanick2023egovlpv2,
  author = {S. Pramanick and Y. Song and S. Nag and K. Q. Lin and H. Shah and M. Z. Shou and R. Chellappa and P. Zhang},
  title = {EgoVLPv2: Egocentric video-language pre-training with fusion in the backbone},
  booktitle = {Proc. IEEE/CVF Int. Conf. Comput. Vis. (ICCV)},
  pages = {5285--5297},
  year = {2023},
}

@inproceedings{wray2021semantic,
  author = {M. Wray and H. Doughty and D. Damen},
  title = {On semantic similarity in video retrieval},
  booktitle = {Proc. IEEE/CVF Conf. Comput. Vis. Pattern Recognit. (CVPR)},
  pages = {3650--3660},
  year = {2021},
}

@inproceedings{mangalam2023egoschema,
  author = {K. Mangalam and R. Akshulakov and J. Malik},
  title = {EgoSchema: A diagnostic benchmark for very long-form video language understanding},
  booktitle = {Adv. Neural Inf. Process. Syst. (NeurIPS)},
  volume = {36},
  pages = {46212--46244},
  year = {2023},
}

@inproceedings{shan2020understanding,
  author = {D. Shan and J. Geng and M. Shu and D. F. Fouhey},
  title = {Understanding human hands in contact at internet scale},
  booktitle = {Proc. IEEE/CVF Conf. Comput. Vis. Pattern Recognit. (CVPR)},
  pages = {9869--9878},
  year = {2020},
}

@inproceedings{cheng2023hands23,
  author = {T. Cheng and D. Shan and A. S. Hassen and R. E. L. Higgins and D. Fouhey},
  title = {Towards a richer 2D understanding of hands at scale},
  booktitle = {Adv. Neural Inf. Process. Syst. (NeurIPS)},
  volume = {36},
  year = {2023},
}

@inproceedings{darkhalil2022visor,
  author = {A. Darkhalil and D. Shan and B. Zhu and J. Ma and A. Kar and R. Higgins and S. Fidler and D. Fouhey and D. Damen},
  title = {EPIC-KITCHENS VISOR benchmark: VIdeo Segmentations and Object Relations},
  booktitle = {Adv. Neural Inf. Process. Syst. (NeurIPS), Datasets and Benchmarks Track},
  volume = {35},
  year = {2022},
}

@inproceedings{liu2022hoi4d,
  author = {Y. Liu and Y. Liu and C. Jiang and K. Lyu and W. Wan and H. Shen and B. Liang and Z. Fu and H. Wang and L. Yi},
  title = {HOI4D: A 4D egocentric dataset for category-level human-object interaction},
  booktitle = {Proc. IEEE/CVF Conf. Comput. Vis. Pattern Recognit. (CVPR)},
  pages = {21013--21022},
  year = {2022},
}

@inproceedings{sener2022assembly101,
  author = {F. Sener and D. Chatterjee and D. Shelepov and K. He and D. Singhania and R. Wang and A. Yao},
  title = {Assembly101: A large-scale multi-view video dataset for understanding procedural activities},
  booktitle = {Proc. IEEE/CVF Conf. Comput. Vis. Pattern Recognit. (CVPR)},
  pages = {21096--21106},
  year = {2022},
}

@inproceedings{banerjee2024hot3d,
  author = {P. Banerjee and S. Shkodrani and P. Moulon and S. Hampali and S. Han and F. Zhang and L. Zhang and J. Fountain and E. Miller and S. Basol and R. Newcombe and R. Wang and J. J. Engel and T. Hodan},
  title = {HOT3D: Hand and object tracking in 3D from egocentric multi-view videos},
  booktitle = {Proc. IEEE/CVF Conf. Comput. Vis. Pattern Recognit. (CVPR)},
  year = {2025},
}

@misc{chavis2025dexfocus,
  author = {Z. Chavis and S. J. Guy and H. S. Park},
  title = {Improving keystep recognition in ego-video via dexterous focus},
  year = {2025},
  eprint = {2506.00827},
  eprinttype = {arxiv},
}

@inproceedings{tan2023egodistill,
  author = {S. Tan and T. Nagarajan and K. Grauman},
  title = {EgoDistill: Egocentric head motion distillation for efficient video understanding},
  booktitle = {Adv. Neural Inf. Process. Syst. (NeurIPS)},
  year = {2023},
}

@inproceedings{tang2023egotracks,
  author = {H. Tang and K. J. Liang and K. Grauman and M. Feiszli and W. Wang},
  title = {EgoTracks: A long-term egocentric visual object tracking dataset},
  booktitle = {Adv. Neural Inf. Process. Syst. (NeurIPS) Datasets and Benchmarks Track},
  year = {2023},
}

@inproceedings{pavlakos2024hamer,
  author = {G. Pavlakos and D. Shan and I. Radosavovic and A. Kanazawa and D. Fouhey and J. Malik},
  title = {Reconstructing hands in 3D with transformers},
  booktitle = {Proc. IEEE/CVF Conf. Comput. Vis. Pattern Recognit. (CVPR)},
  pages = {9826--9836},
  month = jun,
  year = {2024},
}

@inproceedings{fan2023arctic,
  author = {Z. Fan and O. Taheri and D. Tzionas and M. Kocabas and M. Kaufmann and M. J. Black and O. Hilliges},
  title = {ARCTIC: A dataset for dexterous bimanual hand-object manipulation},
  booktitle = {Proc. IEEE/CVF Conf. Comput. Vis. Pattern Recognit. (CVPR)},
  pages = {12943--12954},
  month = jun,
  year = {2023},
}

@article{thakur2024anacto,
  author = {S. Thakur and C. Beyan and P. Morerio and V. Murino and A. Del Bue},
  title = {Anticipating next active objects for egocentric videos},
  journal = {IEEE Access},
  volume = {12},
  pages = {51530--51546},
  year = {2024},
}

@inproceedings{wu2022memvit,
  author = {C.-Y. Wu and Y. Li and K. Mangalam and H. Fan and B. Xiong and J. Malik and C. Feichtenhofer},
  title = {MeMViT: Memory-augmented multiscale vision transformer for efficient long-term video recognition},
  booktitle = {Proc. IEEE/CVF Conf. Comput. Vis. Pattern Recognit. (CVPR)},
  pages = {13587--13597},
  month = jun,
  year = {2022},
}

@inproceedings{balazevic2024mcvit,
  author = {I. Bala\v{z}evi\'{c} and Y. Shi and P. Papalampidi and R. Chaabouni and S. Koppula and O. J. H\'{e}naff},
  title = {Memory consolidation enables long-context video understanding},
  booktitle = {Proc. Int. Conf. Mach. Learn. (ICML)},
  year = {2024},
}

@inproceedings{kwon2021h2o,
  author = {T. Kwon and B. Tekin and J. St\"uhmer and F. Bogo and M. Pollefeys},
  title = {H2O: Two hands manipulating objects for first person interaction recognition},
  booktitle = {Proc. IEEE/CVF Int. Conf. Comput. Vis. (ICCV)},
  pages = {10138--10148},
  month = oct,
  year = {2021},
}

@misc{sigurdsson2018charadesego,
  author = {G. A. Sigurdsson and A. Gupta and C. Schmid and A. Farhadi and K. Alahari},
  title = {Charades-Ego: A large-scale dataset of paired third and first person videos},
  year = {2018},
  eprint = {1804.09626},
  eprinttype = {arxiv},
}

@inproceedings{torralba2011unbiased,
  author = {A. Torralba and A. A. Efros},
  title = {Unbiased look at dataset bias},
  booktitle = {Proc. IEEE Conf. Comput. Vis. Pattern Recognit. (CVPR)},
  pages = {1521--1528},
  month = jun,
  year = {2011},
}

@inproceedings{karpathy2014largescale,
  author = {A. Karpathy and G. Toderici and S. Shetty and T. Leung and R. Sukthankar and L. Fei-Fei},
  title = {Large-scale video classification with convolutional neural networks},
  booktitle = {Proc. IEEE Conf. Comput. Vis. Pattern Recognit. (CVPR)},
  pages = {1725--1732},
  month = jun,
  year = {2014},
}

@inproceedings{tran2015c3d,
  author = {D. Tran and L. Bourdev and R. Fergus and L. Torresani and M. Paluri},
  title = {Learning spatiotemporal features with 3D convolutional networks},
  booktitle = {Proc. IEEE Int. Conf. Comput. Vis. (ICCV)},
  pages = {4489--4497},
  month = dec,
  year = {2015},
}

@inproceedings{qiu2017p3d,
  author = {Z. Qiu and T. Yao and T. Mei},
  title = {Learning spatio-temporal representation with pseudo-3D residual networks},
  booktitle = {Proc. IEEE Int. Conf. Comput. Vis. (ICCV)},
  pages = {5533--5541},
  month = oct,
  year = {2017},
}

@inproceedings{tran2018r2plus1d,
  author = {D. Tran and H. Wang and L. Torresani and J. Ray and Y. LeCun and M. Paluri},
  title = {A closer look at spatiotemporal convolutions for action recognition},
  booktitle = {Proc. IEEE/CVF Conf. Comput. Vis. Pattern Recognit. (CVPR)},
  pages = {6450--6459},
  month = jun,
  year = {2018},
}

@inproceedings{wang2018nonlocal,
  author = {X. Wang and R. Girshick and A. Gupta and K. He},
  title = {Non-local neural networks},
  booktitle = {Proc. IEEE/CVF Conf. Comput. Vis. Pattern Recognit. (CVPR)},
  pages = {7794--7803},
  month = jun,
  year = {2018},
}

@inproceedings{lin2019tsm,
  author = {J. Lin and C. Gan and S. Han},
  title = {TSM: Temporal shift module for efficient video understanding},
  booktitle = {Proc. IEEE/CVF Int. Conf. Comput. Vis. (ICCV)},
  pages = {7083--7093},
  month = oct,
  year = {2019},
}

@inproceedings{feichtenhofer2020x3d,
  author = {C. Feichtenhofer},
  title = {X3D: Expanding architectures for efficient video recognition},
  booktitle = {Proc. IEEE/CVF Conf. Comput. Vis. Pattern Recognit. (CVPR)},
  pages = {203--213},
  month = jun,
  year = {2020},
}

@inproceedings{donahue2015lrcn,
  author = {J. Donahue and L. A. Hendricks and S. Guadarrama and M. Rohrbach and S. Venugopalan and K. Saenko and T. Darrell},
  title = {Long-term recurrent convolutional networks for visual recognition and description},
  booktitle = {Proc. IEEE Conf. Comput. Vis. Pattern Recognit. (CVPR)},
  pages = {2625--2634},
  month = jun,
  year = {2015},
}

@inproceedings{ng2015beyond,
  author = {J. Y.-H. Ng and M. Hausknecht and S. Vijayanarasimhan and O. Vinyals and R. Monga and G. Toderici},
  title = {Beyond short snippets: Deep networks for video classification},
  booktitle = {Proc. IEEE Conf. Comput. Vis. Pattern Recognit. (CVPR)},
  pages = {4694--4702},
  month = jun,
  year = {2015},
}

@inproceedings{srivastava2015unsupervised,
  author = {N. Srivastava and E. Mansimov and R. Salakhutdinov},
  title = {Unsupervised learning of video representations using LSTMs},
  booktitle = {Proc. 32nd Int. Conf. Mach. Learn. (ICML)},
  volume = {37},
  pages = {843--852},
  year = {2015},
}

@inproceedings{shi2015convlstm,
  author = {X. Shi and Z. Chen and H. Wang and D.-Y. Yeung and W.-K. Wong and W.-C. Woo},
  title = {Convolutional LSTM network: A machine learning approach for precipitation nowcasting},
  booktitle = {Adv. Neural Inf. Process. Syst. (NeurIPS)},
  volume = {28},
  pages = {802--810},
  year = {2015},
}

@article{li2018videolstm,
  author = {Z. Li and K. Gavrilyuk and E. Gavves and M. Jain and C. G. M. Snoek},
  title = {VideoLSTM convolves, attends and flows for action recognition},
  journal = {Comput. Vis. Image Underst. (CVIU)},
  volume = {166},
  pages = {41--50},
  year = {2018},
}

@inproceedings{singh2016fpv,
  author = {S. Singh and C. Arora and C. V. Jawahar},
  title = {First person action recognition using deep learned descriptors},
  booktitle = {Proc. IEEE Conf. Comput. Vis. Pattern Recognit. (CVPR)},
  pages = {2620--2628},
  month = jun,
  year = {2016},
}

@inproceedings{ma2016deeperfpv,
  author = {M. Ma and H. Fan and K. M. Kitani},
  title = {Going deeper into first-person activity recognition},
  booktitle = {Proc. IEEE Conf. Comput. Vis. Pattern Recognit. (CVPR)},
  pages = {1894--1903},
  month = jun,
  year = {1903},
  note = {2016, pp. 1894--},
}

@inproceedings{sudhakaran2019lsta,
  author = {S. Sudhakaran and S. Escalera and O. Lanz},
  title = {LSTA: Long short-term attention for egocentric action recognition},
  booktitle = {Proc. IEEE/CVF Conf. Comput. Vis. Pattern Recognit. (CVPR)},
  pages = {9954--9963},
  month = jun,
  year = {2019},
}

@inproceedings{simonyan2014twostream,
  author = {K. Simonyan and A. Zisserman},
  title = {Two-stream convolutional networks for action recognition in videos},
  booktitle = {Adv. Neural Inf. Process. Syst. (NeurIPS)},
  volume = {27},
  pages = {568--576},
  year = {2014},
}

@inproceedings{feichtenhofer2016fusion,
  author = {C. Feichtenhofer and A. Pinz and A. Zisserman},
  title = {Convolutional two-stream network fusion for video action recognition},
  booktitle = {Proc. IEEE Conf. Comput. Vis. Pattern Recognit. (CVPR)},
  pages = {1933--1941},
  month = jun,
  year = {1941},
  note = {2016, pp. 1933--},
}

@inproceedings{feichtenhofer2016stresnet,
  author = {C. Feichtenhofer and A. Pinz and R. P. Wildes},
  title = {Spatiotemporal residual networks for video action recognition},
  booktitle = {Adv. Neural Inf. Process. Syst. (NeurIPS)},
  volume = {29},
  year = {2016},
}

@inproceedings{zhou2018trn,
  author = {B. Zhou and A. Andonian and A. Oliva and A. Torralba},
  title = {Temporal relational reasoning in videos},
  booktitle = {Proc. Eur. Conf. Comput. Vis. (ECCV)},
  volume = {11205},
  pages = {831--846},
  year = {2018},
  note = {LNCS},
}

@inproceedings{girdhar2017actionvlad,
  author = {R. Girdhar and D. Ramanan and A. Gupta and J. Sivic and B. Russell},
  title = {ActionVLAD: Learning spatio-temporal aggregation for action classification},
  booktitle = {Proc. IEEE Conf. Comput. Vis. Pattern Recognit. (CVPR)},
  pages = {971--980},
  month = jul,
  year = {2017},
}

@inproceedings{kazakos2019epicfusion,
  author = {E. Kazakos and A. Nagrani and A. Zisserman and D. Damen},
  title = {EPIC-Fusion: Audio-visual temporal binding for egocentric action recognition},
  booktitle = {Proc. IEEE/CVF Int. Conf. Comput. Vis. (ICCV)},
  pages = {5491--5500},
  month = oct,
  year = {2019},
}

@inproceedings{vaswani2017attention,
  author = {A. Vaswani and N. Shazeer and N. Parmar and J. Uszkoreit and L. Jones and A. N. Gomez and \L. Kaiser and I. Polosukhin},
  title = {Attention is all you need},
  booktitle = {Adv. Neural Inf. Process. Syst. (NeurIPS)},
  volume = {30},
  pages = {5998--6008},
  year = {2017},
}

@inproceedings{dosovitskiy2021vit,
  author = {A. Dosovitskiy and L. Beyer and A. Kolesnikov and D. Weissenborn and X. Zhai and T. Unterthiner and M. Dehghani and M. Minderer and G. Heigold and S. Gelly and J. Uszkoreit and N. Houlsby},
  title = {An image is worth 16x16 words: Transformers for image recognition at scale},
  booktitle = {Proc. Int. Conf. Learn. Represent. (ICLR)},
  year = {2021},
}

@inproceedings{touvron2021deit,
  author = {H. Touvron and M. Cord and M. Douze and F. Massa and A. Sablayrolles and H. J\'{e}gou},
  title = {Training data-efficient image transformers \& distillation through attention},
  booktitle = {Proc. Int. Conf. Mach. Learn. (ICML)},
  volume = {139},
  pages = {10347--10357},
  year = {2021},
}

@inproceedings{liu2021swin,
  author = {Z. Liu and Y. Lin and Y. Cao and H. Hu and Y. Wei and Z. Zhang and S. Lin and B. Guo},
  title = {Swin Transformer: Hierarchical vision transformer using shifted windows},
  booktitle = {Proc. IEEE/CVF Int. Conf. Comput. Vis. (ICCV)},
  pages = {9992--10002},
  month = oct,
  year = {2021},
}

@inproceedings{bertasius2021timesformer,
  author = {G. Bertasius and H. Wang and L. Torresani},
  title = {Is space-time attention all you need for video understanding?},
  booktitle = {Proc. Int. Conf. Mach. Learn. (ICML)},
  volume = {139},
  pages = {813--824},
  year = {2021},
}

@inproceedings{he2022mae,
  author = {K. He and X. Chen and S. Xie and Y. Li and P. Doll\'{a}r and R. Girshick},
  title = {Masked autoencoders are scalable vision learners},
  booktitle = {Proc. IEEE/CVF Conf. Comput. Vis. Pattern Recognit. (CVPR)},
  pages = {16000--16009},
  month = jun,
  year = {2022},
}

@inproceedings{tong2022videomae,
  author = {Z. Tong and Y. Song and J. Wang and L. Wang},
  title = {VideoMAE: Masked autoencoders are data-efficient learners for self-supervised video pre-training},
  booktitle = {Adv. Neural Inf. Process. Syst. (NeurIPS)},
  volume = {35},
  pages = {10078--10093},
  year = {2022},
}

@inproceedings{feichtenhofer2022stmae,
  author = {C. Feichtenhofer and H. Fan and Y. Li and K. He},
  title = {Masked autoencoders as spatiotemporal learners},
  booktitle = {Adv. Neural Inf. Process. Syst. (NeurIPS)},
  volume = {35},
  pages = {35946--35958},
  year = {2022},
}

@inproceedings{wang2023videomaev2,
  author = {L. Wang and B. Huang and Z. Zhao and Z. Tong and Y. He and Y. Wang and Y. Wang and Y. Qiao},
  title = {VideoMAE V2: Scaling video masked autoencoders with dual masking},
  booktitle = {Proc. IEEE/CVF Conf. Comput. Vis. Pattern Recognit. (CVPR)},
  pages = {14549--14560},
  month = jun,
  year = {2023},
}

@inproceedings{arnab2021vivit,
  author = {A. Arnab and M. Dehghani and G. Heigold and C. Sun and M. Lu\v{c}i\'{c} and C. Schmid},
  title = {ViViT: A video vision transformer},
  booktitle = {Proc. IEEE/CVF Int. Conf. Comput. Vis. (ICCV)},
  pages = {6836--6846},
  month = oct,
  year = {2021},
}

@inproceedings{li2022mvitv2,
  author = {Y. Li and C.-Y. Wu and H. Fan and K. Mangalam and B. Xiong and J. Malik and C. Feichtenhofer},
  title = {MViTv2: Improved multiscale vision transformers for classification and detection},
  booktitle = {Proc. IEEE/CVF Conf. Comput. Vis. Pattern Recognit. (CVPR)},
  pages = {4804--4814},
  month = jun,
  year = {2022},
}

@inproceedings{patrick2021motionformer,
  author = {M. Patrick and D. Campbell and Y. M. Asano and I. Misra and F. Metze and C. Feichtenhofer and A. Vedaldi and J. F. Henriques},
  title = {Keeping your eye on the ball: Trajectory attention in video transformers},
  booktitle = {Adv. Neural Inf. Process. Syst. (NeurIPS)},
  volume = {34},
  pages = {12493--12506},
  year = {2021},
}

@inproceedings{wei2022maskfeat,
  author = {C. Wei and H. Fan and S. Xie and C.-Y. Wu and A. Yuille and C. Feichtenhofer},
  title = {Masked feature prediction for self-supervised visual pre-training},
  booktitle = {Proc. IEEE/CVF Conf. Comput. Vis. Pattern Recognit. (CVPR)},
  pages = {14668--14678},
  month = jun,
  year = {2022},
}

@article{li2023uniformer,
  author = {K. Li and Y. Wang and J. Zhang and P. Gao and G. Song and Y. Liu and H. Li and Y. Qiao},
  title = {UniFormer: Unifying convolution and self-attention for visual recognition},
  journal = {IEEE Trans. Pattern Anal. Mach. Intell. (TPAMI)},
  volume = {45},
  number = {10},
  pages = {12581--12600},
  month = oct,
  year = {2023},
}

@inproceedings{yan2022mtv,
  author = {S. Yan and X. Xiong and A. Arnab and Z. Lu and M. Zhang and C. Sun and C. Schmid},
  title = {Multiview transformers for video recognition},
  booktitle = {Proc. IEEE/CVF Conf. Comput. Vis. Pattern Recognit. (CVPR)},
  pages = {3333--3343},
  month = jun,
  year = {2022},
}

@inproceedings{ryoo2021tokenlearner,
  author = {M. S. Ryoo and AJ Piergiovanni and A. Arnab and M. Dehghani and A. Angelova},
  title = {TokenLearner: Adaptive space-time tokenization for videos},
  booktitle = {Adv. Neural Inf. Process. Syst. (NeurIPS)},
  volume = {34},
  pages = {12786--12797},
  year = {2021},
}

@inproceedings{videoclip2021,
  author = {H. Xu and G. Ghosh and P.-Y. Huang and D. Okhonko and A. Aghajanyan and F. Metze and L. Zettlemoyer and C. Feichtenhofer},
  title = {VideoCLIP: Contrastive pre-training for zero-shot video-text understanding},
  booktitle = {Proc. Conf. Empirical Methods Nat. Lang. Process. (EMNLP)},
  pages = {6787--6800},
  year = {2021},
}

@inproceedings{zhang2023helpinghands,
  author = {C.~Zhang and A.~Gupta and A.~Zisserman},
  title = {Helping hands: An object-aware ego-centric video recognition model},
  booktitle = {Proc. IEEE/CVF Int. Conf. Comput. Vis. (ICCV)},
  pages = {13901--13912},
  month = oct,
  year = {2023},
}

@inproceedings{girdhar2023imagebind,
  author = {R.~Girdhar and A.~El-Nouby and Z.~Liu and M.~Singh and K.~V.~Alwala and A.~Joulin and I.~Misra},
  title = {ImageBind: One embedding space to bind them all},
  booktitle = {Proc. IEEE/CVF Conf. Comput. Vis. Pattern Recognit. (CVPR)},
  pages = {15180--15190},
  month = jun,
  year = {2023},
}

@inproceedings{simon2017hand,
  author = {T. Simon and H. Joo and I. Matthews and Y. Sheikh},
  title = {Hand keypoint detection in single images using multiview bootstrapping},
  booktitle = {Proc. IEEE Conf. Comput. Vis. Pattern Recognit. (CVPR)},
  pages = {4645--4653},
  month = jul,
  year = {2017},
}

@article{cao2019openpose,
  author = {Z. Cao and G. Hidalgo and T. Simon and S.-E. Wei and Y. Sheikh},
  title = {OpenPose: Realtime multi-person 2D pose estimation using part affinity fields},
  journal = {IEEE Trans. Pattern Anal. Mach. Intell. (TPAMI)},
  volume = {43},
  number = {1},
  pages = {172--186},
  month = jan,
  year = {2021},
}

@misc{zhang2020mediapipe,
  author = {F. Zhang and V. Bazarevsky and A. Vakunov and A. Tkachenka and G. Sung and C.-L. Chang and M. Grundmann},
  title = {MediaPipe Hands: On-device real-time hand tracking},
  year = {2020},
  eprint = {2006.10214},
  eprinttype = {arxiv},
}

@article{romero2017mano,
  author = {J. Romero and D. Tzionas and M. J. Black},
  title = {Embodied hands: Modeling and capturing hands and bodies together},
  journal = {ACM Trans. Graph. (Proc. SIGGRAPH Asia)},
  volume = {36},
  number = {6},
  pages = {245:1--245:17},
  month = nov,
  year = {2017},
}

@inproceedings{moon2020interhand,
  author = {G. Moon and S.-I. Yu and H. Wen and T. Shiratori and K. M. Lee},
  title = {InterHand2.6M: A dataset and baseline for 3D interacting hand pose estimation from a single RGB image},
  booktitle = {Proc. Eur. Conf. Comput. Vis. (ECCV)},
  volume = {12365},
  pages = {548--564},
  year = {2020},
  note = {LNCS},
}

@inproceedings{ohkawa2023assemblyhands,
  author = {T. Ohkawa and K. He and F. Sener and T. Hodan and L. Tran and C. Keskin},
  title = {AssemblyHands: Towards egocentric activity understanding via 3D hand pose estimation},
  booktitle = {Proc. IEEE/CVF Conf. Comput. Vis. Pattern Recognit. (CVPR)},
  pages = {12999--13008},
  month = jun,
  year = {2023},
}

@inproceedings{lin2021metro,
  author = {K. Lin and L. Wang and Z. Liu},
  title = {End-to-end human pose and mesh reconstruction with transformers},
  booktitle = {Proc. IEEE/CVF Conf. Comput. Vis. Pattern Recognit. (CVPR)},
  pages = {1954--1963},
  month = jun,
  year = {1963},
  note = {2021, pp. 1954--},
}

@inproceedings{lin2021meshgraphormer,
  author = {K. Lin and L. Wang and Z. Liu},
  title = {Mesh Graphormer},
  booktitle = {Proc. IEEE/CVF Int. Conf. Comput. Vis. (ICCV)},
  pages = {12939--12948},
  month = oct,
  year = {2021},
}

@inproceedings{park2022handoccnet,
  author = {J. Park and Y. Oh and G. Moon and H. Choi and K. M. Lee},
  title = {HandOccNet: Occlusion-robust 3D hand mesh estimation network},
  booktitle = {Proc. IEEE/CVF Conf. Comput. Vis. Pattern Recognit. (CVPR)},
  pages = {1496--1505},
  month = jun,
  year = {2022},
}

@inproceedings{potamias2025wilor,
  author = {R. A. Potamias and J. Zhang and J. Deng and S. Zafeiriou},
  title = {WiLoR: End-to-end 3D hand localization and reconstruction in-the-wild},
  booktitle = {Proc. IEEE/CVF Conf. Comput. Vis. Pattern Recognit. (CVPR)},
  pages = {12242--12254},
  month = jun,
  year = {2025},
}

@inproceedings{zhang2025hawor,
  author = {J. Zhang and J. Deng and M. Ma and R. A. Potamias},
  title = {HaWoR: World-space hand motion reconstruction from egocentric videos},
  booktitle = {Proc. IEEE/CVF Conf. Comput. Vis. Pattern Recognit. (CVPR)},
  month = jun,
  year = {2025},
}

@inproceedings{yu2025dynhamr,
  author = {Z. Yu and S. Zafeiriou and T. Birdal},
  title = {Dyn-HaMR: Recovering 4D interacting hand motion from a dynamic camera},
  booktitle = {Proc. IEEE/CVF Conf. Comput. Vis. Pattern Recognit. (CVPR)},
  pages = {27716--27726},
  month = jun,
  year = {2025},
}

@inproceedings{prakash2024wildhands,
  author = {A. Prakash and R. Tu and M. Chang and S. Gupta},
  title = {3D hand pose estimation in everyday egocentric images},
  booktitle = {Proc. Eur. Conf. Comput. Vis. (ECCV)},
  volume = {15136},
  pages = {183--202},
  year = {2024},
  note = {LNCS},
}

@article{furnari2017next,
  author = {A. Furnari and S. Battiato and K. Grauman and G. M. Farinella},
  title = {Next-active-object prediction from egocentric videos},
  journal = {J. Vis. Commun. Image Represent.},
  volume = {49},
  pages = {401--411},
  year = {2017},
}

@inproceedings{ragusa2023stillfast,
  author = {F. Ragusa and G. M. Farinella and A. Furnari},
  title = {StillFast: An end-to-end approach for short-term object interaction anticipation},
  booktitle = {Proc. IEEE/CVF Conf. Comput. Vis. Pattern Recognit. Workshops (CVPRW)},
  pages = {3635--3644},
  month = jun,
  year = {2023},
}

@inproceedings{nagarajan2019hotspots,
  author = {T. Nagarajan and C. Feichtenhofer and K. Grauman},
  title = {Grounded human-object interaction hotspots from video},
  booktitle = {Proc. IEEE/CVF Int. Conf. Comput. Vis. (ICCV)},
  pages = {8688--8697},
  month = oct,
  year = {2019},
}

@inproceedings{nagarajan2020egotopo,
  author = {T. Nagarajan and Y. Li and C. Feichtenhofer and K. Grauman},
  title = {EGO-TOPO: Environment affordances from egocentric video},
  booktitle = {Proc. IEEE/CVF Conf. Comput. Vis. Pattern Recognit. (CVPR)},
  pages = {163--172},
  month = jun,
  year = {2020},
}

@inproceedings{luo2022affordance,
  author = {H. Luo and W. Zhai and J. Zhang and Y. Cao and D. Tao},
  title = {Learning affordance grounding from exocentric images},
  booktitle = {Proc. IEEE/CVF Conf. Comput. Vis. Pattern Recognit. (CVPR)},
  pages = {2252--2261},
  month = jun,
  year = {2022},
}

@article{feix2016grasp,
  author = {T. Feix and J. Romero and H.-B. Schmiedmayer and A. M. Dollar and D. Kragic},
  title = {The GRASP taxonomy of human grasp types},
  journal = {IEEE Trans. Human-Mach. Syst. (THMS)},
  volume = {46},
  number = {1},
  pages = {66--77},
  month = feb,
  year = {2016},
}

@inproceedings{brahmbhatt2020contactpose,
  author = {S. Brahmbhatt and C. Tang and C. D. Twigg and C. C. Kemp and J. Hays},
  title = {ContactPose: A dataset of grasps with object contact and hand pose},
  booktitle = {Proc. Eur. Conf. Comput. Vis. (ECCV)},
  volume = {12358},
  pages = {361--378},
  year = {2020},
  note = {LNCS},
}

@inproceedings{taheri2020grab,
  author = {O. Taheri and N. Ghorbani and M. J. Black and D. Tzionas},
  title = {GRAB: A dataset of whole-body human grasping of objects},
  booktitle = {Proc. Eur. Conf. Comput. Vis. (ECCV)},
  volume = {12349},
  pages = {581--600},
  year = {2020},
  note = {LNCS},
}

@inproceedings{ji2020actiongenome,
  author = {J. Ji and R. Krishna and L. Fei-Fei and J. C. Niebles},
  title = {Action Genome: Actions as compositions of spatio-temporal scene graphs},
  booktitle = {Proc. IEEE/CVF Conf. Comput. Vis. Pattern Recognit. (CVPR)},
  pages = {10236--10247},
  month = jun,
  year = {2020},
}

@inproceedings{herzig2022orvit,
  author = {R. Herzig and E. Ben-Avraham and K. Mangalam and A. Bar and G. Chechik and A. Rohrbach and T. Darrell and A. Globerson},
  title = {Object-Region Video Transformers},
  booktitle = {Proc. IEEE/CVF Conf. Comput. Vis. Pattern Recognit. (CVPR)},
  pages = {3148--3159},
  month = jun,
  year = {2022},
}

@inproceedings{radevski2021stlt,
  author = {G. Radevski and M.-F. Moens and T. Tuytelaars},
  title = {Revisiting spatio-temporal layouts for compositional action recognition},
  booktitle = {Proc. British Mach. Vis. Conf. (BMVC)},
  year = {2021},
}

@inproceedings{wang2018stgraph,
  author = {X. Wang and A. Gupta},
  title = {Videos as space-time region graphs},
  booktitle = {Proc. Eur. Conf. Comput. Vis. (ECCV)},
  volume = {11209},
  pages = {413--431},
  year = {2018},
  note = {LNCS},
}

@inproceedings{hoireasoning2022,
  author = {J. Ma and D. Damen},
  title = {Hand-object interaction reasoning},
  booktitle = {Proc. 18th IEEE Int. Conf. Advanced Video Signal-Based Surveillance (AVSS)},
  pages = {1--8},
  year = {2022},
}

@misc{dessalene2020egoomg,
  author = {E. Dessalene and M. Maynord and C. Devaraj and C. Ferm\"uller and Y. Aloimonos},
  title = {Egocentric object manipulation graphs},
  year = {2020},
  eprint = {2006.03201},
  eprinttype = {arxiv},
}

@inproceedings{rodin2024easg,
  author = {I. Rodin and A. Furnari and K. Min and S. Tripathi and G. M. Farinella},
  title = {Action scene graphs for long-form understanding of egocentric videos},
  booktitle = {Proc. IEEE/CVF Conf. Comput. Vis. Pattern Recognit. (CVPR)},
  pages = {18622--18632},
  month = jun,
  year = {2024},
}

@misc{rodin2025easgbench,
  author = {I. Rodin and T.-Y. Wu and K. Min and S. Nittur Sridhar and A. Furnari and S. Tripathi and G. M. Farinella},
  title = {EASG-Bench: Video Q\&A benchmark with egocentric action scene graphs},
  year = {2025},
  eprint = {2506.05787},
  eprinttype = {arxiv},
}

@inproceedings{rai2021homage,
  author = {N. Rai and H. Chen and J. Ji and R. Desai and K. Kozuka and S. Ishizaka and E. Adeli and J. C. Niebles},
  title = {Home Action Genome: Cooperative compositional action understanding},
  booktitle = {Proc. IEEE/CVF Conf. Comput. Vis. Pattern Recognit. (CVPR)},
  pages = {11184--11193},
  month = jun,
  year = {2021},
}

@inproceedings{misra2016shuffle,
  author = {I. Misra and C. L. Zitnick and M. Hebert},
  title = {Shuffle and learn: Unsupervised learning using temporal order verification},
  booktitle = {Proc. Eur. Conf. Comput. Vis. (ECCV)},
  volume = {9905},
  pages = {527--544},
  year = {2016},
  note = {LNCS},
}

@inproceedings{goletto2024amego,
  author = {G. Goletto and T. Nagarajan and G. Averta and D. Damen},
  title = {AMEGO: Active memory from long egocentric videos},
  booktitle = {Proc. Eur. Conf. Comput. Vis. (ECCV)},
  volume = {15071},
  pages = {92--110},
  year = {2024},
  note = {LNCS},
}

@inproceedings{plizzari2025egotempo,
  author = {C. Plizzari and A. Tonioni and Y. Xian and A. Kulshrestha and F. Tombari},
  title = {Omnia de EgoTempo: Benchmarking temporal understanding of multi-modal LLMs in egocentric videos},
  booktitle = {Proc. IEEE/CVF Conf. Comput. Vis. Pattern Recognit. (CVPR)},
  pages = {24129--24138},
  month = jun,
  year = {2025},
}

@inproceedings{yan2018stgcn,
  author = {S. Yan and Y. Xiong and D. Lin},
  title = {Spatial temporal graph convolutional networks for skeleton-based action recognition},
  booktitle = {Proc. 32nd AAAI Conf. Artif. Intell. (AAAI)},
  pages = {7444--7452},
  year = {2018},
}

@inproceedings{cong2021sttran,
  author = {Y. Cong and W. Liao and H. Ackermann and B. Rosenhahn and M. Y. Yang},
  title = {Spatial-temporal transformer for dynamic scene graph generation},
  booktitle = {Proc. IEEE/CVF Int. Conf. Comput. Vis. (ICCV)},
  pages = {16372--16382},
  month = oct,
  year = {2021},
}

@inproceedings{wang2024oed,
  author = {G. Wang and Z. Li and Q. Chen and Y. Liu},
  title = {OED: Towards one-stage end-to-end dynamic scene graph generation},
  booktitle = {Proc. IEEE/CVF Conf. Comput. Vis. Pattern Recognit. (CVPR)},
  pages = {27938--27947},
  month = jun,
  year = {2024},
}

@inproceedings{peddi2024scenesayer,
  author = {R. Peddi and S. Singh and Saurabh and P. Singla and V. Gogate},
  title = {Towards scene graph anticipation},
  booktitle = {Proc. Eur. Conf. Comput. Vis. (ECCV)},
  volume = {15146},
  pages = {159--175},
  year = {2024},
  note = {LNCS},
}

@inproceedings{nguyen2025hyperglm,
  author = {T.-T. Nguyen and P. Nguyen and J. Cothren and A. Yilmaz and K. Luu},
  title = {HyperGLM: HyperGraph for video scene graph generation and anticipation},
  booktitle = {Proc. IEEE/CVF Conf. Comput. Vis. Pattern Recognit. (CVPR)},
  pages = {29150--29160},
  month = jun,
  year = {2025},
}

@inproceedings{li2025visa,
  author = {Y. Li and Z. Li and H. Chen and L. Xu},
  title = {Unbiased video scene graph generation via visual and semantic dual debiasing},
  booktitle = {Proc. IEEE/CVF Conf. Comput. Vis. Pattern Recognit. (CVPR)},
  month = jun,
  year = {2025},
}

@article{li2024sggsurvey,
  author = {H. Li and G. Zhu and L. Zhang and Y. Jiang and Y. Dang and H. Hou and P. Shen and X. Zhao and S. A. A. Shah and M. Bennamoun},
  title = {Scene graph generation: A comprehensive survey},
  journal = {Neurocomputing},
  volume = {566},
  pages = {127052},
  year = {2024},
}

@inproceedings{li2025samjam,
  author = {J. Li and F. J. Pena Cantu and E. Yu and A. Wong and Y. Cui and Y. Chen},
  title = {SAMJAM: Zero-shot video scene graph generation for egocentric kitchen videos},
  booktitle = {Proc. IEEE/CVF Conf. Comput. Vis. Pattern Recognit. Workshops (CVPRW)},
  pages = {467--473},
  month = jun,
  year = {2025},
}

@inproceedings{yu2023sevila,
  author = {S. Yu and J. Cho and P. Yadav and M. Bansal},
  title = {Self-chained image-language model for video localization and question answering},
  booktitle = {Adv. Neural Inf. Process. Syst. (NeurIPS)},
  volume = {36},
  year = {2023},
}

@inproceedings{yu2025framevoyager,
  author = {S. Yu and C. Jin and H. Wang and Z. Chen and S. Jin and Z. Zuo and X. Xu and Z. Sun and B. Zhang and J. Wu and H. Zhang and Q. Sun},
  title = {Frame-Voyager: Learning to query frames for video large language models},
  booktitle = {Proc. Int. Conf. Learn. Represent. (ICLR)},
  year = {2025},
}

@inproceedings{qiu2025step,
  author = {H. Qiu and M. Gao and L. Qian and K. Pan and Q. Yu and J. Li and W. Wang and S. Tang and Y. Zhuang and T.-S. Chua},
  title = {STEP: Enhancing video-LLMs' compositional reasoning by spatio-temporal graph-guided self-training},
  booktitle = {Proc. IEEE/CVF Conf. Comput. Vis. Pattern Recognit. (CVPR)},
  pages = {3284--3294},
  month = jun,
  year = {2025},
}

@misc{zemskova2026focusgraph,
  author = {T. Zemskova and S. Andryushenko and I. Obrubov and V. Khoruzhaia and E. Eroshenko and E. Derevyanka and D. Yudin},
  title = {FocusGraph: Graph-structured frame selection for embodied long video question answering},
  year = {2026},
  eprint = {2603.04349},
  eprinttype = {arxiv},
}

@misc{hu2025graphvideoagent,
  author = {M. Chu and Y. Li and T.-S. Chua},
  title = {Understanding long videos via LLM-powered entity relation graphs},
  year = {2025},
  eprint = {2501.15953},
  eprinttype = {arxiv},
}

@inproceedings{pei2025egohod,
  author = {B. Pei and Y. Huang and J. Xu and G. Chen and Y. He and L. Yang and Y. Wang and W. Xie and Y. Qiao and F. Wu and L. Wang},
  title = {Modeling fine-grained hand-object dynamics for egocentric video representation learning},
  booktitle = {Proc. Int. Conf. Learn. Represent. (ICLR)},
  year = {2025},
}

@inproceedings{kipf2022savi,
  author = {T. Kipf and G. F. Elsayed and A. Mahendran and A. Stone and S. Sabour and G. Heigold and R. Jonschkowski and A. Dosovitskiy and K. Greff},
  title = {Conditional object-centric learning from video},
  booktitle = {Proc. Int. Conf. Learn. Represent. (ICLR)},
  year = {2022},
}

@inproceedings{zhang2024objectprompt,
  author = {C. Zhang and C. Fu and S. Wang and N. Agarwal and K. Lee and C. Choi and C. Sun},
  title = {Object-centric video representation for long-term action anticipation},
  booktitle = {Proc. IEEE/CVF Winter Conf. Appl. Comput. Vis. (WACV)},
  pages = {6751--6761},
  month = jan,
  year = {2024},
}

@inproceedings{li2021weaklyhoi,
  author = {S. Li and Y. Du and A. Torralba and J. Sivic and B. Russell},
  title = {Weakly supervised human-object interaction detection in video via contrastive spatiotemporal regions},
  booktitle = {Proc. IEEE/CVF Int. Conf. Comput. Vis. (ICCV)},
  pages = {1845--1855},
  month = oct,
  year = {2021},
}

@article{zhou2022coop,
  author = {K. Zhou and J. Yang and C. C. Loy and Z. Liu},
  title = {Learning to prompt for vision-language models},
  journal = {Int. J. Comput. Vis. (IJCV)},
  volume = {130},
  number = {9},
  pages = {2337--2348},
  year = {2022},
}

@inproceedings{zhou2022cocoop,
  author = {K. Zhou and J. Yang and C. C. Loy and Z. Liu},
  title = {Conditional prompt learning for vision-language models},
  booktitle = {Proc. IEEE/CVF Conf. Comput. Vis. Pattern Recognit. (CVPR)},
  pages = {16816--16825},
  month = jun,
  year = {2022},
}

@inproceedings{jia2022vpt,
  author = {M. Jia and L. Tang and B.-C. Chen and C. Cardie and S. Belongie and B. Hariharan and S.-N. Lim},
  title = {Visual prompt tuning},
  booktitle = {Proc. Eur. Conf. Comput. Vis. (ECCV)},
  volume = {13693},
  pages = {709--727},
  year = {2022},
  note = {LNCS},
}

@inproceedings{khattak2023maple,
  author = {M. U. Khattak and H. Rasheed and M. Maaz and S. Khan and F. S. Khan},
  title = {MaPLe: Multi-modal prompt learning},
  booktitle = {Proc. IEEE/CVF Conf. Comput. Vis. Pattern Recognit. (CVPR)},
  pages = {19113--19122},
  month = jun,
  year = {2023},
}

@misc{wang2021actionclip,
  author = {M. Wang and J. Xing and Y. Liu},
  title = {ActionCLIP: A new paradigm for video action recognition},
  year = {2021},
  eprint = {2109.08472},
  eprinttype = {arxiv},
}

@inproceedings{ju2022prompting,
  author = {C. Ju and T. Han and K. Zheng and Y. Zhang and W. Xie},
  title = {Prompting visual-language models for efficient video understanding},
  booktitle = {Proc. Eur. Conf. Comput. Vis. (ECCV)},
  volume = {13695},
  pages = {105--124},
  year = {2022},
  note = {LNCS},
}

@inproceedings{wasim2023vitaclip,
  author = {S. T. Wasim and M. Naseer and S. Khan and F. S. Khan and M. Shah},
  title = {Vita-CLIP: Video and text adaptive CLIP via multimodal prompting},
  booktitle = {Proc. IEEE/CVF Conf. Comput. Vis. Pattern Recognit. (CVPR)},
  pages = {23034--23044},
  month = jun,
  year = {2023},
}

@article{zhang2024vlmsurvey,
  author = {J. Zhang and J. Huang and S. Jin and S. Lu},
  title = {Vision-language models for vision tasks: A survey},
  journal = {IEEE Trans. Pattern Anal. Mach. Intell. (TPAMI)},
  volume = {46},
  number = {8},
  pages = {5625--5644},
  month = aug,
  year = {2024},
}

@inproceedings{xu2023pov,
  author = {B. Xu and S. Zheng and Q. Jin},
  title = {POV: Prompt-oriented view-agnostic learning for egocentric hand-object interaction in the multi-view world},
  booktitle = {Proc. ACM Int. Conf. Multimedia (ACM MM)},
  pages = {2807--2816},
  year = {2023},
}

@inproceedings{jia2024actionprompt,
  author = {C. Jia and M. Luo and Z. Chang and Z. Dang and M. Han and M. Wang and G. Dai and S. Dang and J. Wang},
  title = {Generating action-conditioned prompts for open-vocabulary video action recognition},
  booktitle = {Proc. ACM Int. Conf. Multimedia (ACM MM)},
  pages = {4640--4649},
  year = {2024},
}

@inproceedings{kundu2024discovering,
  author = {S. Kundu and S. Trehan and S. N. Aakur},
  title = {Discovering novel actions from open world egocentric videos with object-grounded visual commonsense reasoning},
  booktitle = {Proc. Eur. Conf. Comput. Vis. (ECCV)},
  year = {2024},
  note = {LNCS},
}

@misc{bansal2024hoiref,
  author = {S. Bansal and C. Wray and D. Damen},
  title = {HOI-Ref: Hand-object interaction referral in egocentric vision},
  year = {2024},
  eprint = {2404.09933},
  eprinttype = {arxiv},
}

@inproceedings{yang2025egolife,
  author = {J. Yang and S. Liu and H. Guo and Y. Dong and X. Zhang and S. Zhang and P. Wang and Z. Zhou and B. Xie and Z. Wang and B. Ouyang and Z. Lin and M. Cominelli and Z. Cai and Y. Zhang and P. Zhang and F. Hong and J. Widmer and F. Gringoli and L. Yang and B. Li and Z. Liu},
  title = {EgoLife: Towards egocentric life assistant},
  booktitle = {Proc. IEEE/CVF Conf. Comput. Vis. Pattern Recognit. (CVPR)},
  pages = {28885--28900},
  month = jun,
  year = {2025},
}

@inproceedings{wang2023holoassist,
  author = {X. Wang and T. Kwon and M. Rad and B. Pan and I. Chakraborty and S. Andrist and D. Bohus and A. Feniello and B. Tekin and F. V. Frujeri and N. Joshi and M. Pollefeys},
  title = {HoloAssist: An egocentric human interaction dataset for interactive AI assistants in the real world},
  booktitle = {Proc. IEEE/CVF Int. Conf. Comput. Vis. (ICCV)},
  pages = {20270--20281},
  month = oct,
  year = {2023},
}

@inproceedings{flaborea2024prego,
  author = {A. Flaborea and G. M. D'Amely di Melendugno and L. Plini and L. Scofano and E. De Matteis and A. Furnari and G. M. Farinella and F. Galasso},
  title = {PREGO: Online mistake detection in PRocedural EGOcentric videos},
  booktitle = {Proc. IEEE/CVF Conf. Comput. Vis. Pattern Recognit. (CVPR)},
  pages = {18483--18492},
  month = jun,
  year = {2024},
}

@inproceedings{lee2024egoper,
  author = {S.-P. Lee and Z. Lu and Z. Zhang and M. Hoai and E. Elhamifar},
  title = {Error detection in egocentric procedural task videos},
  booktitle = {Proc. IEEE/CVF Conf. Comput. Vis. Pattern Recognit. (CVPR)},
  pages = {18655--18666},
  month = jun,
  year = {2024},
}

@inproceedings{peddi2024captaincook4d,
  author = {R. Peddi and S. Saurabh and A. Vupputuri and T. Maganti and V. Gogate and S. Singh and P. Singla},
  title = {CaptainCook4D: A dataset for understanding errors in procedural activities},
  booktitle = {Adv. Neural Inf. Process. Syst. (NeurIPS) Datasets and Benchmarks Track},
  year = {2024},
}

@misc{kong2025ariagen2,
  author = {C. Kong and others},
  title = {Aria Gen 2 Pilot Dataset},
  year = {2025},
  eprint = {2510.16134},
  eprinttype = {arxiv},
}

@inproceedings{nair2022r3m,
  author = {S. Nair and A. Rajeswaran and V. Kumar and C. Finn and A. Gupta},
  title = {R3M: A universal visual representation for robot manipulation},
  booktitle = {Proc. Conf. Robot Learn. (CoRL)},
  volume = {205},
  pages = {892--909},
  year = {2022},
  note = {Proc. Mach. Learn. Res. (PMLR)},
}

@inproceedings{zitkovich2023rt2,
  author = {B. Zitkovich and T. Yu and S. Xu and P. Xu and T. Xiao and F. Xia and J. Wu and P. Wohlhart and S. Welker and A. Wahid and others},
  title = {RT-2: Vision-language-action models transfer web knowledge to robotic control},
  booktitle = {Proc. Conf. Robot Learn. (CoRL)},
  volume = {229},
  pages = {2165--2183},
  year = {2023},
  note = {Proc. Mach. Learn. Res. (PMLR)},
}

@inproceedings{kim2024openvla,
  author = {M. J. Kim and K. Pertsch and S. Karamcheti and T. Xiao and A. Balakrishna and S. Nair and R. Rafailov and E. Foster and G. Lam and P. Sanketi and Q. Vuong and T. Kollar and B. Burchfiel and R. Tedrake and D. Sadigh and S. Levine and P. Liang and C. Finn},
  title = {OpenVLA: An open-source vision-language-action model},
  booktitle = {Proc. Conf. Robot Learn. (CoRL)},
  year = {2024},
}

@misc{black2024pi0,
  author = {K. Black and N. Brown and D. Driess and A. Esmail and M. Equi and C. Finn and N. Fusai and L. Groom and K. Hausman and B. Ichter and S. Jakubczak and T. Jones and L. Ke and S. Levine and A. Li-Bell and M. Mothukuri and S. Nair and K. Pertsch and L. X. Shi and J. Tanner and Q. Vuong and A. Walling and H. Wang and U. Zhilinsky},
  title = {$\pi_0$: A vision-language-action flow model for general robot control},
  year = {2024},
  eprint = {2410.24164},
  eprinttype = {arxiv},
}

@misc{liu2025egozero,
  author = {V. Liu and A. Adeniji and H. Zhan and R. Bhirangi and P. Abbeel and L. Pinto},
  title = {EgoZero: Robot learning from smart glasses},
  year = {2025},
  eprint = {2505.20290},
  eprinttype = {arxiv},
}

@inproceedings{jiang2025dexmimicgen,
  author = {Z. Jiang and Y. Xie and K. Lin and Z. Xu and W. Wan and A. Mandlekar and L. Fan and Y. Zhu},
  title = {DexMimicGen: Automated data generation for bimanual dexterous manipulation via imitation learning},
  booktitle = {Proc. IEEE Int. Conf. Robot. Autom. (ICRA)},
  year = {2025},
}

@misc{vinod2025egovlm,
  author = {A. Vinod and S. Pandit and A. Vavre and L. Liu},
  title = {EgoVLM: Policy optimization for egocentric video understanding},
  year = {2025},
  eprint = {2506.03097},
  eprinttype = {arxiv},
}

@inproceedings{tu2025playerone,
  author = {Y. Tu and H. Luo and X. Chen and X. Bai and F. Wang and H. Zhao},
  title = {PlayerOne: Egocentric world simulator},
  booktitle = {Adv. Neural Inf. Process. Syst. (NeurIPS)},
  year = {2025},
  note = {. (Oral.)},
}

@misc{bai2025wholebody,
  author = {Y. Bai and D. Tran and A. Bar and Y. LeCun and T. Darrell and J. Malik},
  title = {Whole-body conditioned egocentric video prediction},
  year = {2025},
  eprint = {2506.21552},
  eprinttype = {arxiv},
}

@inproceedings{egoillusion2025,
  author = {A. Seth and U. Tyagi and R. Selvakumar and N. Anand and S. Kumar and S. Ghosh and R. Duraiswami and C. Agarwal and D. Manocha},
  title = {EGOILLUSION: Benchmarking hallucinations in egocentric video understanding},
  booktitle = {Proc. Conf. Empirical Methods Nat. Lang. Process. (EMNLP)},
  pages = {28461--28480},
  year = {2025},
}

\end{document}